\pdfoutput=1
\documentclass[nohyperref]{article}

\usepackage[final]{microtype}
\usepackage{graphicx}
\usepackage{subfigure}
\usepackage{booktabs} % for professional tables

\usepackage{hyperref}

\usepackage[accepted]{icml2022}
\usepackage{amsmath}
\usepackage{amssymb}
\usepackage{mathtools}
\usepackage{amsthm}
\usepackage{bbm}
\usepackage{enumitem}

\usepackage[capitalize,noabbrev]{cleveref}

\theoremstyle{plain}
\newtheorem{theorem}{Theorem}[section]

\newtheorem{lemma}[theorem]{Lemma}

\theoremstyle{definition}

\newtheorem{assumption}[theorem]{Assumption}
\theoremstyle{remark}

\newcommand{\grad}{\nabla}

\newcommand{\iprod}[2]{\langle #1, #2 \rangle}

\newcommand{\abs}[1]{\left|{#1}\right|}
\newcommand{\norm}[1]{\left\|{#1} \right\|}

\newcommand{\cO}{\mathcal{O}}

\newcommand{\cX}{\mathcal{X}}

\newcommand{\Z}{\mathbb{Z}}

\newcommand{\R}{\mathbb{R}}

\newcommand{\E}{\mathbb{E}}
\renewcommand{\Pr}{\mathbb{P}}

\newcommand{\indicator}{\mathbbm{1}}

\renewcommand{\L}{\mathcal{L}}

\newcommand{\cB}{\mathcal{B}}

\usepackage[textsize=tiny]{todonotes}

\icmltitlerunning{NN Weights Do Not Converge to Stationary Points}

\begin{document}

\twocolumn[
\icmltitle{
Neural Network Weights Do Not Converge to Stationary Points: 
\\An Invariant Measure Perspective
}

% It is OKAY to include author information, even for blind
% submissions: the style file will automatically remove it for you
% unless you've provided the [accepted] option to the icml2022
% package.

% List of affiliations: The first argument should be a (short)
% identifier you will use later to specify author affiliations
% Academic affiliations should list Department, University, City, Region, Country
% Industry affiliations should list Company, City, Region, Country

% You can specify symbols, otherwise they are numbered in order.
% Ideally, you should not use this facility. Affiliations will be numbered
% in order of appearance and this is the preferred way.
\icmlsetsymbol{equal}{*}

\begin{icmlauthorlist}
\icmlauthor{Jingzhao Zhang}{thu}
\icmlauthor{Haochuan Li}{MIT}
\icmlauthor{Suvrit Sra}{MIT}
\icmlauthor{Ali Jadbabaie}{MIT}

%\icmlauthor{}{sch}
\end{icmlauthorlist}

\icmlaffiliation{thu}{IIIS, Tsinghua University}
\icmlaffiliation{MIT}{Massachusetts Institute of Technology}

\icmlcorrespondingauthor{Jingzhao Zhang}{jingzhaoz@mail.tsinghua.edu.cn}
\icmlcorrespondingauthor{Haochuan Li}{haochuan@mit.edu}

% You may provide any keywords that you
% find helpful for describing your paper; these are used to populate
% the "keywords" metadata in the PDF but will not be shown in the document
\icmlkeywords{Nonconvex optimization, deep learning, gradient methods}

\vskip 0.3in
]

% this must go after the closing bracket ] following \twocolumn[ ...

% This command actually creates the footnote in the first column
% listing the affiliations and the copyright notice.
% The command takes one argument, which is text to display at the start of the footnote.
% The \icmlEqualContribution command is standard text for equal contribution.
% Remove it (just {}) if you do not need this facility.

\printAffiliationsAndNotice{}  % leave blank if no need to mention equal contribution
% \printAffiliationsAndNotice{\icmlEqualContribution} % otherwise use the standard text.

\begin{abstract}

This work examines the deep disconnect between existing theoretical analyses of gradient-based algorithms and the practice of training  deep neural networks. Specifically, we provide numerical evidence that in large-scale neural network training (e.g., ImageNet + ResNet101, and WT103 + TransformerXL models), the neural network's weights \emph{do not} converge to stationary points where the gradient of the loss is zero. Remarkably, however, we observe that even though the weights do not converge to stationary points, the progress in minimizing the loss function halts and training loss stabilizes. Inspired by this observation, we propose a new perspective based on ergodic theory of dynamical systems to explain it. Rather than studying the evolution of weights, we study the evolution of the distribution of weights. We prove convergence of the distribution of weights to an approximate invariant measure,  thereby explaining how the training loss can stabilize without weights necessarily converging to stationary points. We further discuss how this perspective can better align optimization theory with empirical observations in machine learning practice.

\end{abstract}

\section{Introduction}
It would not be controversial to claim that currently there exists a wide gulf between theoretical investigations of convergence to (approximate) stationary points for non-convex optimization problems and the empirical performance of popular algorithms used in deep learning practice. Due to the intrinsic intractability of general nonconvex problems, theoretical analysis of nonconvex optimization problems often focuses on the rates of convergence of gradient norm $\|\nabla f(\theta)\|$ instead of the suboptimality $f(\theta) - \min_\theta f(\theta)$. The vast theoretical literature on optimization for machine learning has documented the recent progress in this area. In particular, optimal gradient-based algorithms and rates have been identified in various nonconvex settings, including deterministic, stochastic and finite-sum problems~\citep{Carmon2017a, Arjevani2019, Fang2018a}.

In addition to theoretical interest in nonconvex problems, a practical motivation for studying nonconvex convergence analyses is to improve the large-scale optimization methods that are used in machine learning practice, especially in training deep neural networks.  As neural network models allow for efficient gradient evaluations, gradient-based algorithms remain the dominant methods to tune network parameters. Naturally, great effort has been dedicated to theoretical understanding of gradient-based optimizers. 

But despite the rapid progress in the theory of gradient-based algorithms, this theory has had a limited impact on real-world neural network training. And the gap between theory and practice is as wide as ever. For example, even though the variance reduction technique theoretically accelerates convergence, recent empirical evidence in~\citep{Defazio2018} suggests that it may be ineffective in speeding up neural network training. On the other extreme, ADAM~\citep{Kingma2014} is among the most popular algorithms in neural network training, yet its theoretical convergence was proven to be incorrect~\citep{Reddi2019}. Despite dubious theoretical properties, ADAM is still among the most effective optimizers.

Our goal is to address the ineffectiveness of applying theoretical convergence rates to stationarity in neural network training by identifying a fundamental gap between theoretical convergence and empirical convergence. First, we provide evidence that in many challenging experiments  (e.g., ImageNet, Wiki103) where the model does not overfit the data, gradient-based optimization methods do not converge to stationary points as theory mandates. This mismatch questions applicability of usual theory as applied to training neural networks.  The reason for such a surprising divide is that most optimization analyses for deep learning either assume smoothness directly which leads to convergence to stationary points using classical analyses, or prove smoothness and fast convergence by relying explicitly on overparametrization. However, our empirical investigations reveal that the key premise of the theory--pointwise convergence to a fixed point---\emph{may not happen at all in practice!}

Motivated by this observation, we aim to answer the following question in the rest of this work: \emph{how should one define and analyze the convergence of gradient-based optimization methods, when the training loss seems to converge, yet the gradient norm does not converge to $0$?}\footnotemark\footnotetext{More pedantically, the iterates do not converge to even an $\epsilon$-stationary point as predicted by standard theory.}

% \begin{center}
% \begin{minipage}{.9\linewidth}

% \end{minipage}
% \end{center}

We propose a new lens through which one should view convergence: rather than convergence of weights, we postulate that the convergence should be viewed in terms of invariant measures as used in the ergodic theory of dynamical systems. Building on classical results from this literature, we then show how this new perspective is also consistent with some curious findings in neural network training, such as relaxed smoothness~\citep{Zhang2019e} and edge of stability~\citep{Cohen2021, wu2018sgd}. More concretely, our contributions are summarized as follows:
\vspace{-0.1cm}
\begin{itemize}[leftmargin=*,itemsep=0pt]
	\item We empirically verify through ResNet training and Transformer-XL training in a wide range of applications that the iterates do not converge to a stationary point as existing theory predicts.
	\item We propose an invariant measure perspective from dynamical systems to explain why the training loss can converge without the iterates converging to stationary points.
	\item Most importantly, we show that our theorems on diminishing gain of the loss without vanishing of the gradient apply to neural network training even without standard global Lipschitzness or smoothness assumptions. 
\end{itemize}

It is worth noting that our analysis for deep learning, though holds under a very generic setup without assuming overfitting or bounded Lipschitzness, only states vanishing change in average training loss. It does not comment on the actual loss values. Consequently, much remains to be done based on our proposed view. However, our observations relate to interesting phenomena such as decay of function values, edge of stability, and relative smoothness. We conclude our work with a detailed discussion of the above points. We believe that it provides a paradigm shift in how convergence in deep learning should be defined and studied.

\vspace*{-5pt}
\subsection{Related work}

% Many recent results inspired us to investigate the oscillatory behavior of neural network training. One line of works is on the empirical investigations of neural network reproducibility. In~\citep{Henderson2017}, authors analyzed the stability of policy reward in reinforcement learning and found large variations. In~\citep{Madhyastha2019}, authors studied the instability for interpretation mechanisms. In~\citep{Bhojanapalli2021}, authors found that though image classification has relatively stable classification accuracy, the actual prediction on individual images has large variations. We learned from recent studies~\citep{Cohen2021,Zhang2019e, zhang2020stochastic} that previous analysis assumptions on noise and smoothness not only have large variations but also adapt to the step size choice. These observations motivate us to rethink the convergence proofs used in classical optimization analysis. In addition, a few very recent results reported simialrly large oscillations in Cifar10 training~\citep{li2020reconciling,kunin2021rethinking, lobacheva2021periodic}, though the authors focus on SDE approximation or batch normalization.  Our work instead focuses on the connection to nonconvex optimization theorems. Specifically we show that at the end of training when training loss has converged,  even the full batch gradient norm does not converge to zero.

Several recent empirical findings discuss the instability of neural network predictions even after training loss has converged, and they inspire us to investigate whether convergence to stationary points actually happens. \citet{Henderson2017} analyze the stability of policy reward in reinforcement learning and observe large variations. \citet{Madhyastha2019} study the instability for interpretation mechanisms. In~\citep{Bhojanapalli2021}, the authors note that though image classification has relatively stable accuracy, the actual prediction on individual images has large variation. A few very recent results report similar large oscillations in Cifar10 training~\citep{li2020reconciling,kunin2021rethinking, lobacheva2021periodic}, though the authors focus on SDE approximation or batch normalization. Our work instead focuses on the connection to nonconvex optimization theorems. In addition, we learned from recent studies~\citep{Cohen2021,Zhang2019e, zhang2020stochastic} that assumptions on noise and smoothness not only fail but can further adversarially adapt to the step size choice, further suggesting that optimizers may not find stationary points in deep learning.

On the theory side, two lines of work study convergence beyond finding stationary points, and hence are closely related to this paper. One line studies the non-convergence of dynamics of algorithms in games or multiobjective optimization~\citep{Hsieh2019, cheung2019vortices,papadimitriou2019game,letcher2020impossibility, flokas2020no}. Another models SGD dynamics via  Langevin dynamics~\citep{cheng2020stochastic, li2020reconciling, gurbuzbalaban2021heavy}. Our work differs from the Langevin dynamics view in that we do not aim to achieve global mixing. As a consequence, we avoid the unrealistic assumption in Langevin analysis that the noise level is inversely proportional to the step size.

% !TeX root = ../main.tex
\begin{figure*}[htbp]
	\centering	\vspace{-0.4cm}
	{
		\includegraphics[scale=0.23]{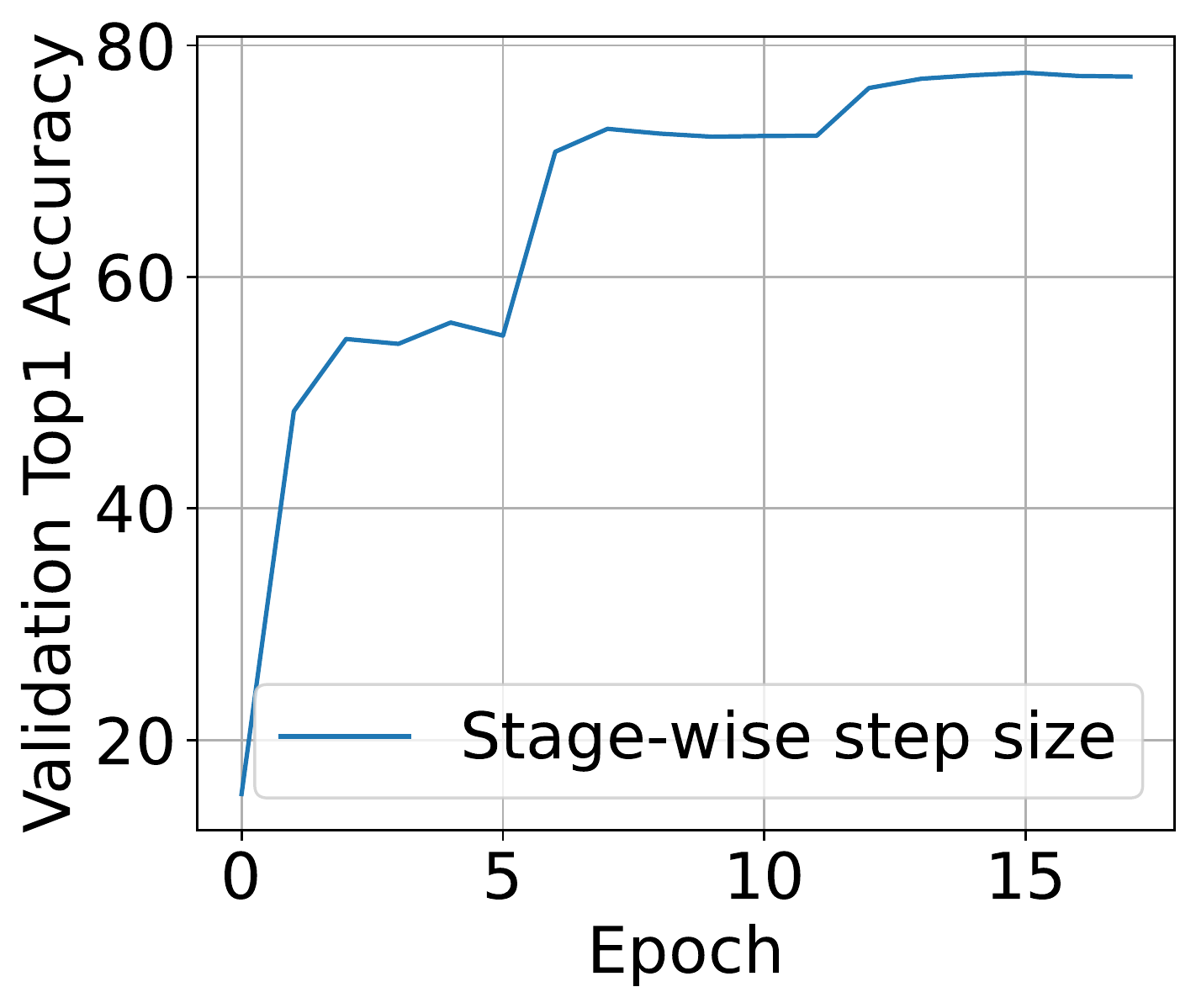}
	}
	{
		\includegraphics[scale=0.23]{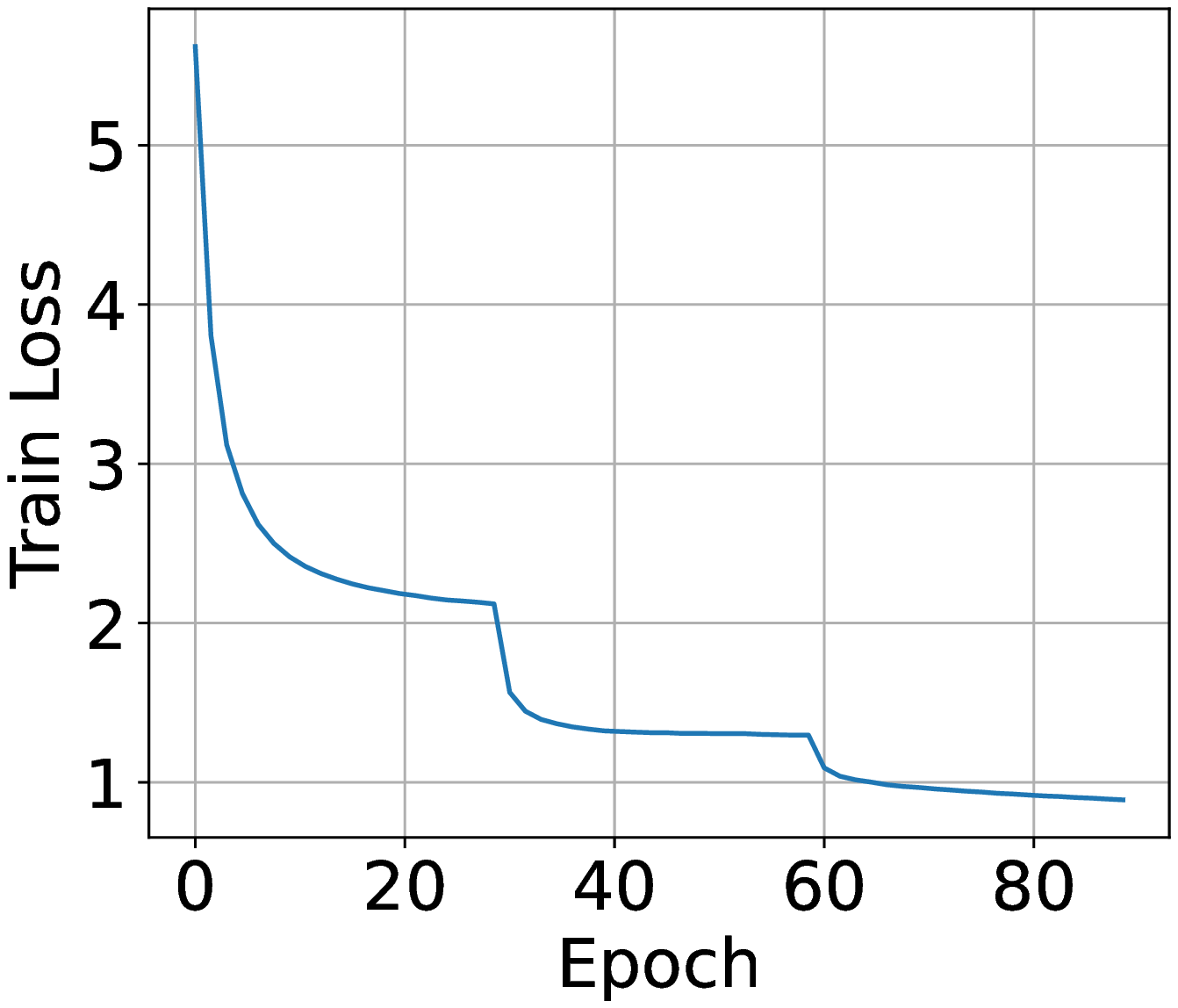}
	}
	{
		\includegraphics[scale=0.23]{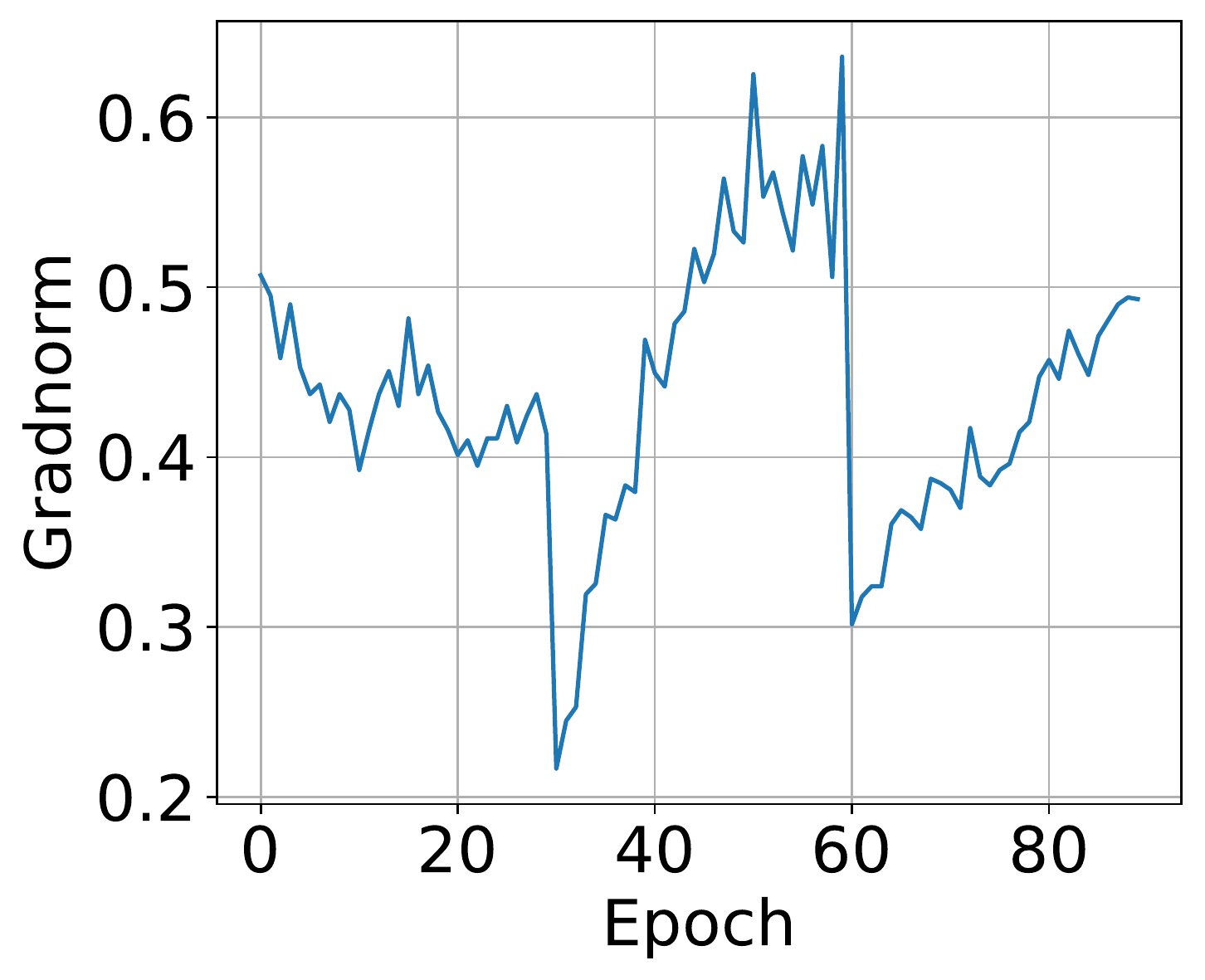}
	}
	{
		\includegraphics[scale=0.23]{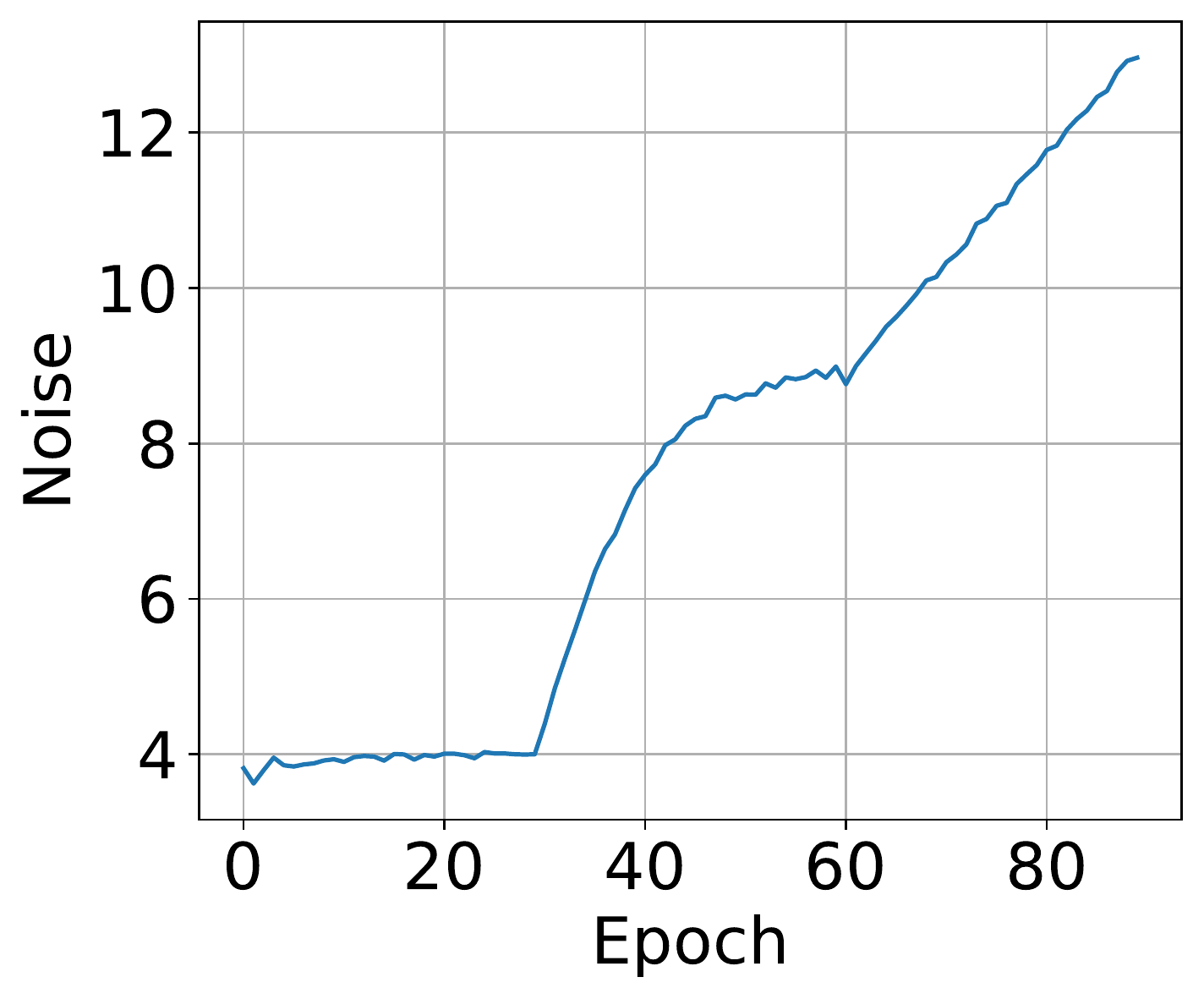}
	}
% 	{
% 		\includegraphics[scale=0.25]{figures/imagenet-baseline-train-sharpness}
% 	}
	\vspace{-0.4cm}
	\caption{
		The validation accuracy and the quantities of interest \eqref{eq:quantities} for the default training schedule of ImageNet + ResNet101 experiment.
	}
	\label{fig:imagenet-baseline}	

\end{figure*}

\begin{figure*}[htbp]
	\centering
	\vspace{-0.5cm}
	
	{
		\includegraphics[scale=0.23]{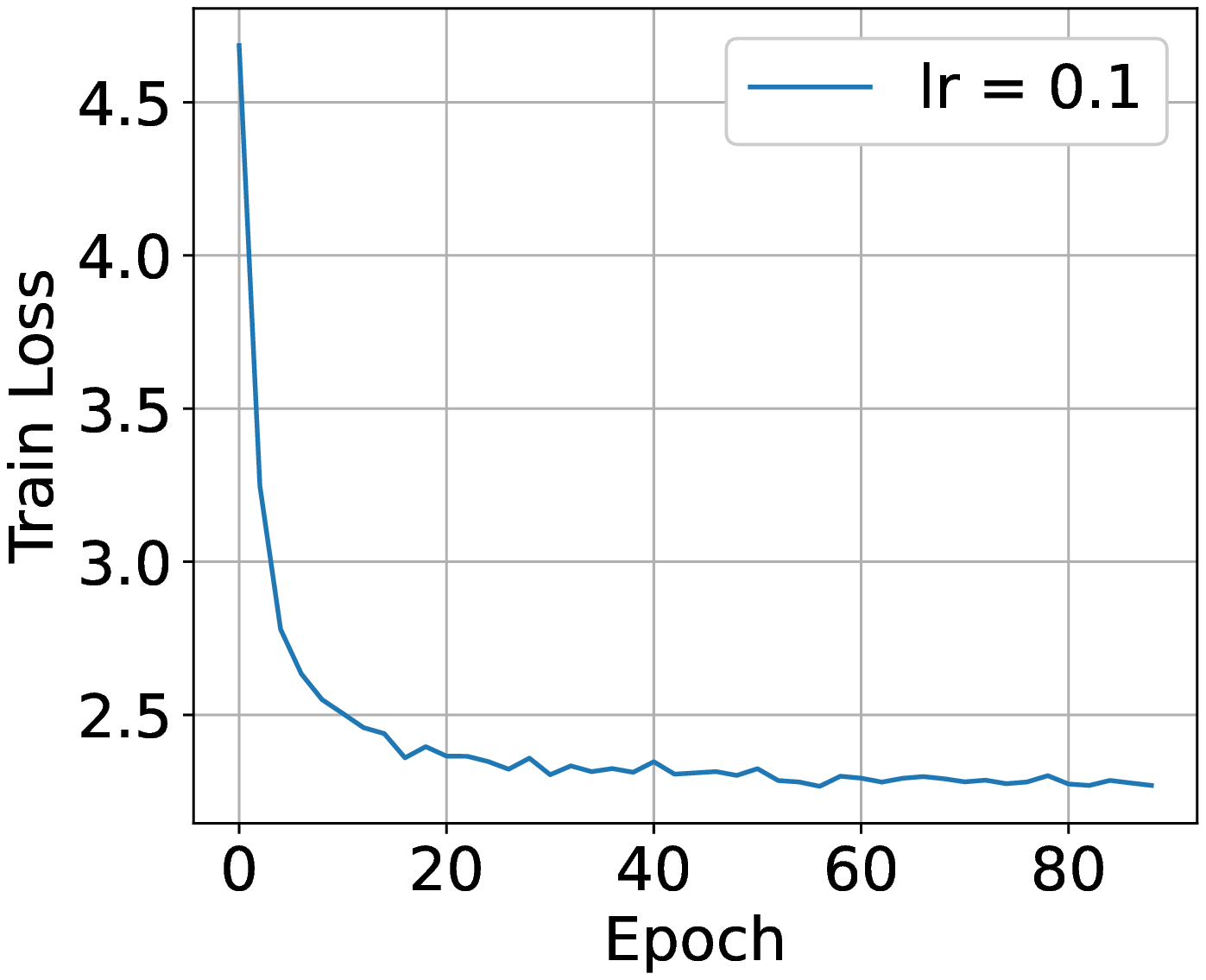}
	}	
	{
		\includegraphics[scale=0.23]{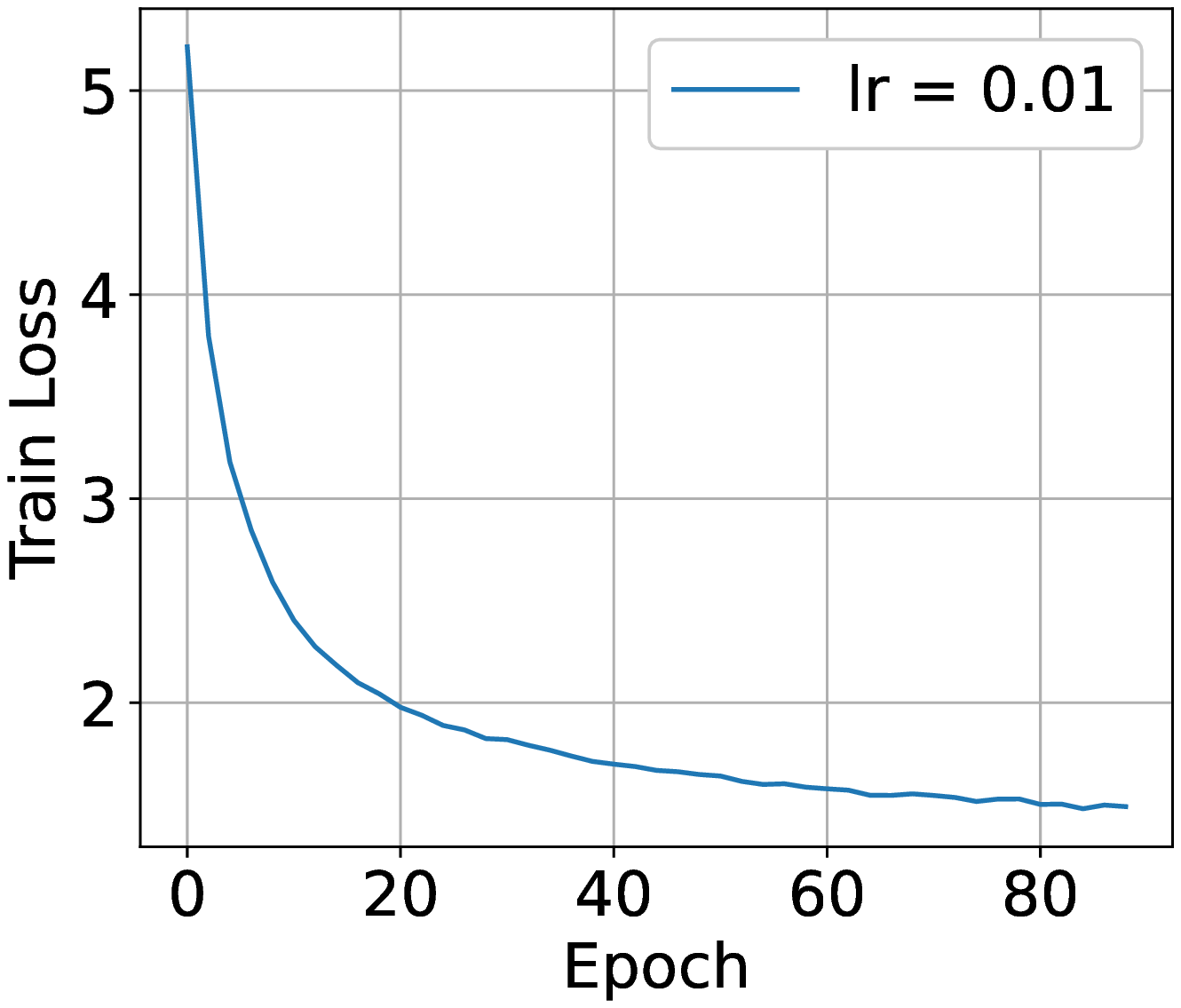}
	}
	{
		\includegraphics[scale=0.23]{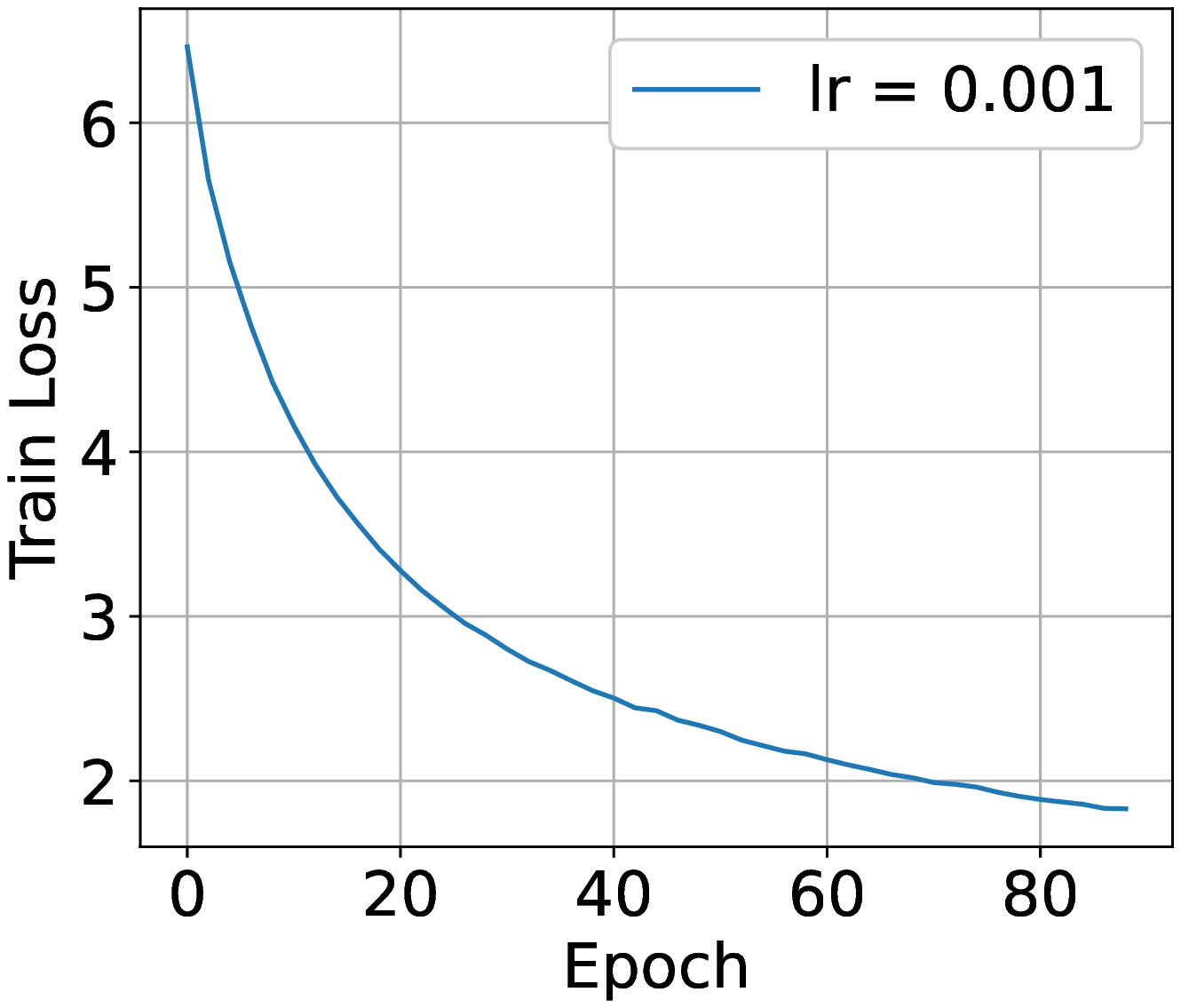}
	}
	{
		\includegraphics[scale=0.23]{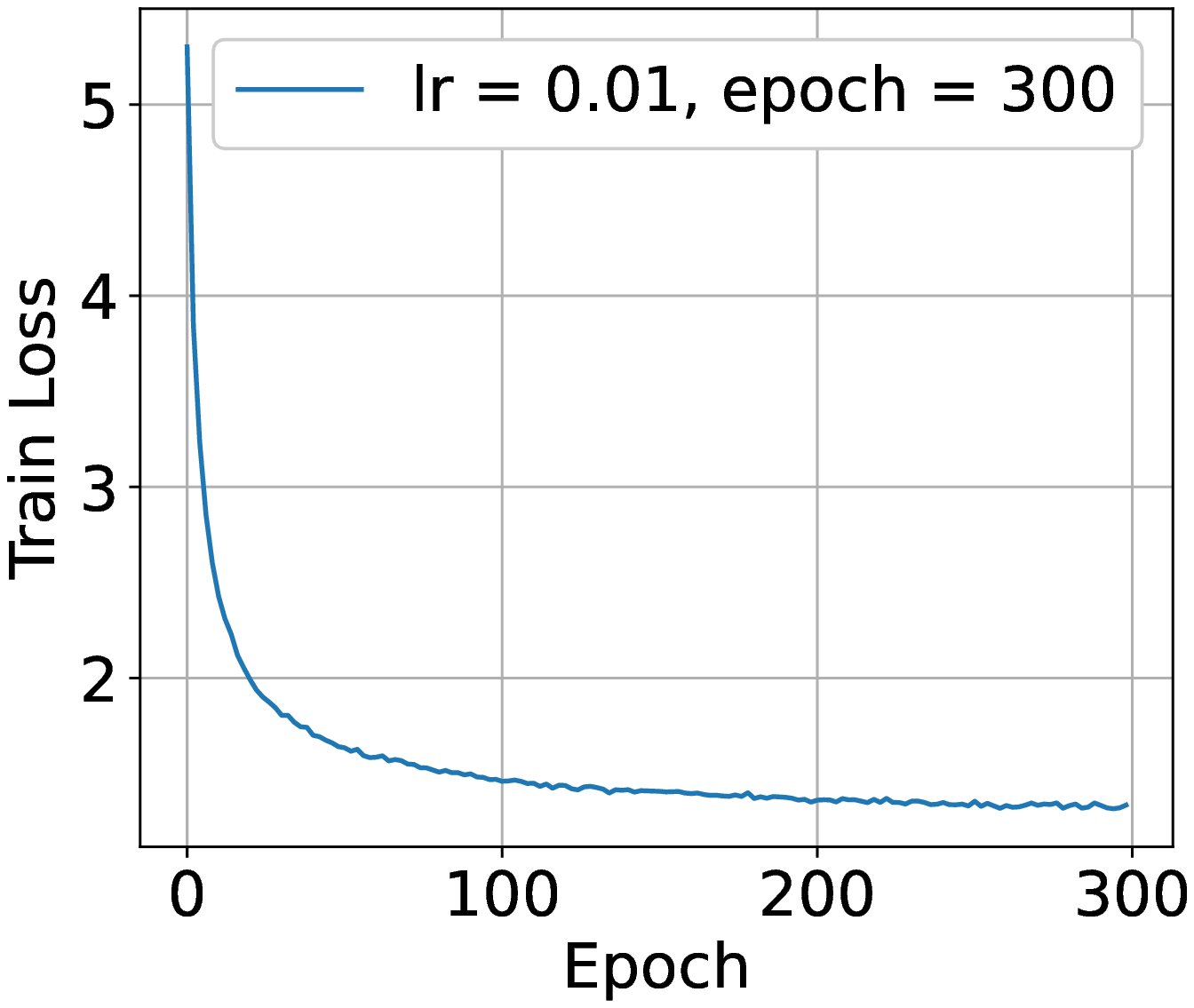}
	}

	{
		\includegraphics[scale=0.22]{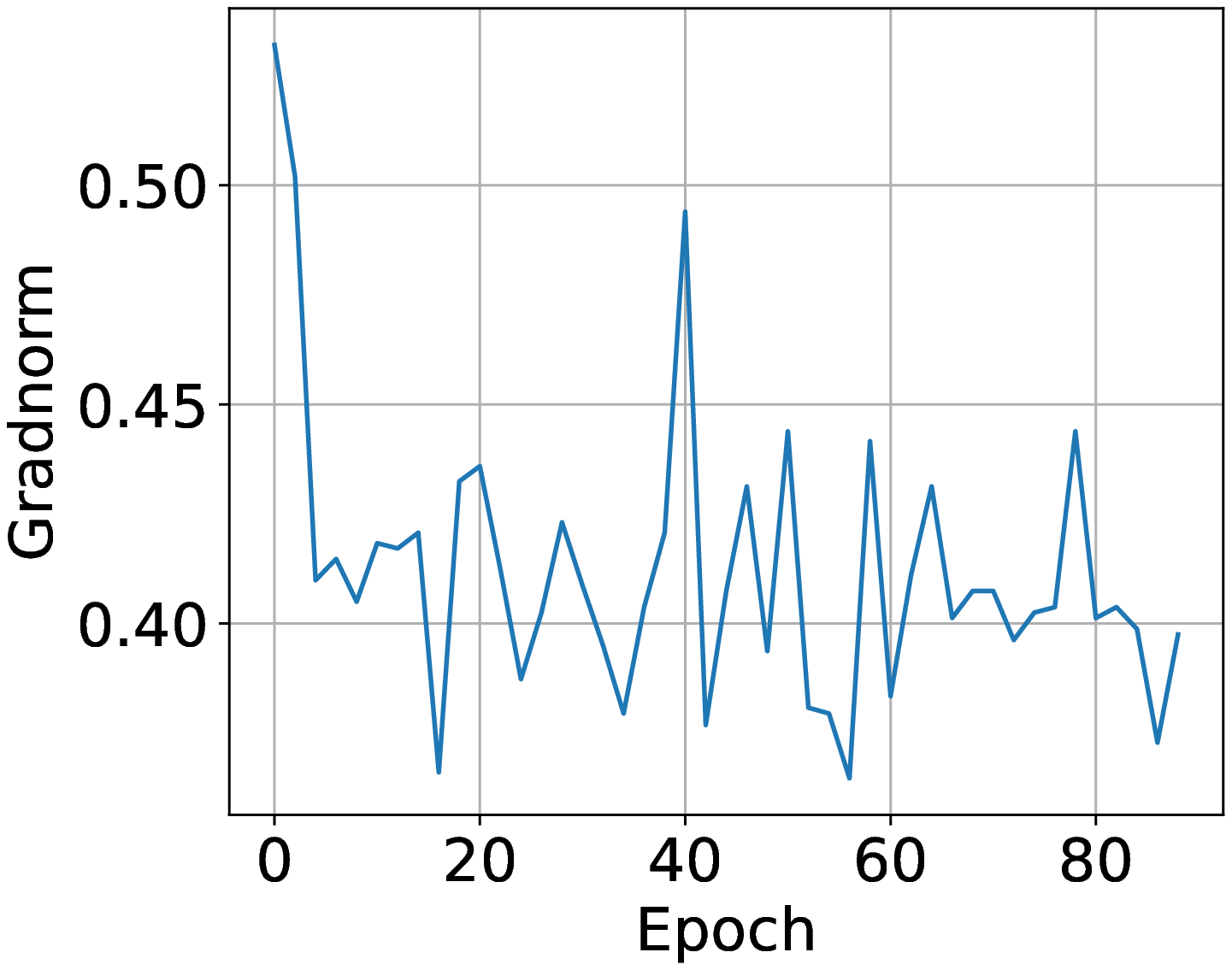}
	}
	{
		\includegraphics[scale=0.22]{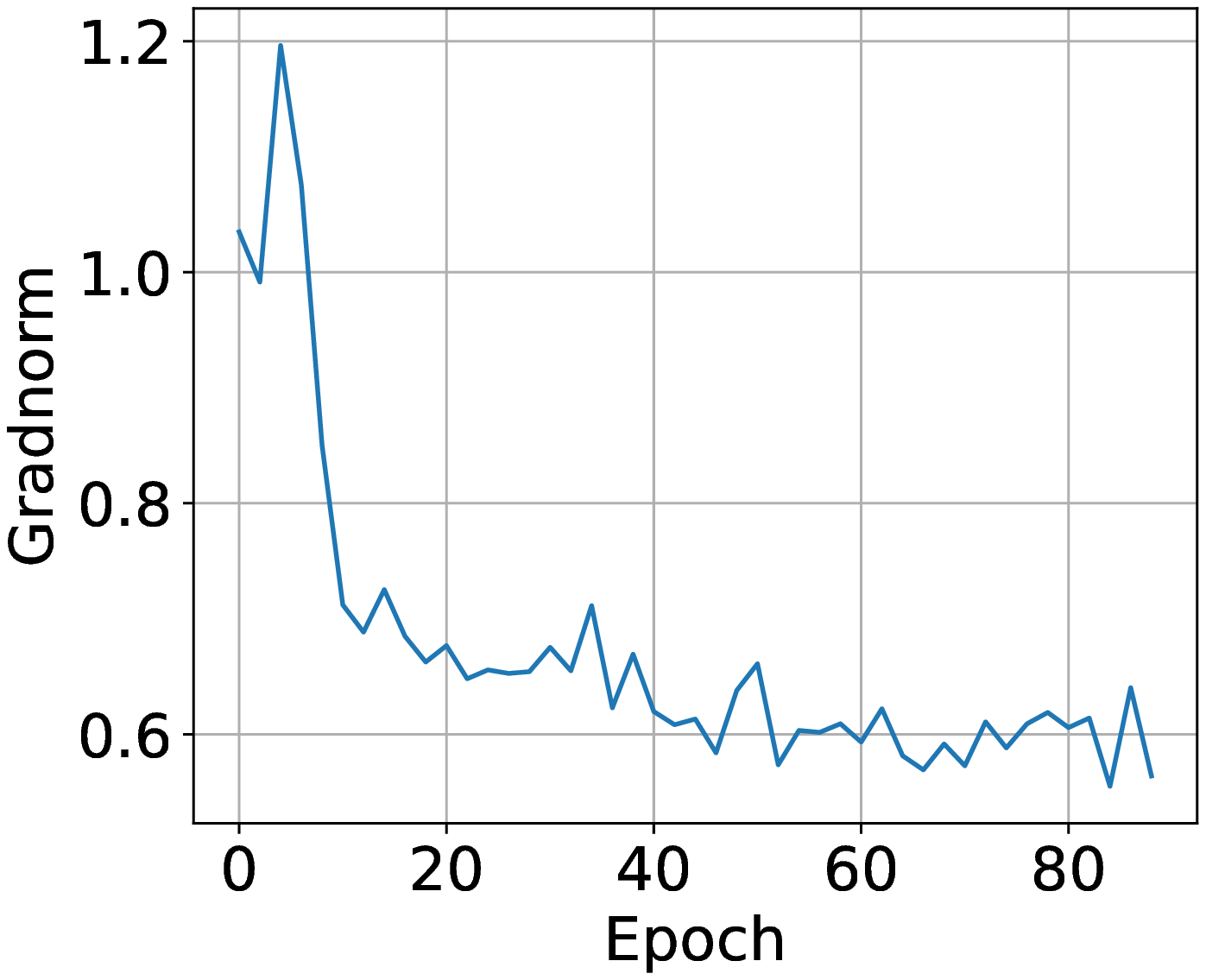}
	}
	{
		\includegraphics[scale=0.22]{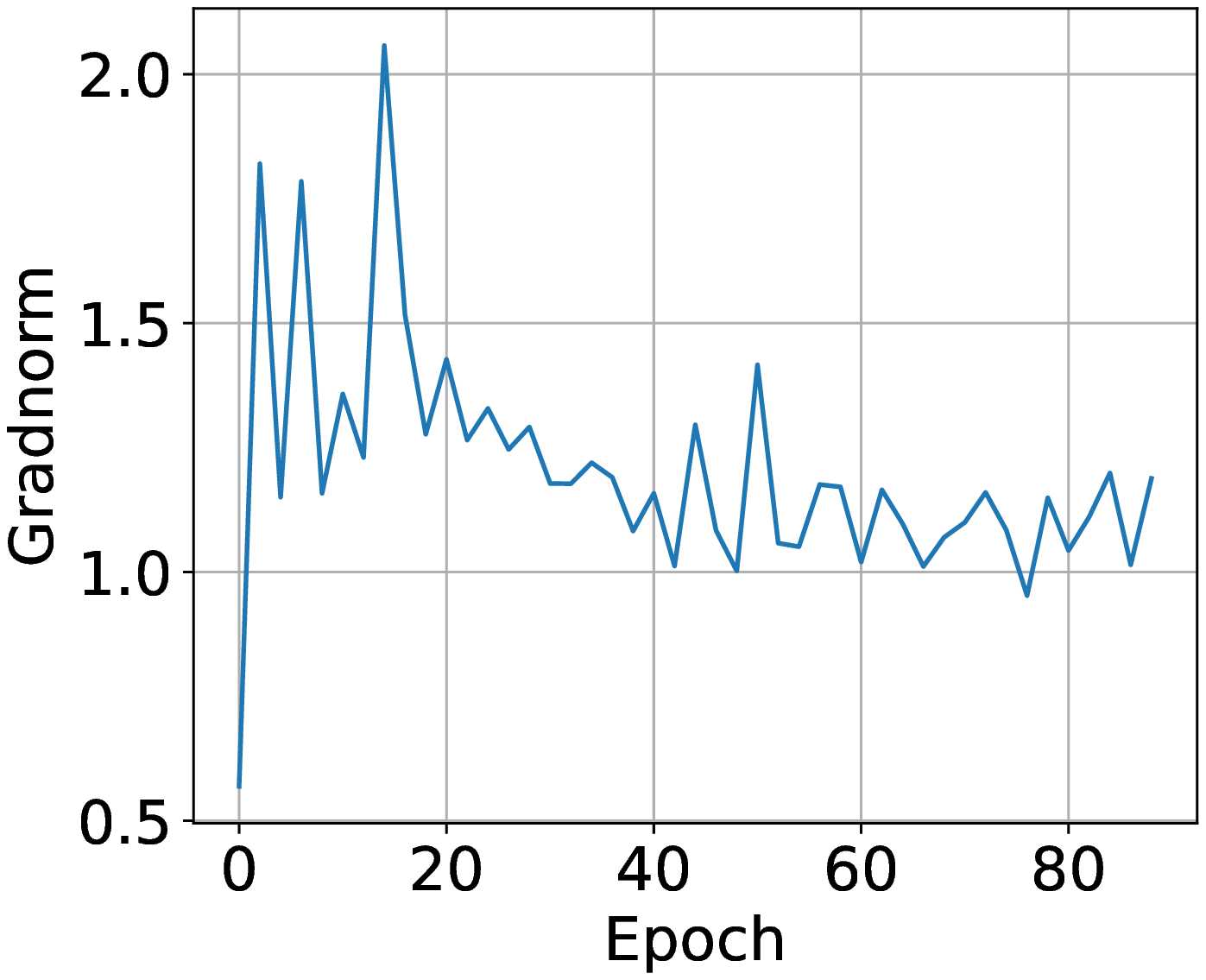}
	}
	{
		\includegraphics[scale=0.22]{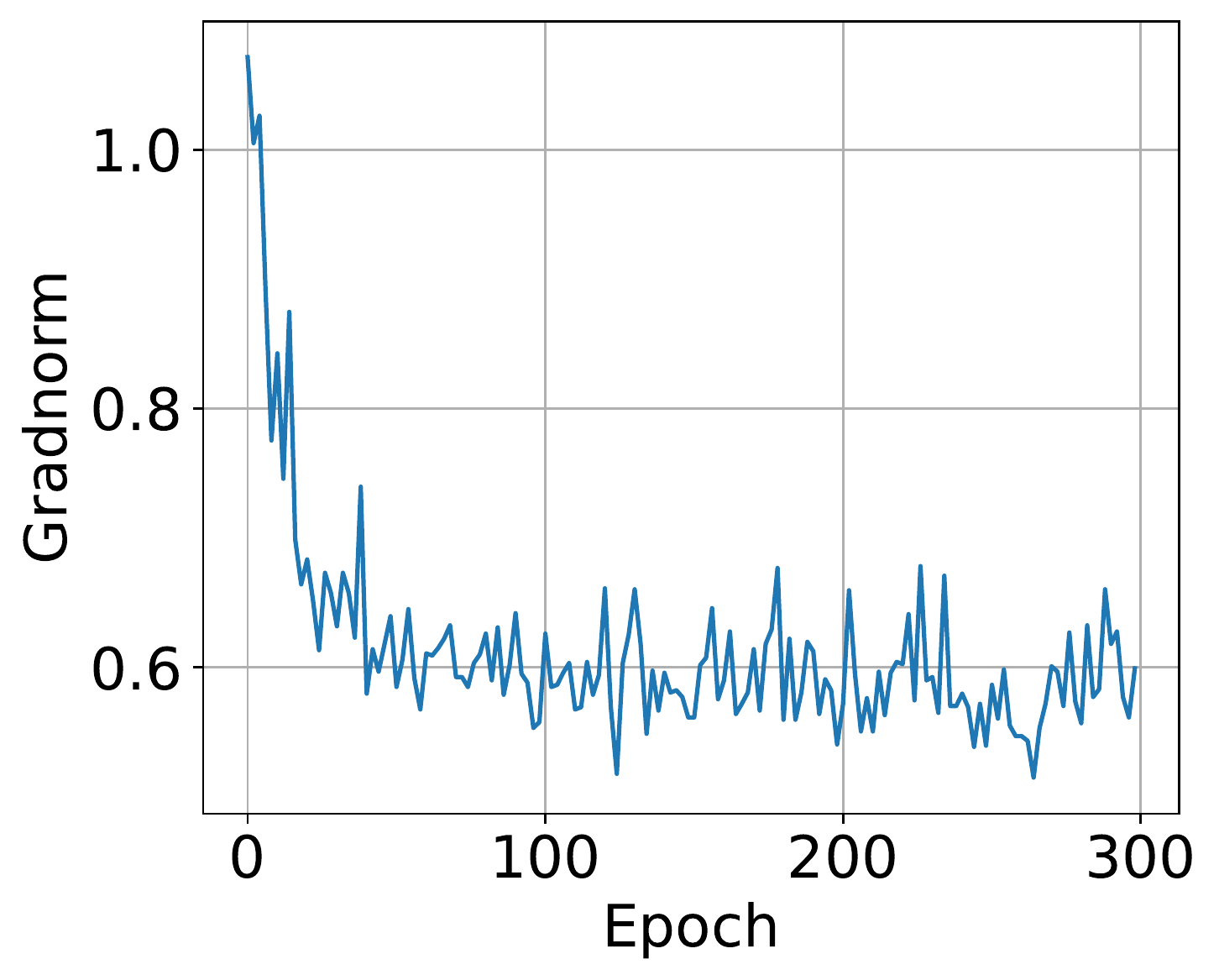}
	}

	{
		\includegraphics[scale=0.22]{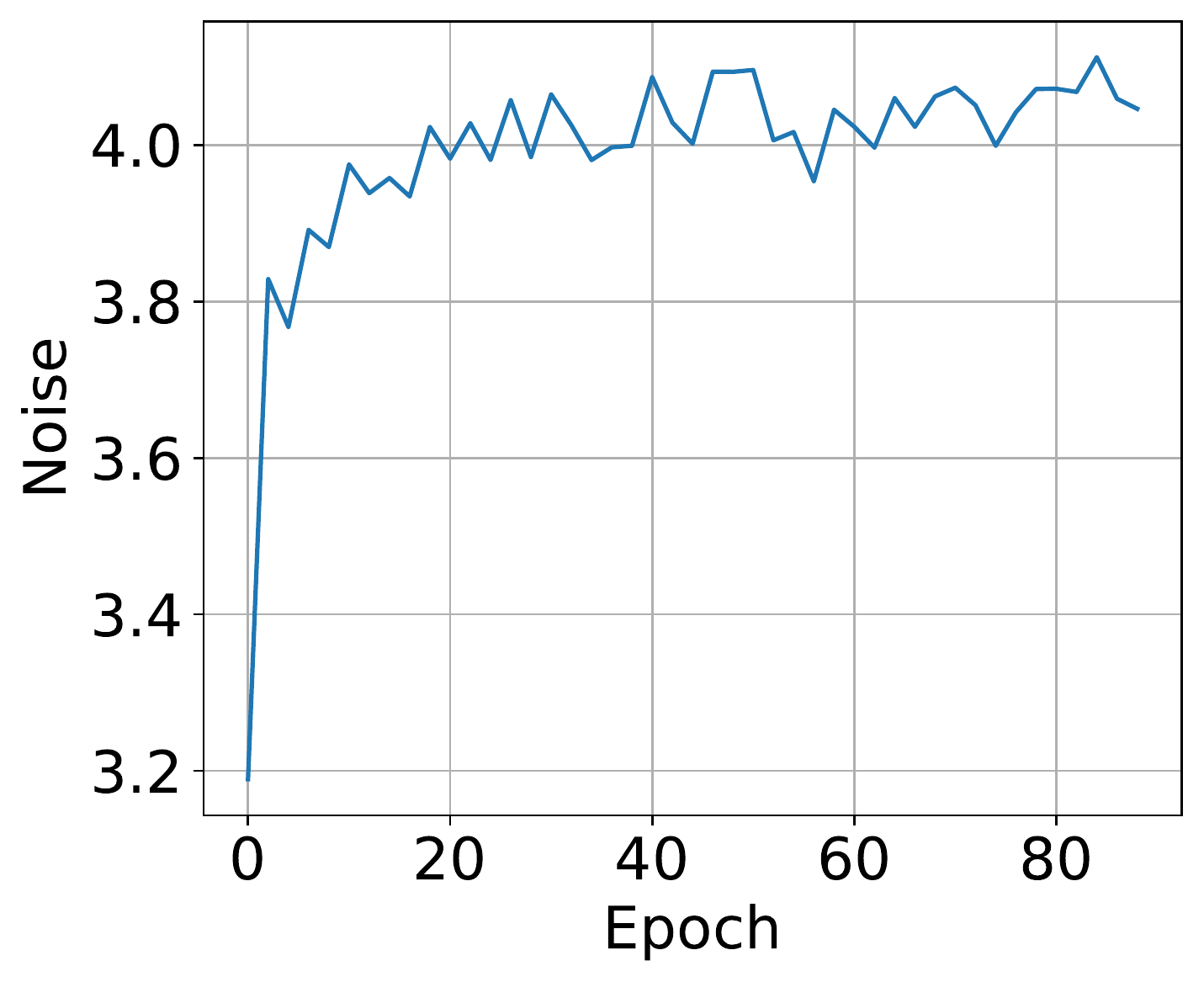}
	}
	{
		\includegraphics[scale=0.22]{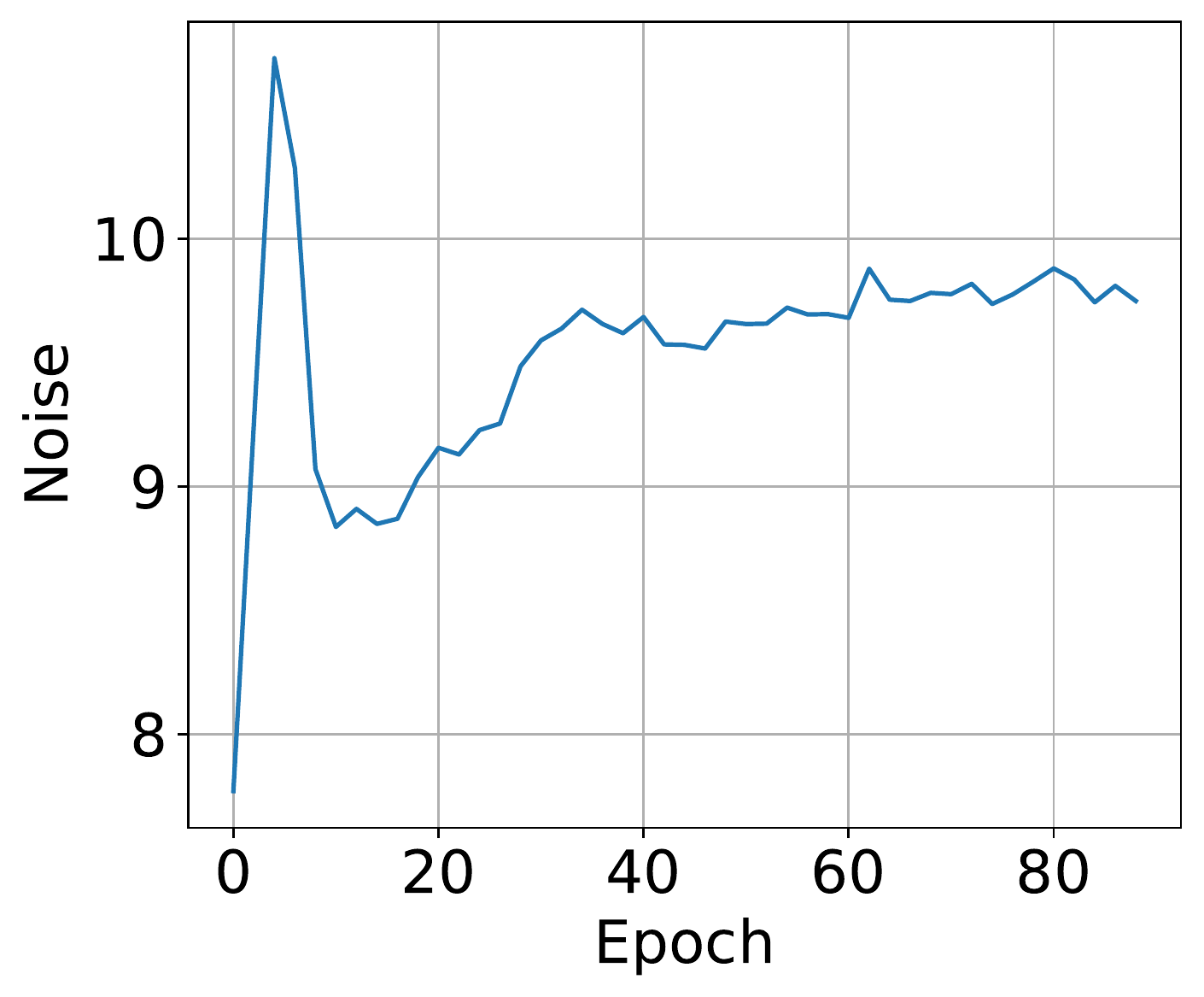}
	}
	{
		\includegraphics[scale=0.22]{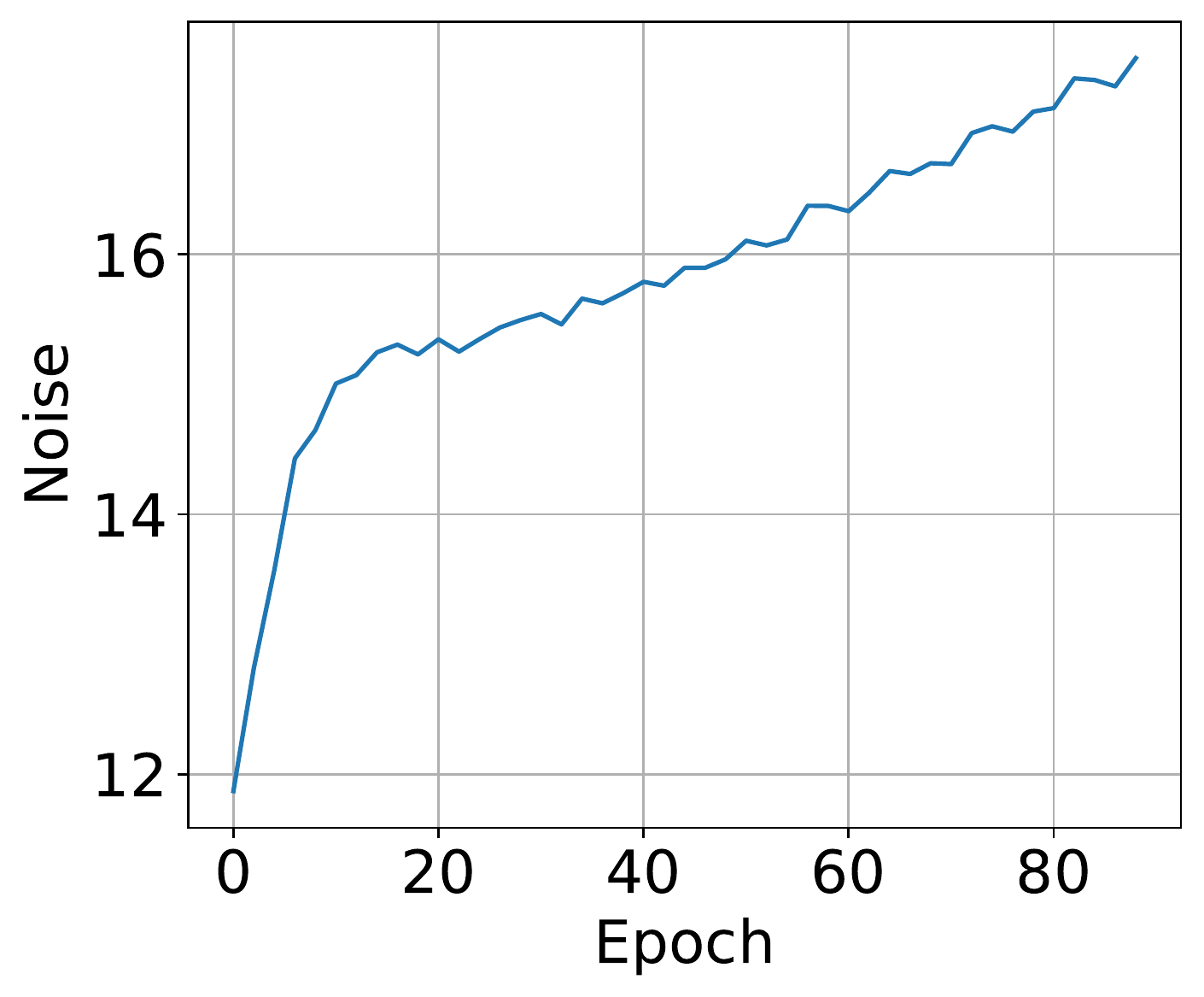}
	}
	{
		\includegraphics[scale=0.22]{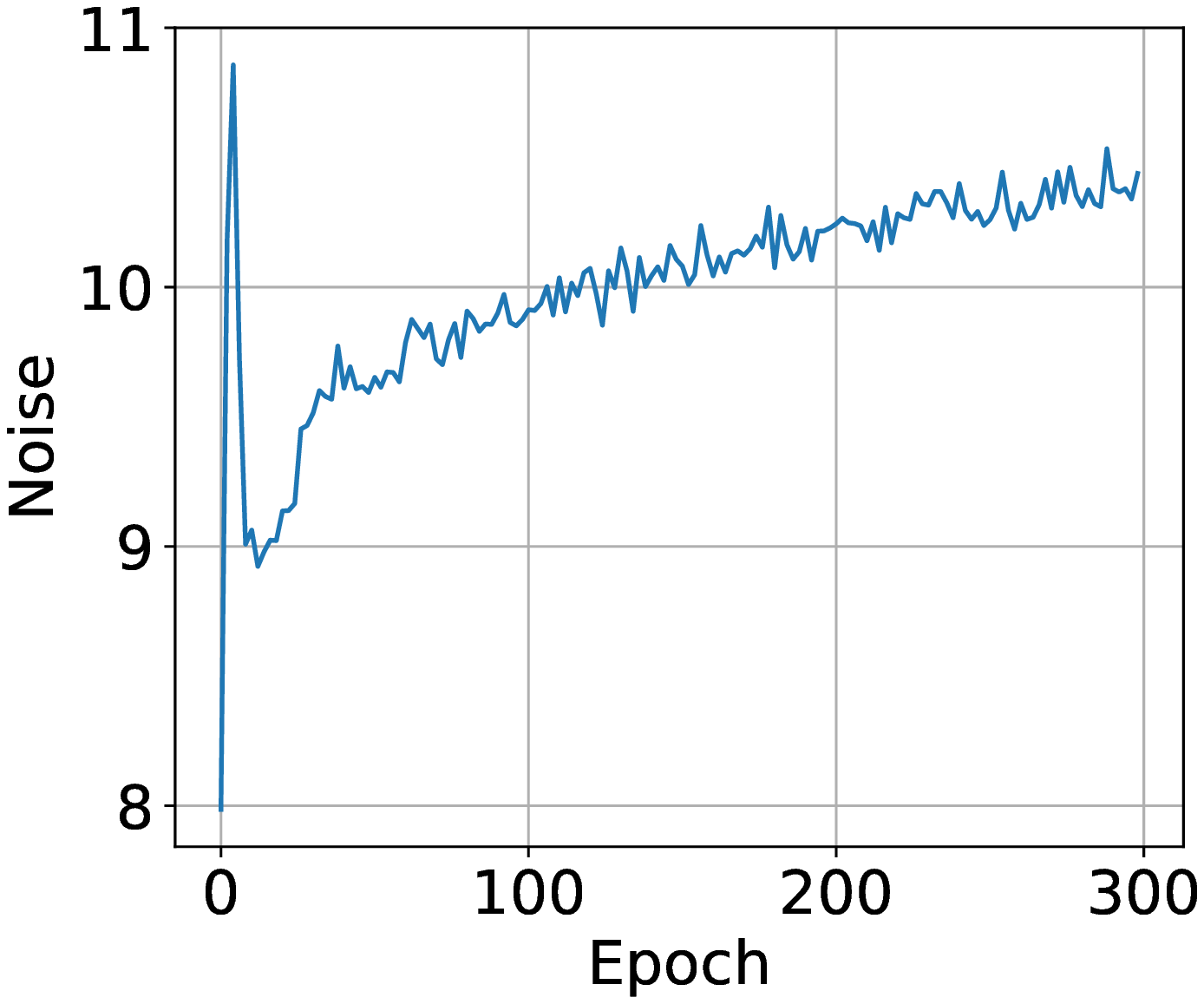}
	}

	\vspace{-0.3cm}
	\caption{
		The quantities of interest \eqref{eq:quantities} vs epoch for the constant learning rate training schedule in ImageNet experiments. The learning rate is set to be $0.1, 0.01, 0.001, 0.01$ respectively starting from the left column. All models are trained for 90 epochs, except that the last experiment in the column ran for 300 epochs
	}\vspace{-0.2cm}
	\label{fig:imagenet-constant}
\end{figure*}
\begin{figure*}[htbp]
	\centering
% 	\hrule
% 	\vspace{0.5cm}
	{
		\includegraphics[scale=0.23]{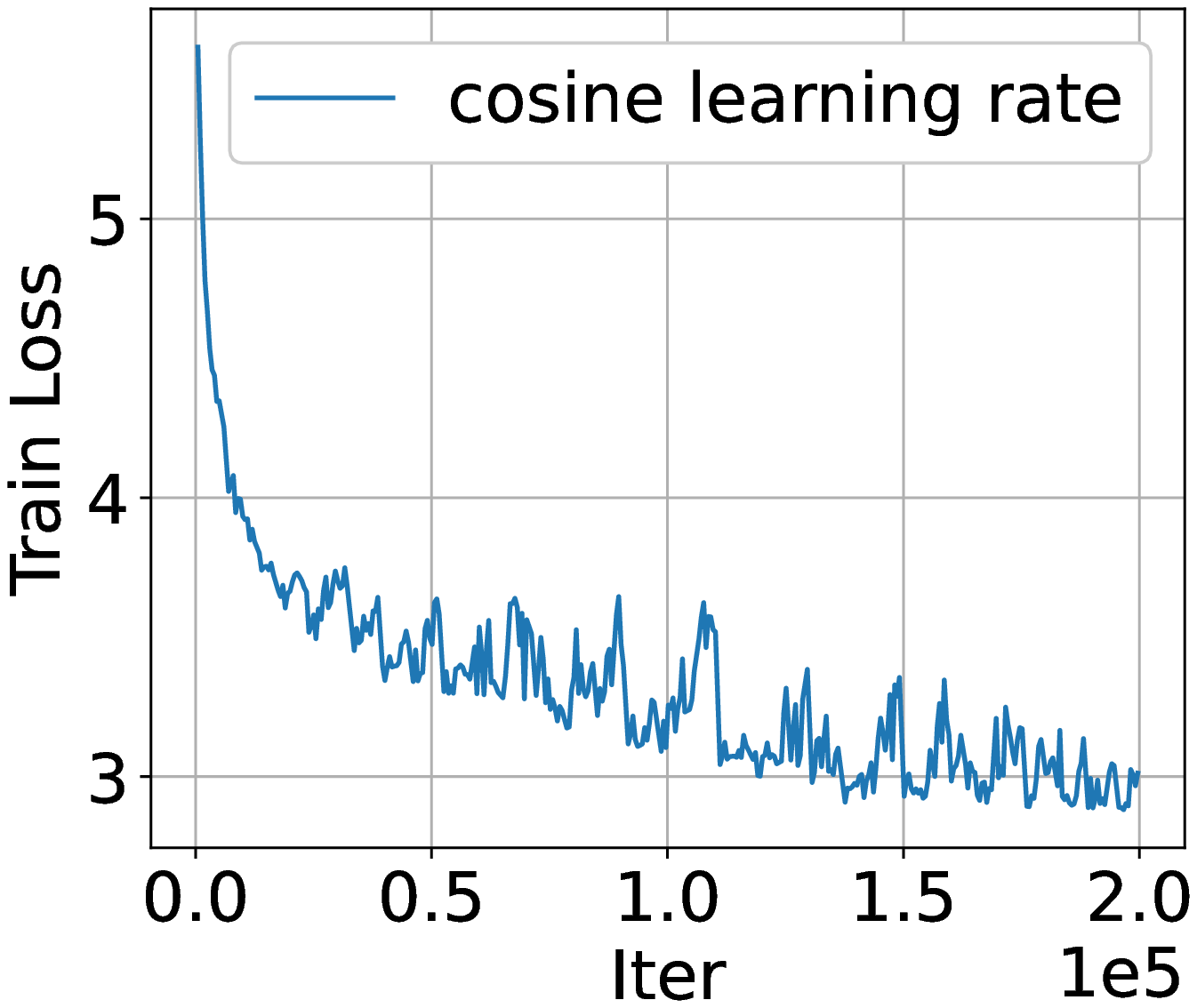}
	}	
	{
		\includegraphics[scale=0.23]{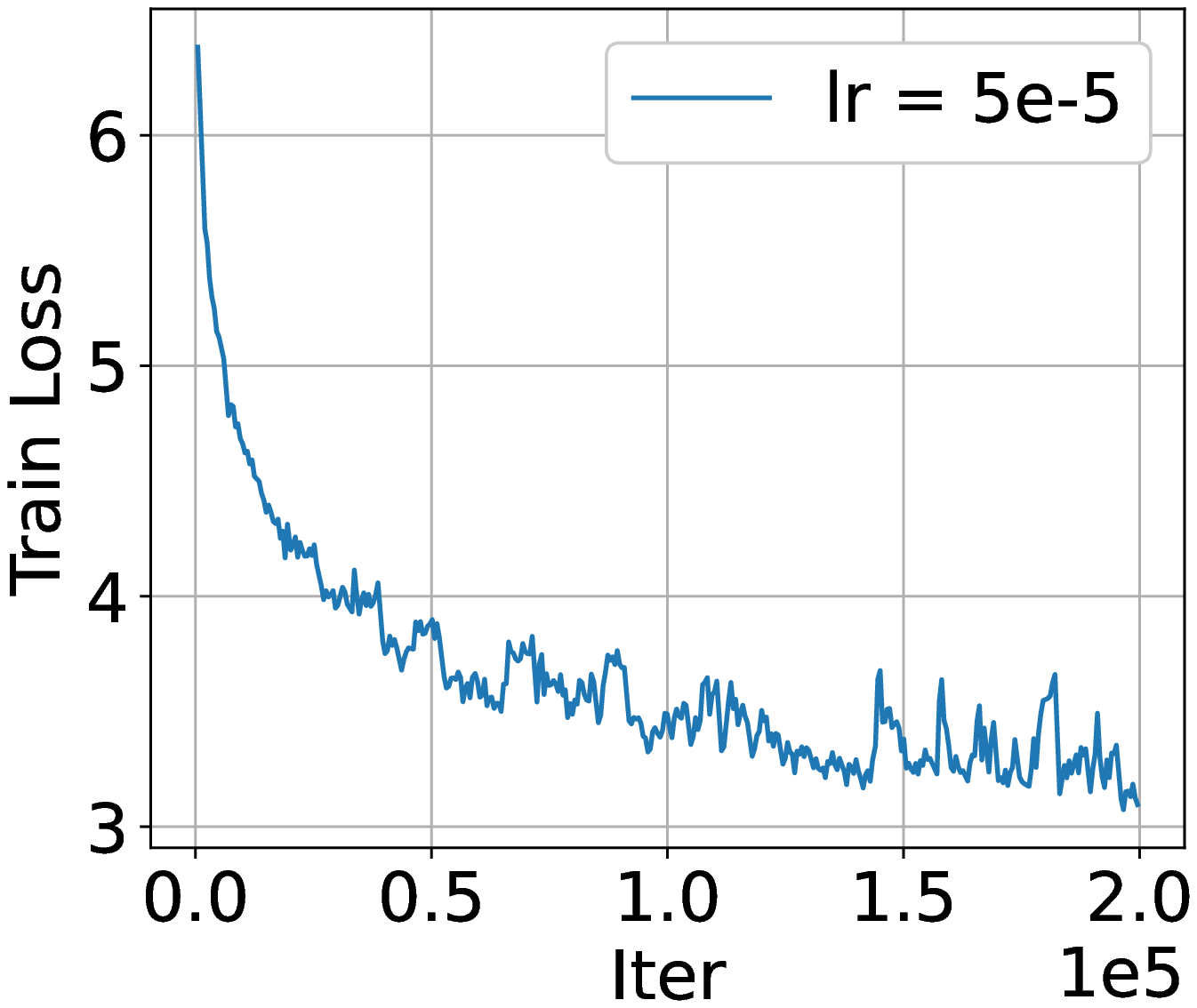}
	}	
	{
		\includegraphics[scale=0.23]{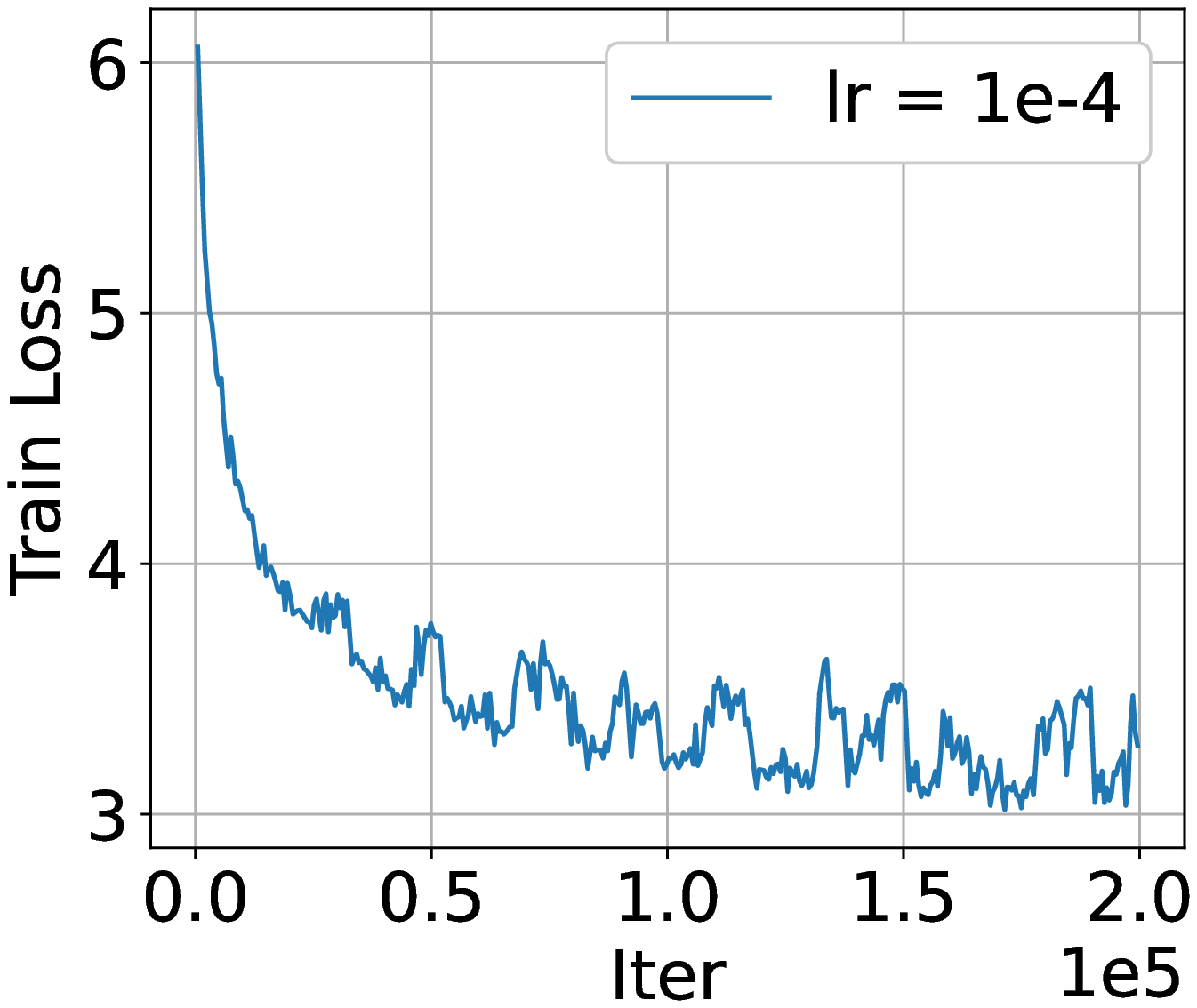}
	}	
	{
		\includegraphics[scale=0.23]{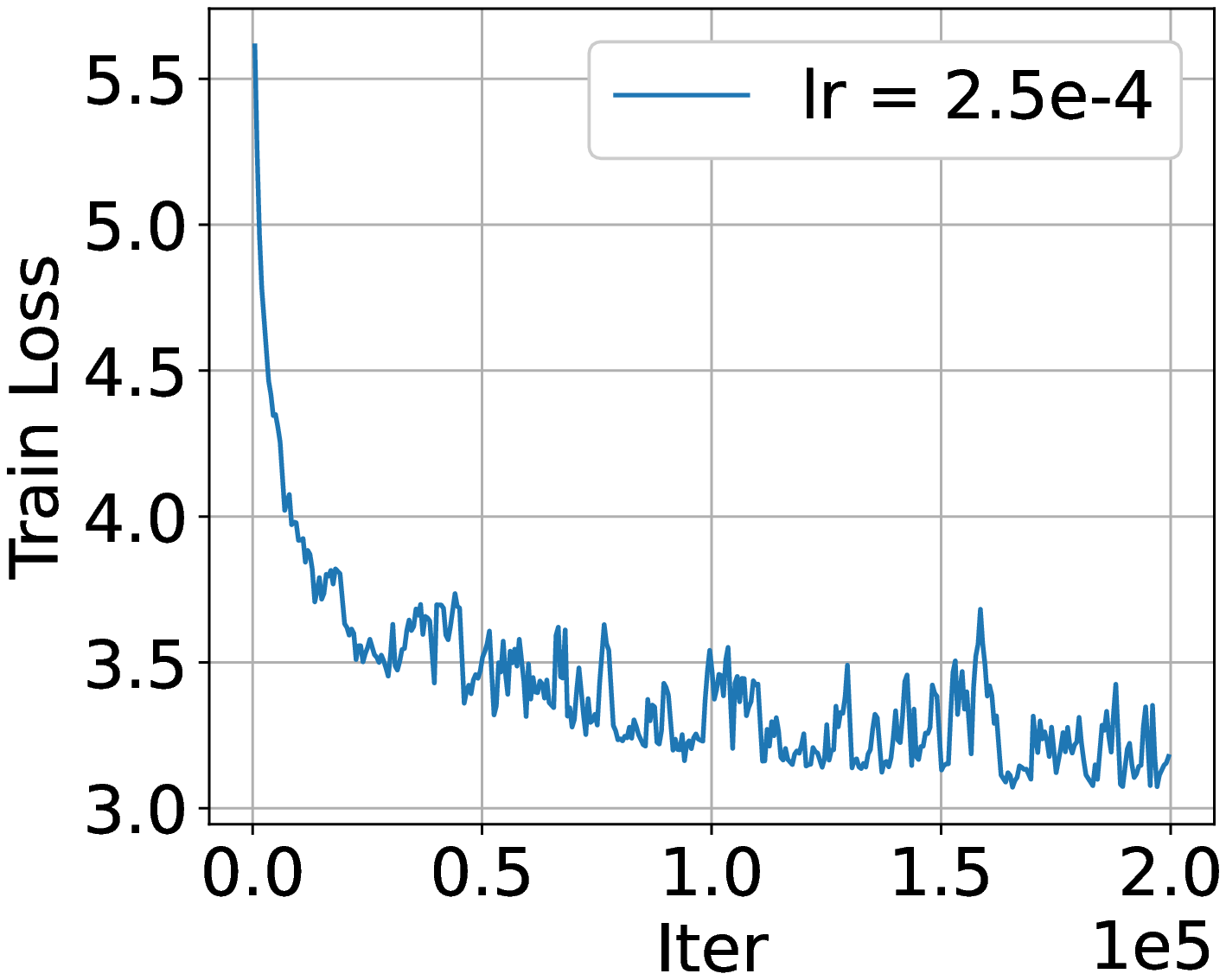}
	}

	{
		\includegraphics[scale=0.23]{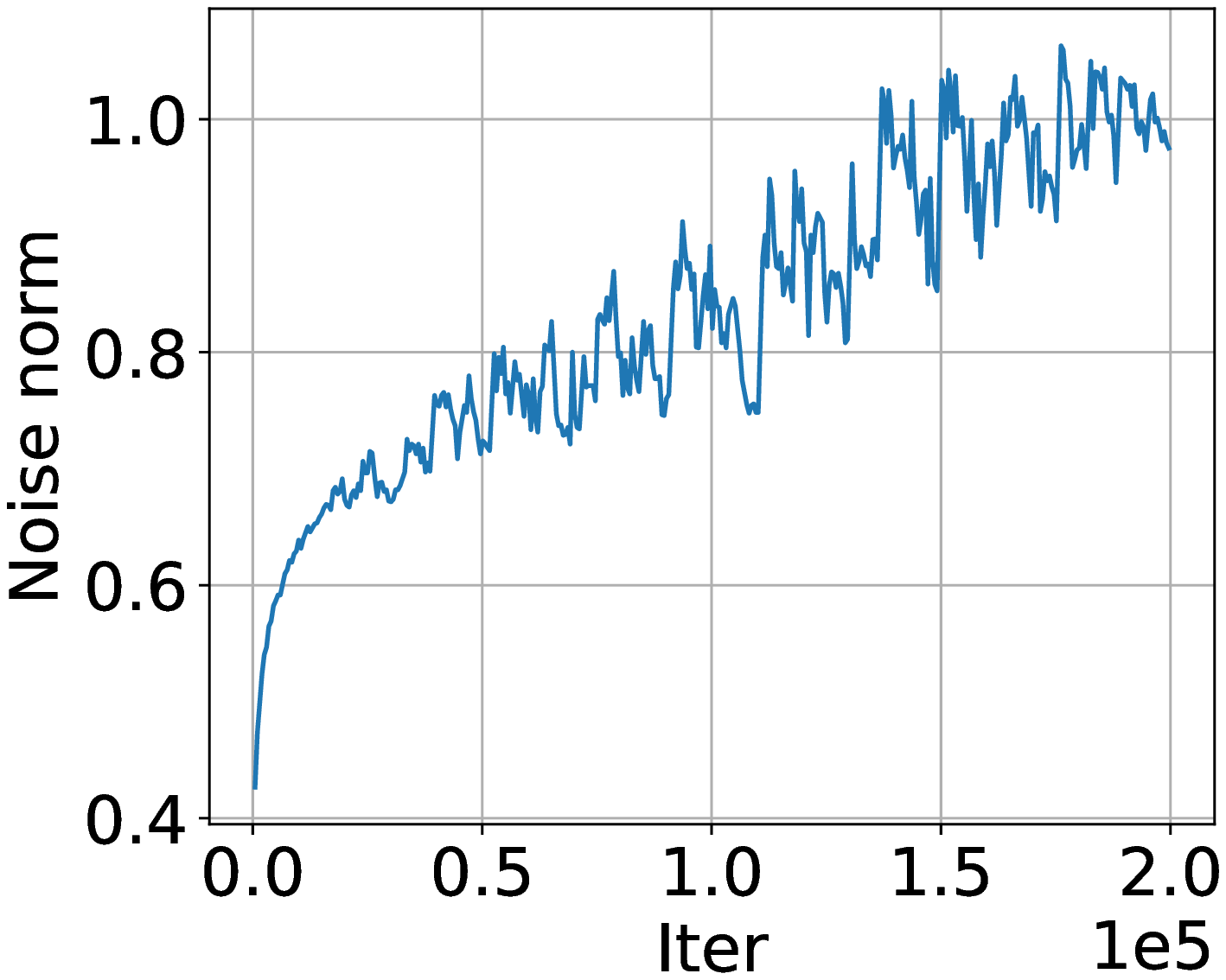}
	}
	{
		\includegraphics[scale=0.23]{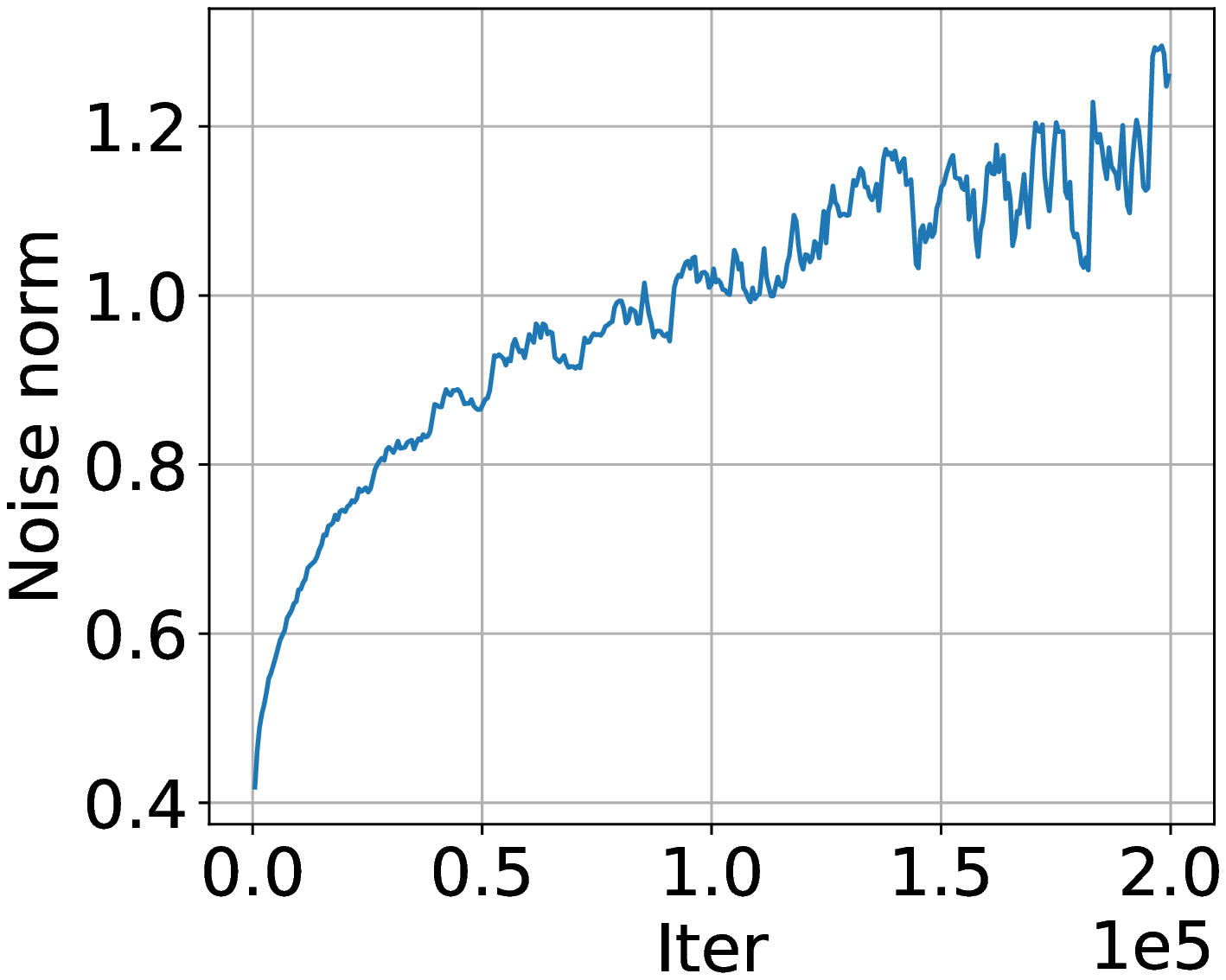}
	}
	{
		\includegraphics[scale=0.23]{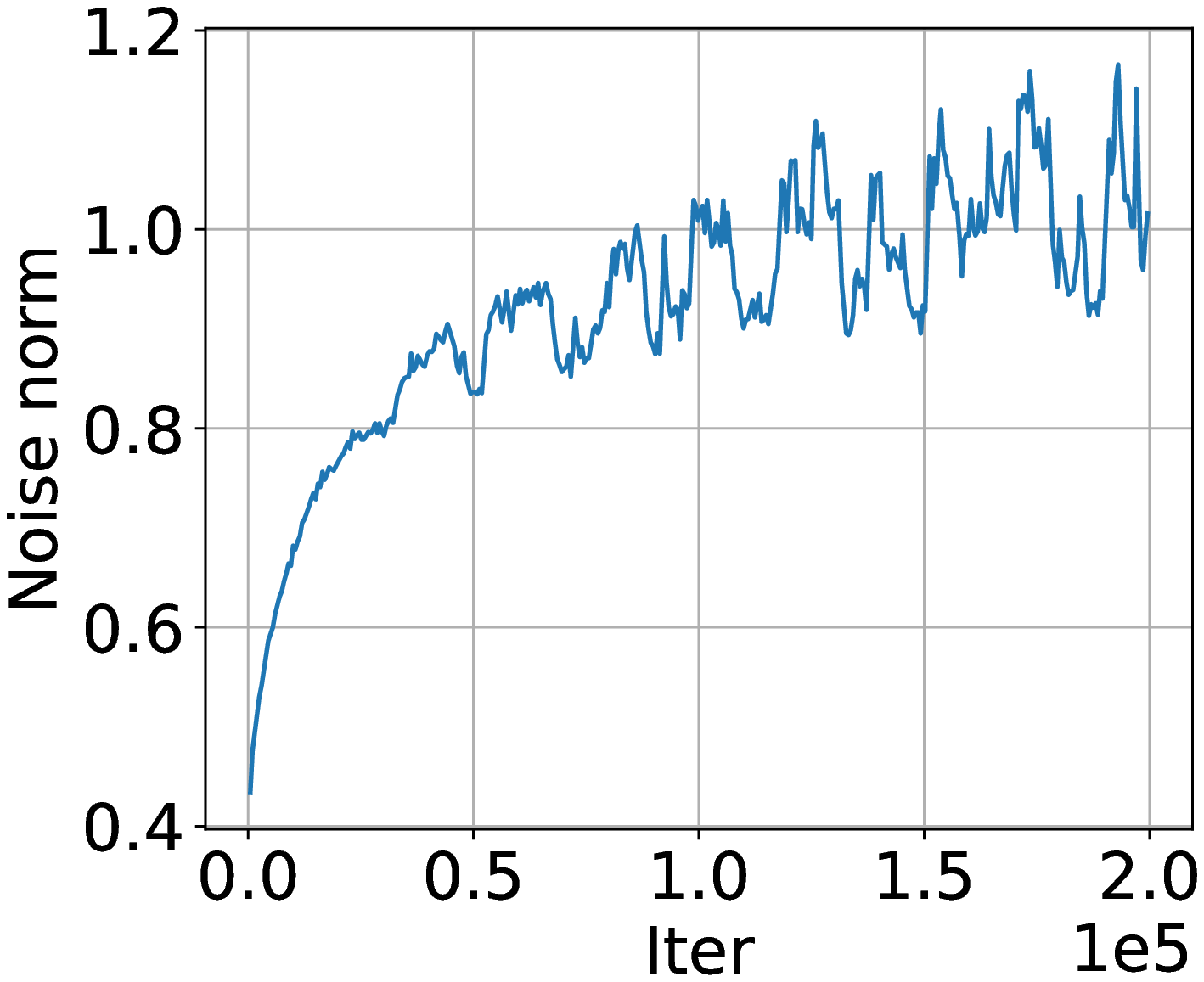}
	}
	{
		\includegraphics[scale=0.23]{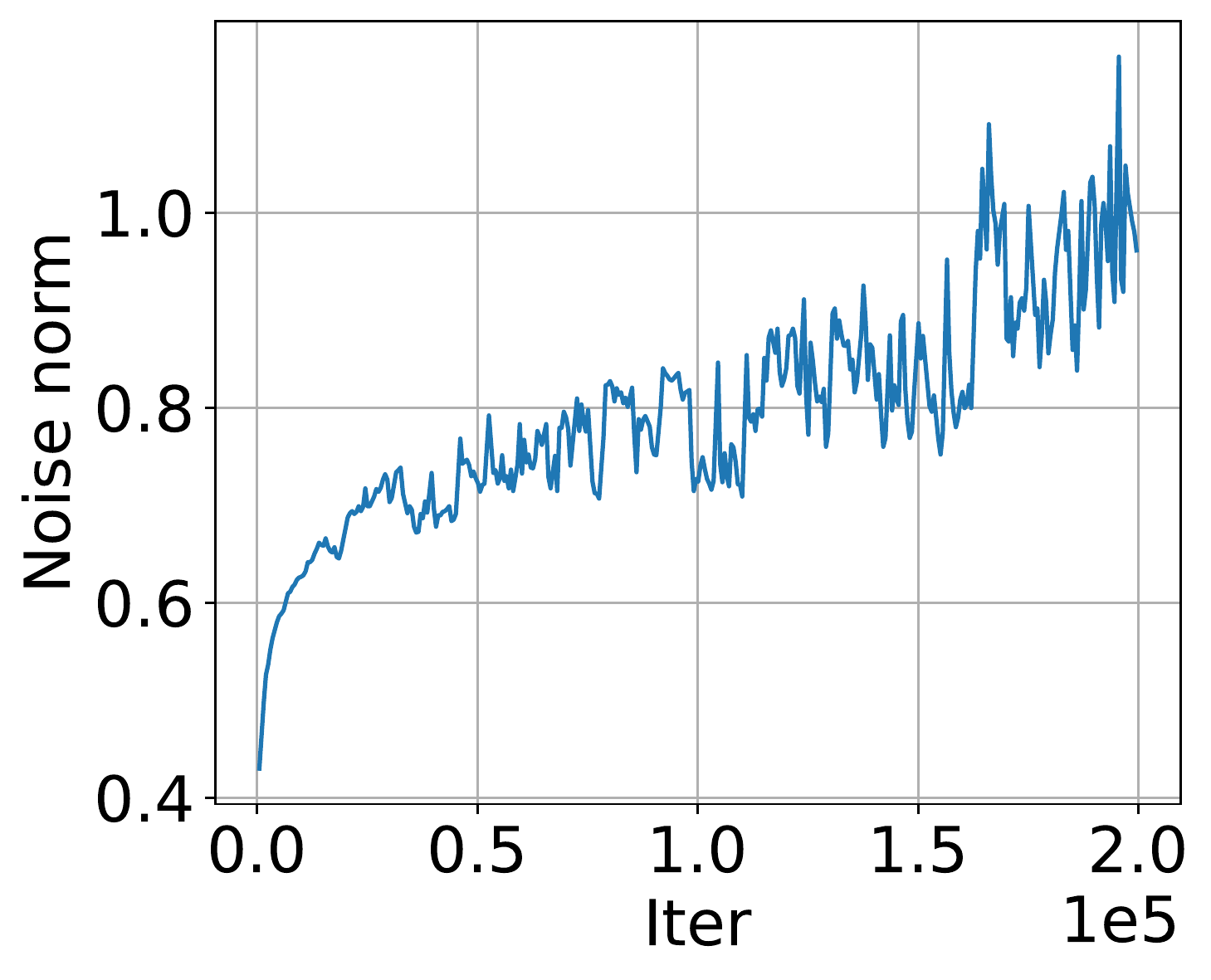}
	}

	{
		\includegraphics[scale=0.23]{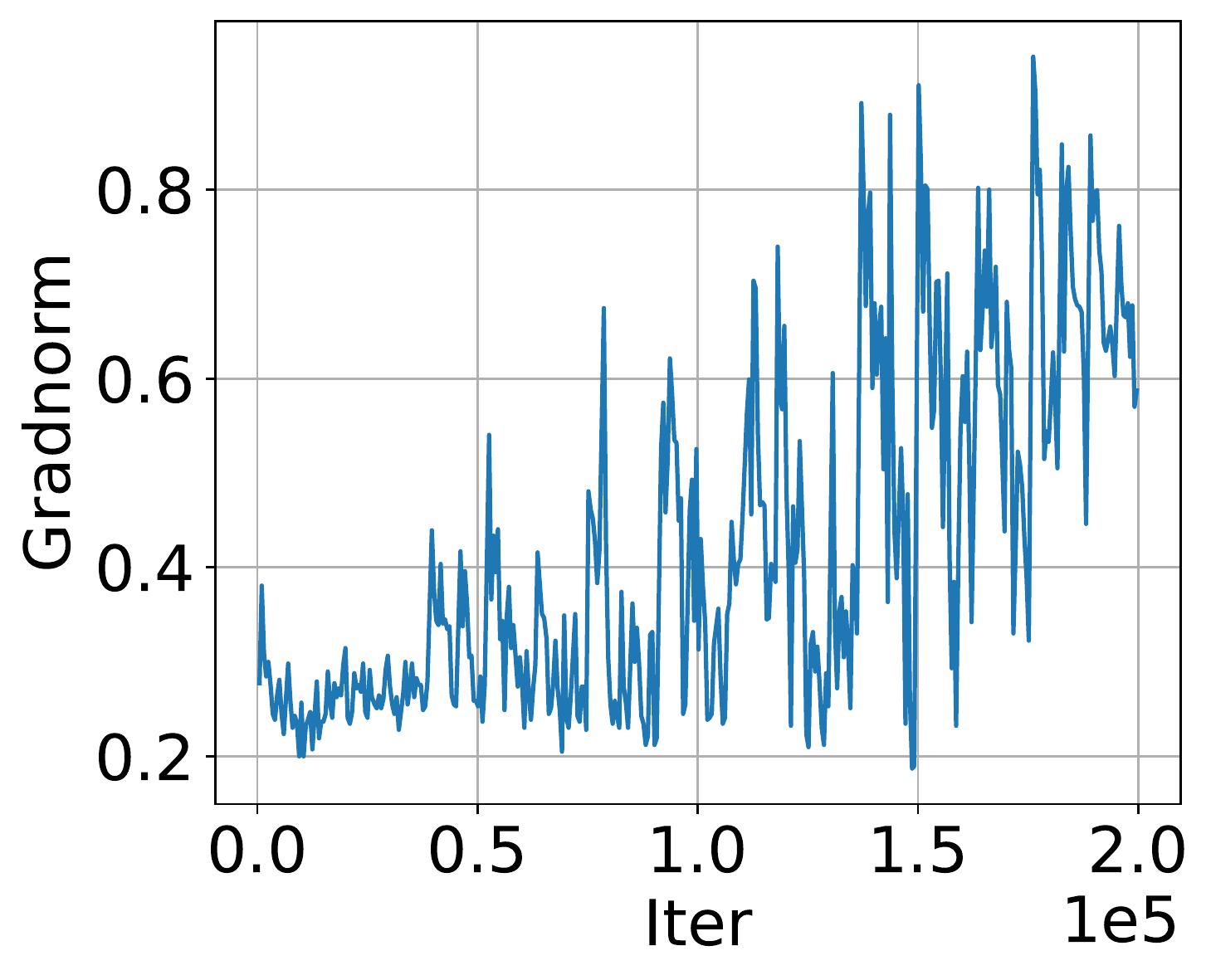}
	}
	{
		\includegraphics[scale=0.23]{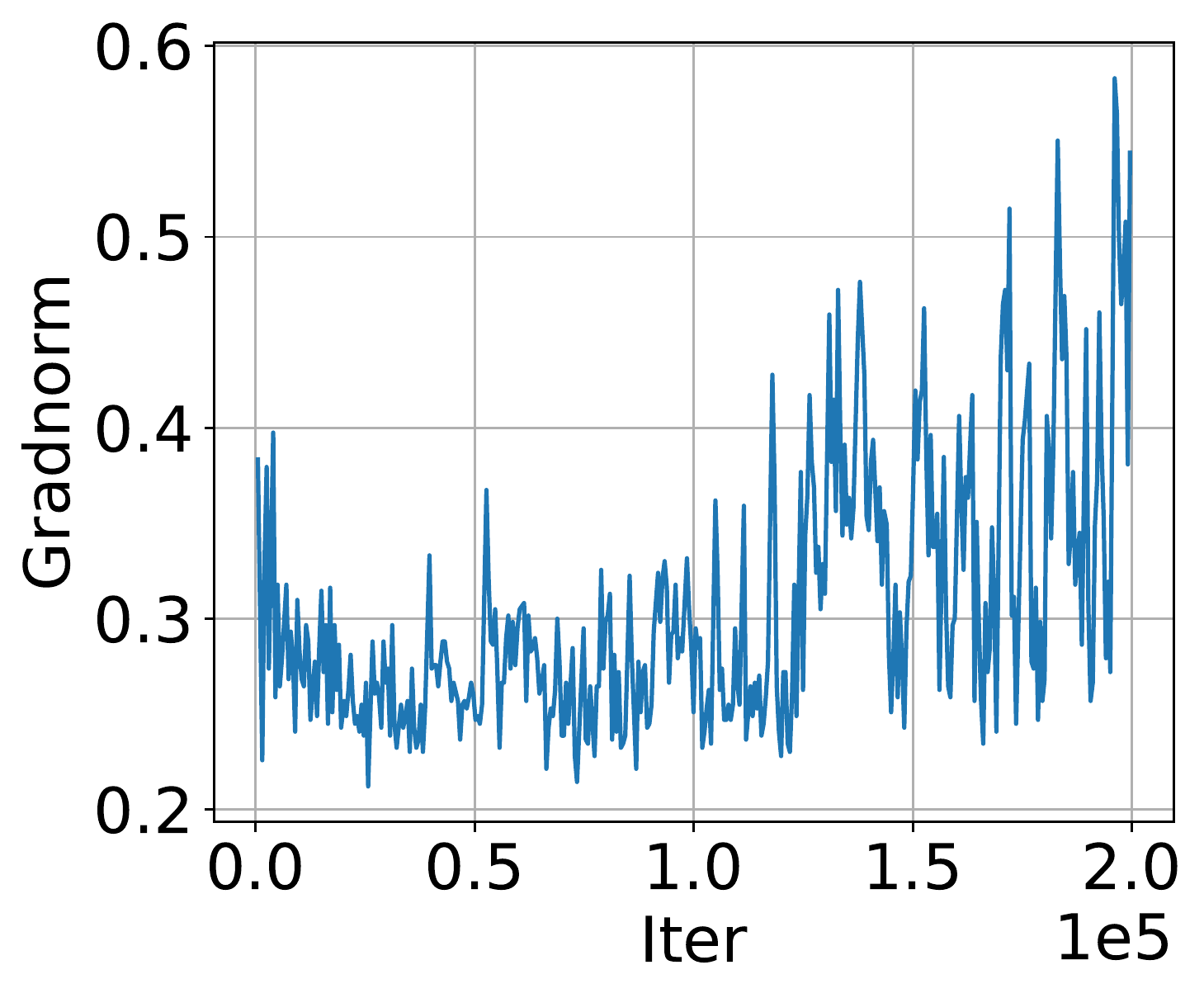}
	}
	{
		\includegraphics[scale=0.23]{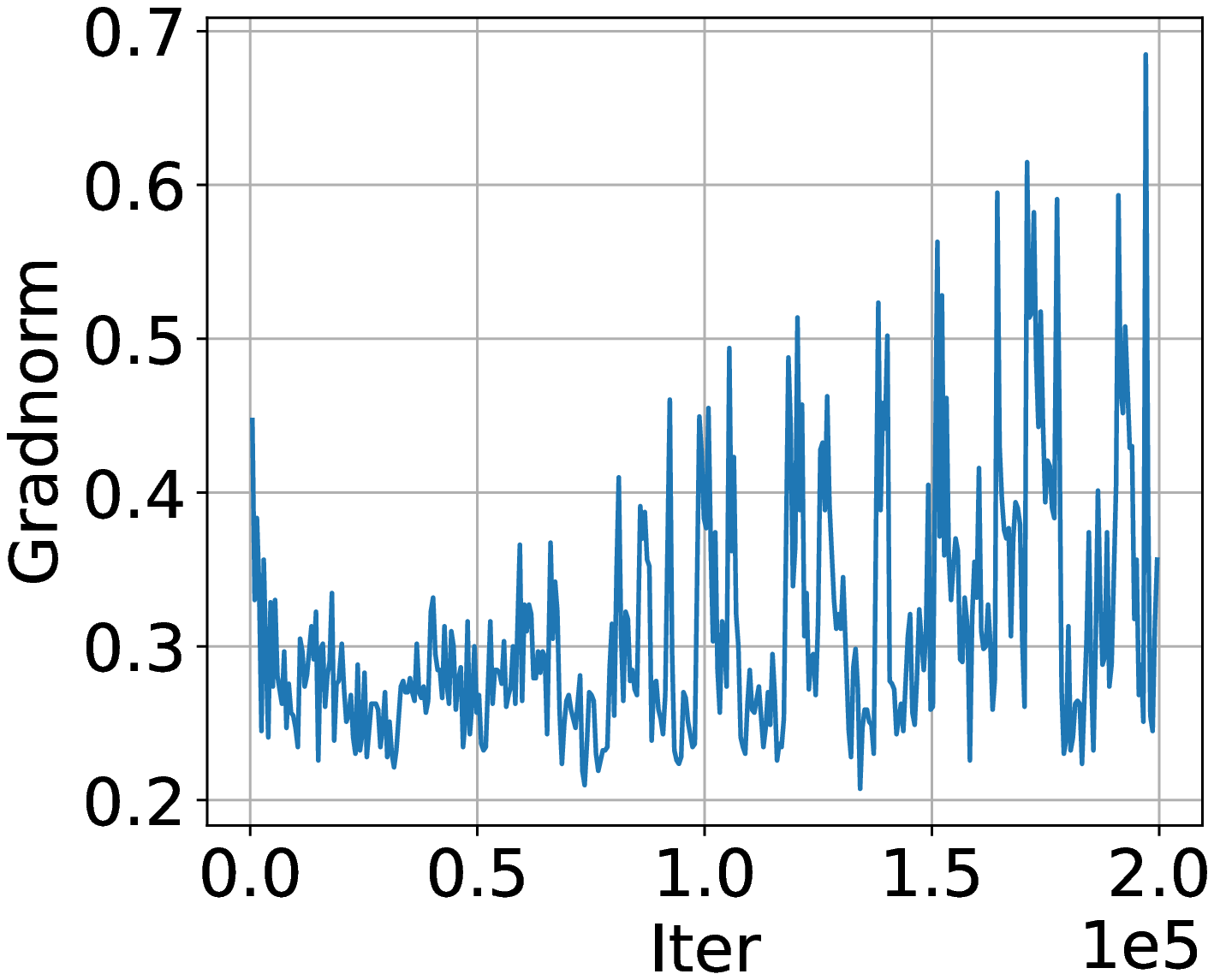}
	}
	{
		\includegraphics[scale=0.23]{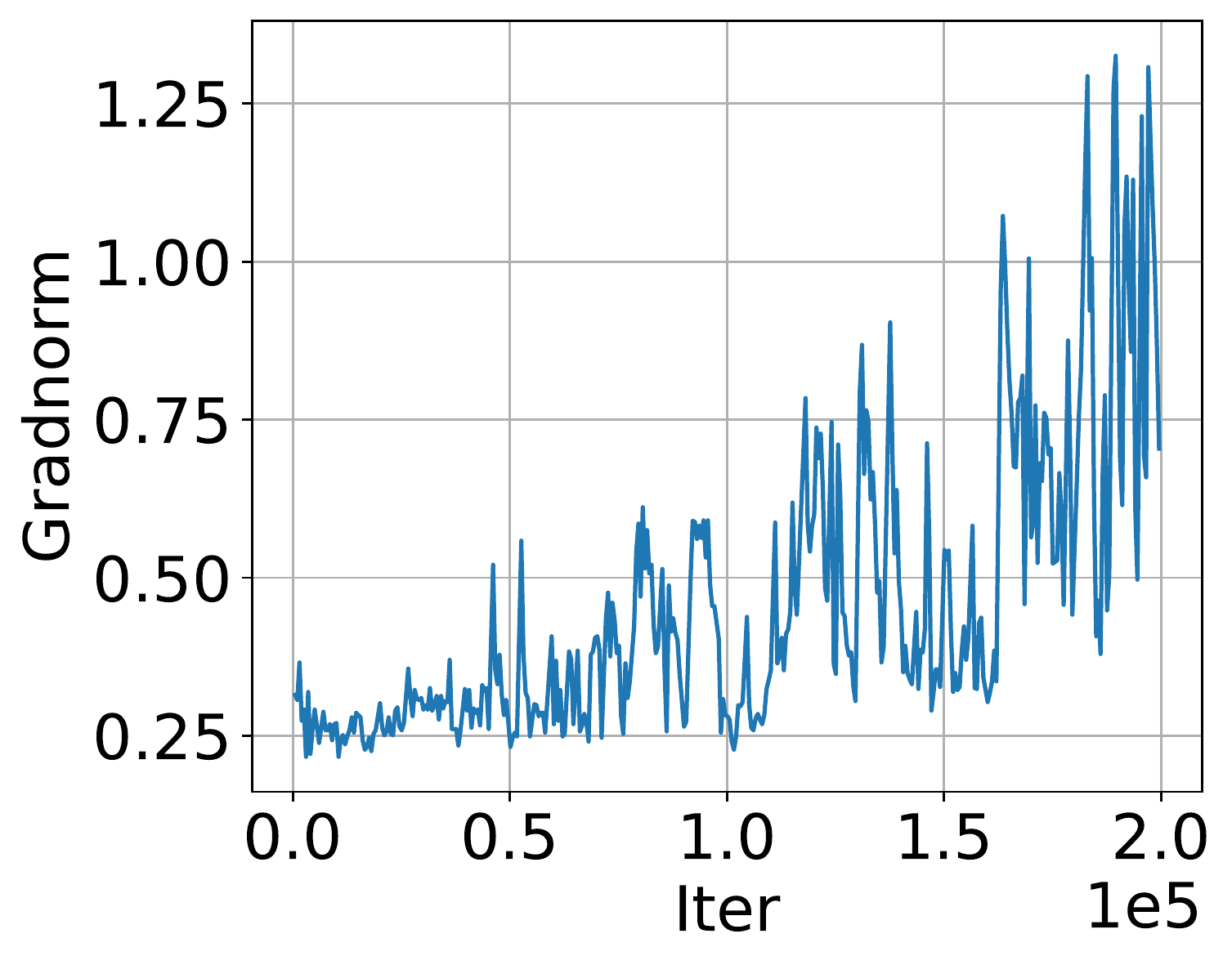}
	}

	\vspace{-0.5cm}
	\caption{
		The estimated stats vs epoch for the transformer XL training. The learning rate is set to be cosine learning rate with  $\eta = 0.00025$ in the first column. The learning rates are constant learning rates with $\eta = 0.00005, 0.0001, 0.00025$ from the second to the last column. 
	} 
	\label{fig:transformer-stats}
\end{figure*}

\vspace*{-5pt}
\section{A motivating example: ImageNet + ResNet}\label{sec:example}

We start our exposition by providing some experimental result showing that the traditional notion of convergence for nonconvex functions \emph{does not} occur in deep neural network training. Our experiments are based on one of the most popular training schemes, where we train ResNet101 on ImageNet. More details can be found later in Appendix~\ref{sec:exp} and in our released \href{https://github.com/JingzhaoZhang/NN-training-converge-to-nonstationary-points}{code repository}. 

To explain the quantities of interest, we first define our notation. Let $S=\{(x^i,y^i)\}^N_{i=1}$ be the dataset. Let $f(x, \theta)$ to denote the neural network function with model parameters $\theta$ and  input data $x$. We use  $\ell$ to denote the loss function such as cross-entropy after softmax. We would like to investigate, during training, the evolution of the following quantities: training loss, gradient norm, and noise. Mathematically, they are defined as follows respectively,
\begin{align}\label{eq:quantities}
	&\L_S(\theta_k) := \tfrac{1}{N}\textstyle\sum_{i=1}^N \ell(f(x^i,\theta_k),y^i),\\
	& \norm{\grad L_S(\theta_k)}_2 :=  \|\tfrac{1}{N}\textstyle\sum_{i=1}^N \frac{\partial}{\partial \theta} \ell(f(x^i,\theta_k),y^i) \|_2, \nonumber\\
	&\sigma(\theta_k) := \sqrt{ \tfrac{1}{N} \textstyle\sum_{i=1}^N \norm{\grad L_S(\theta_k)  - \frac{\partial}{\partial \theta} \ell(f(x^i,\theta_k),y^i)  }_2^2 },\nonumber 
\end{align}
where $\norm{\cdot}_2$ is the standard vector $\ell_2$ norm.

We adopt the standard training schedule following~\citep{He2016}, i.e., the learning rate starts at $0.1$ and is decayed by a factor of 10 every 30 epochs. The evolution of the aforementioned quantities is plotted in Figure~\ref{fig:imagenet-baseline}. We make the following immediate observations:
\begin{itemize}[leftmargin=*,itemsep=0.05cm]
	\item Within each period where the step size is held constant, the change in loss converges to 0.
	\item The gradient norm does not converge to $0$ despite the fact that the loss function converges. In fact, the gradient norm stays roughly unchanged.
	\item The noise level (in the stochastic gradient) increases during training.
\end{itemize}

The above observations suggest that there is \emph{ a tremendous gap between theory and practice}. Much of the research on nonconvex optimization theory focuses on the convergence rate of gradient norms under a bounded-smoothness, bounded-noise setup. Faster algorithms are designed under this guidance. However, in practice, we find that the convergence of the training loss \emph{does not} require the convergence of gradient norms. This mismatch may be the reason why techniques such as variance reduction or local regularization combined with Nesterov-momentum have had limited practical use, despite their massive theoretical popularity.

\vspace{-0.3cm}
\section{A systematic investigation}

In this section, we will provide a set of experiments to systematically understand when and (hopefully) why the neural network parameters do not converge to stationary points as theory mandates. In particular, we will try to test the following \textbf{hypotheses} in the experiments:
\vspace{-0.1cm}

\begin{enumerate}[leftmargin=*,itemsep=0.02cm]
    \item The nonconvergence is due to the fact that the step size is not small enough or the model is not trained long enough.
    \item This phenomenon is restricted to the ResNet + ImageNet task, or models with non-differentiable ReLUs. 
    % \item The step size decreased too fast before the gradient norm could converge.
    \item The large gradient norm is due to estimation error.
\end{enumerate}
In the end, we will see that these hypotheses \textbf{fail} to hold and that the phenomenon is quite common in large-scale tasks.

\subsection{Different learning rates and training schedules}

One immediate question following the observation in Figure~\ref{fig:imagenet-baseline} is whether the observed phenomenon holds solely for a particular stage-wise learning rate, which is not very common in theoretical analysis. To address this question, we run the same ResNet101 model on ImageNet just as before, except that we now use a constant learning rate across all 90 epochs of training. The evolutions of the quantities in~\eqref{eq:quantities} are summarized in Figure~\ref{fig:imagenet-constant}. A quick glance at the plots verifies that the gradient norm does not converge to $0$ in any of the experiments.  We further notice that, surprisingly, a smaller learning rate leads to a larger gradient norm, larger stochastic gradient noise intensity, and larger sharpness as observed in \citep{Cohen2021}. We will further discuss the implications of these observations later in the paper.

As the loss curves in the last two rows of Figure~\ref{fig:imagenet-constant} are still decreasing, another question could be that we didn't run the experiment long enough to achieve actual convergence. To address this question,  we continue the second row experiment (step size $\eta = 0.01$) for 300 epochs and present the result in the rightmost column of Figure~\ref{fig:imagenet-constant}. We can see that no clear progress was made after about 50 epochs. 

The above experiments show that in ImageNet + ResNet101 experiment, the parameters do not converge to stationary points. In the next section, we test whether this phenomenon is restricted to the particular data set and architecture.

% \begin{figure*}[htbp]
% 	\centering
% 	\hrule
% 	\vspace{0.1cm}

% % 	{
% % 		\includegraphics[scale=0.25]{figures/imagenet-cstlr001-ep300-train-sharpness}
% % 	}
%     \vspace{-0.5cm}
% 	\caption{
% 	The estimated stats vs epoch for training ResNet on ImageNet using a constant learning rate $\eta=0.01$ for 300 epochs.
% }\vspace{-0.5cm}
% \label{fig:imagenet-long}
% \end{figure*}

\vspace{-0.2cm}

\subsection{Transformer XL experiments}\label{sec:exp-tran}

We run Transformer-XL training on WT103 dataset for the language modeling task following the implementation of the original authors~\citep{Dai2019}. Our training procedure is exactly the same as the official code, except that we reduce the number of attention layers for the baseline model from 6 to 4. Aside from training with a cosine learning rate schedule with initial learning rate $\eta = 0.00025$, we also experimented with different constant learning rates. The result is summarized in Figure~\ref{fig:transformer-stats}. We found that the observations made before also apply to transformer XL.

\subsection{Refuted hypotheses from the systematic study}

\begin{figure*}[t]
	\centering
	{
		\includegraphics[scale=0.22]{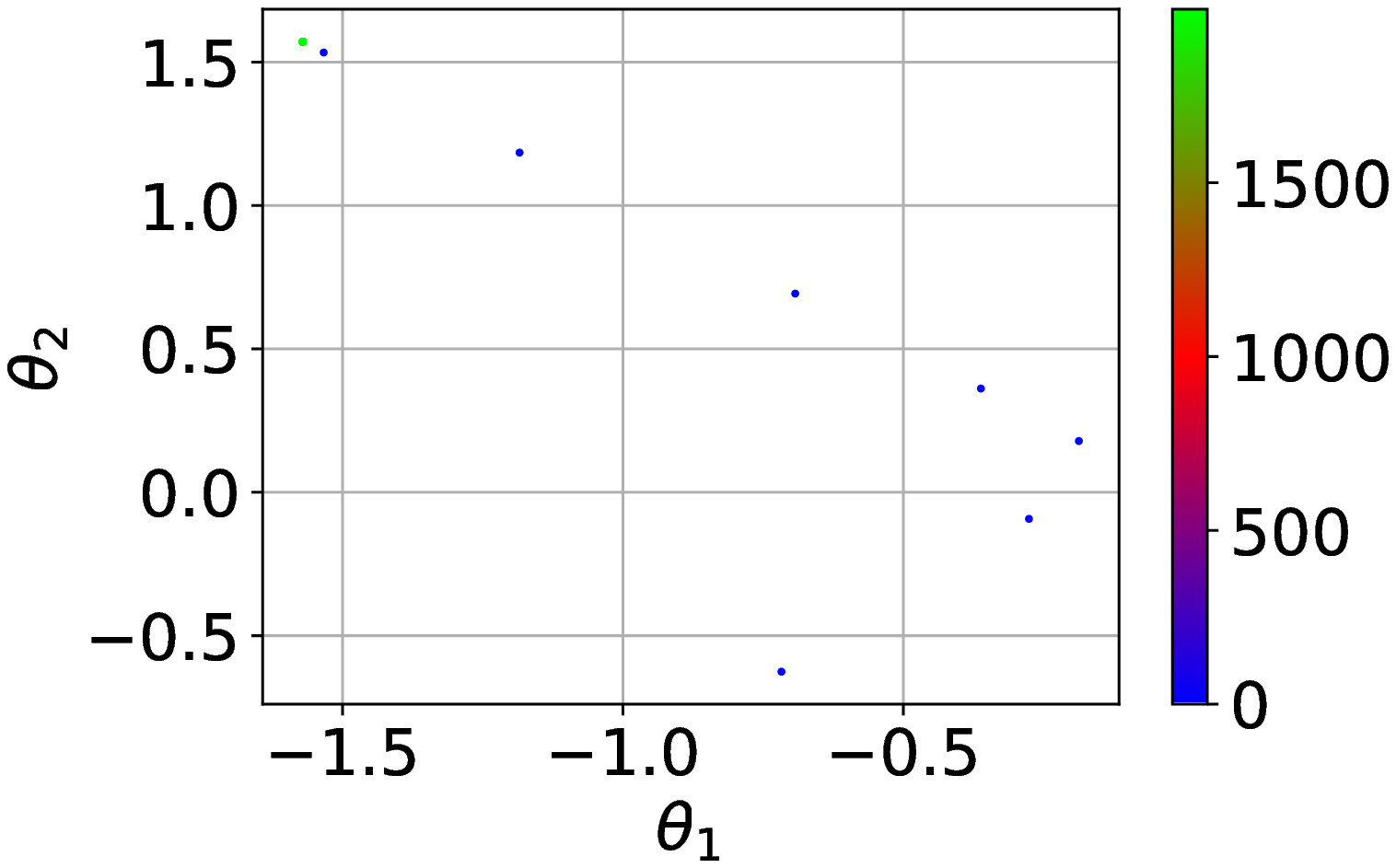}
	}\hspace{-0.2cm}
	{
		\includegraphics[scale=0.22]{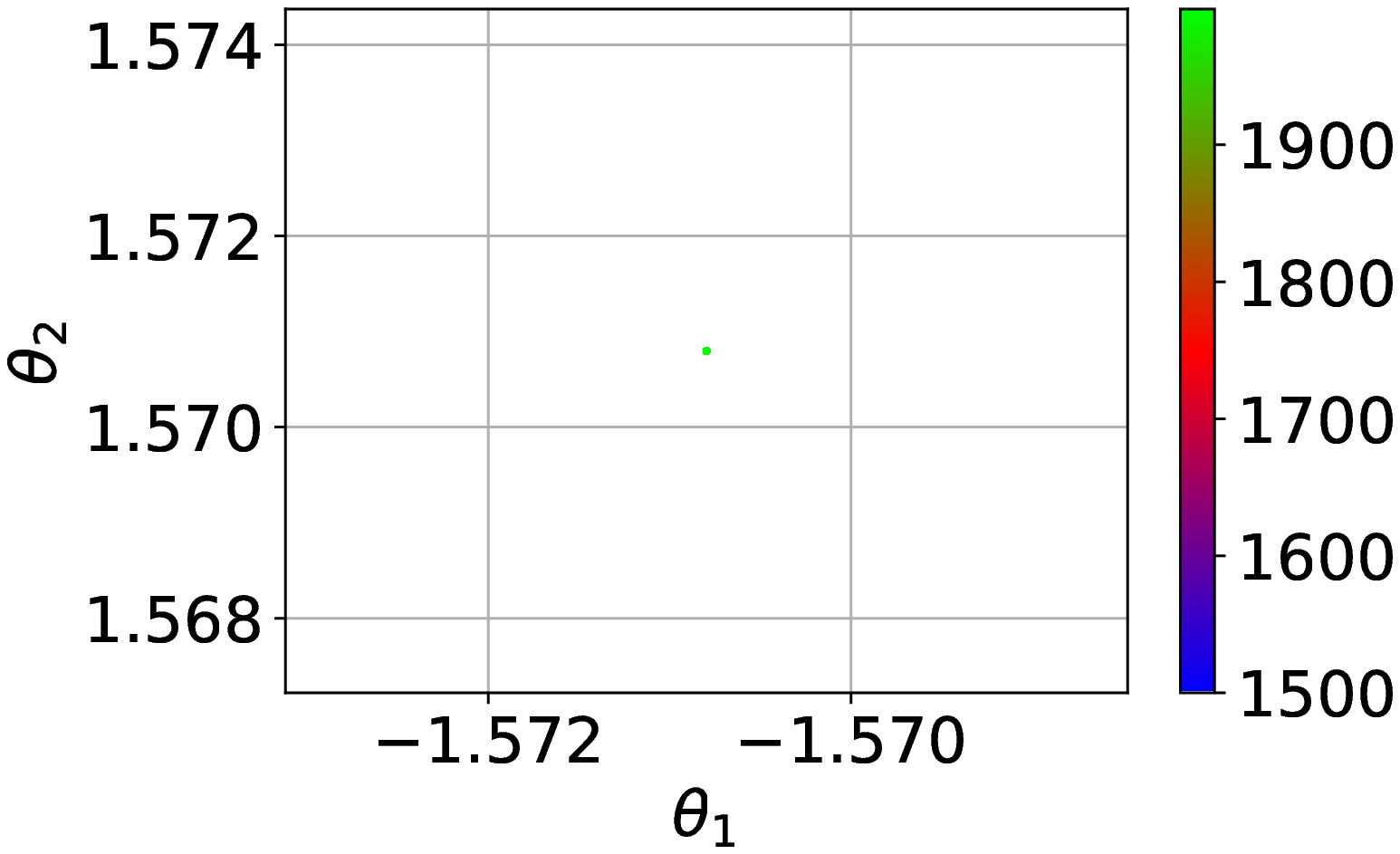}
	}
	{
		\includegraphics[scale=0.22]{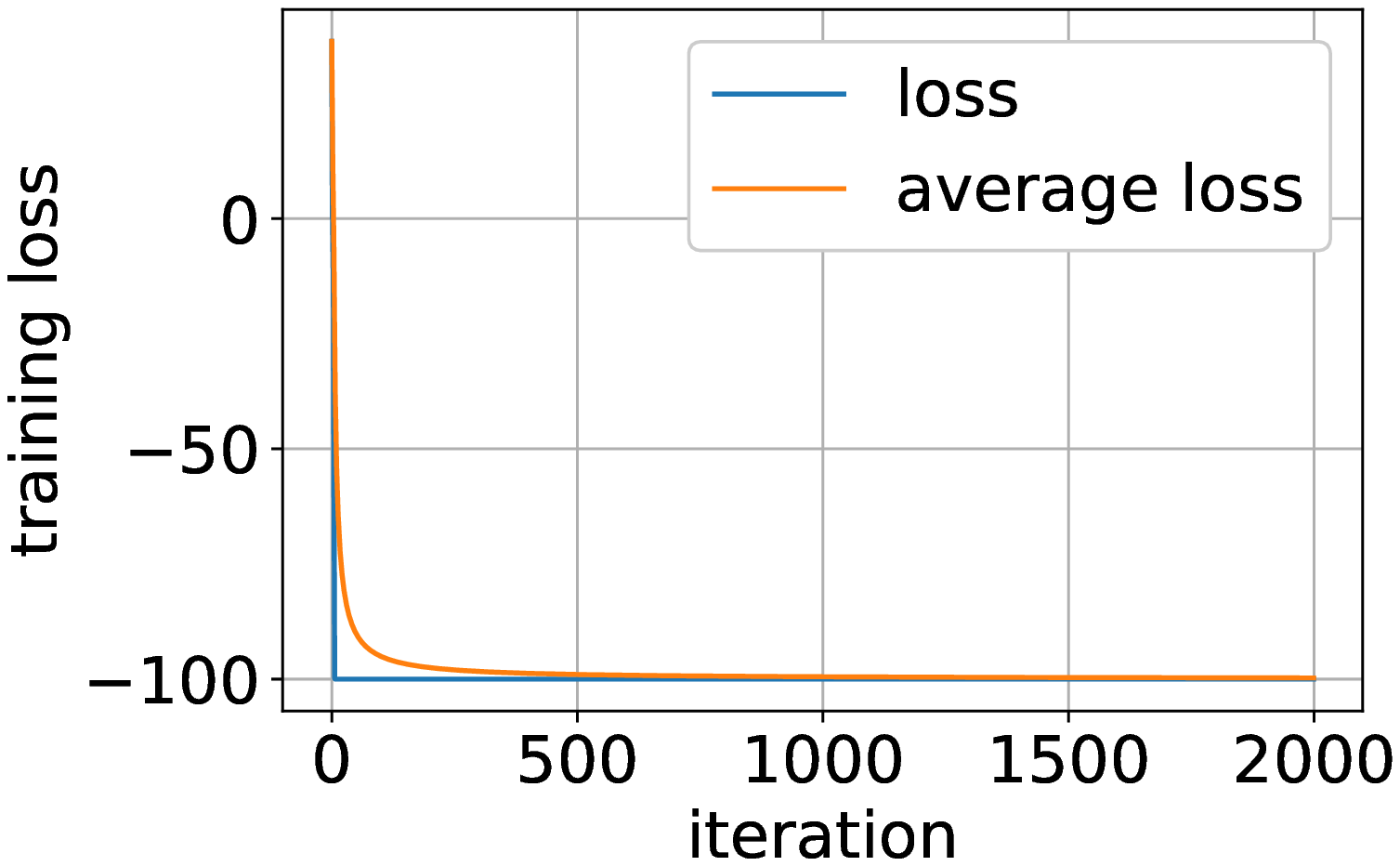}
	}	
	{
		\includegraphics[scale=0.22]{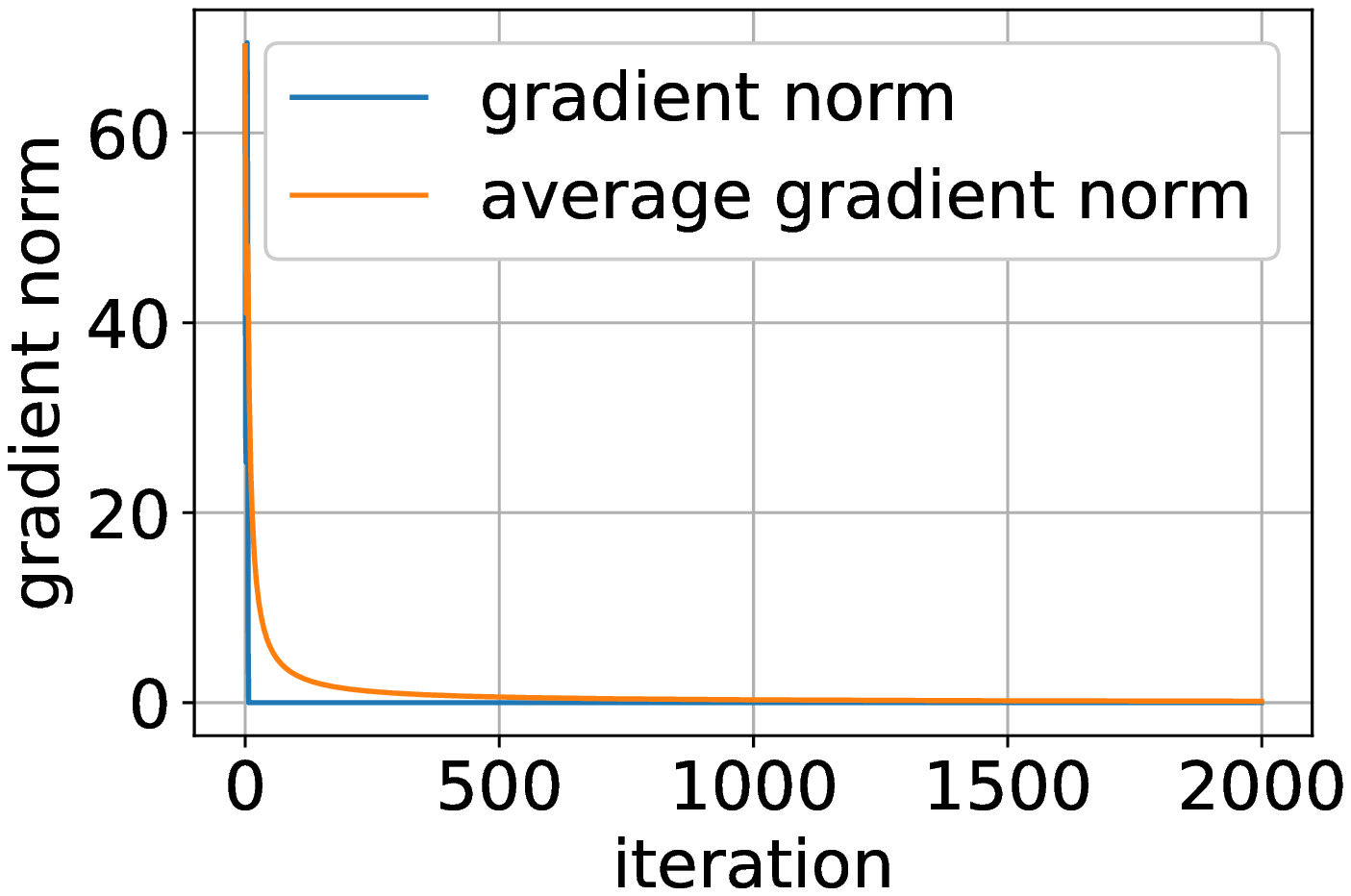}
	}
	
	\hspace{0.3cm}
	{
		\includegraphics[scale=0.22]{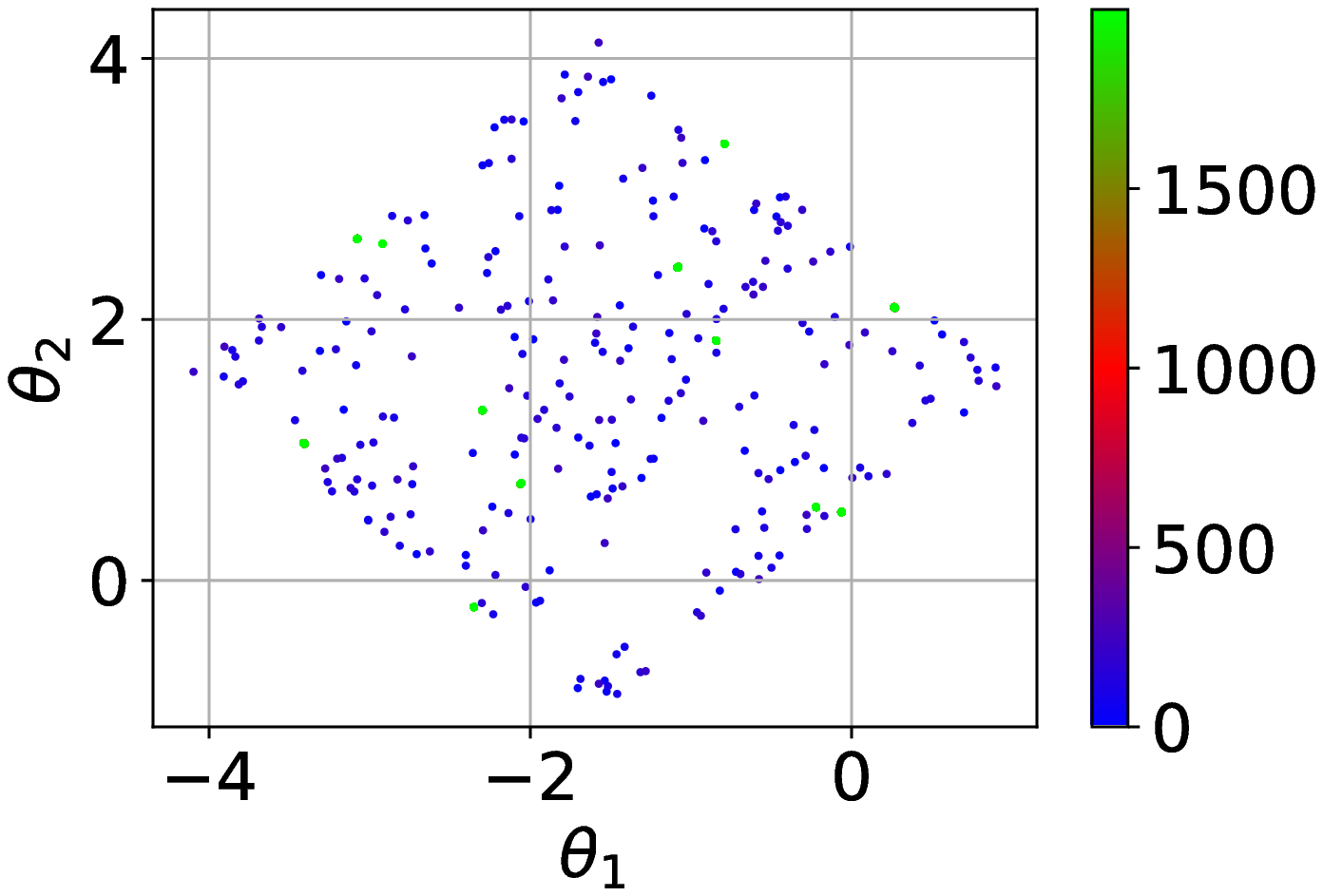}
	}\hspace{0.4cm}
	{
		\includegraphics[scale=0.22]{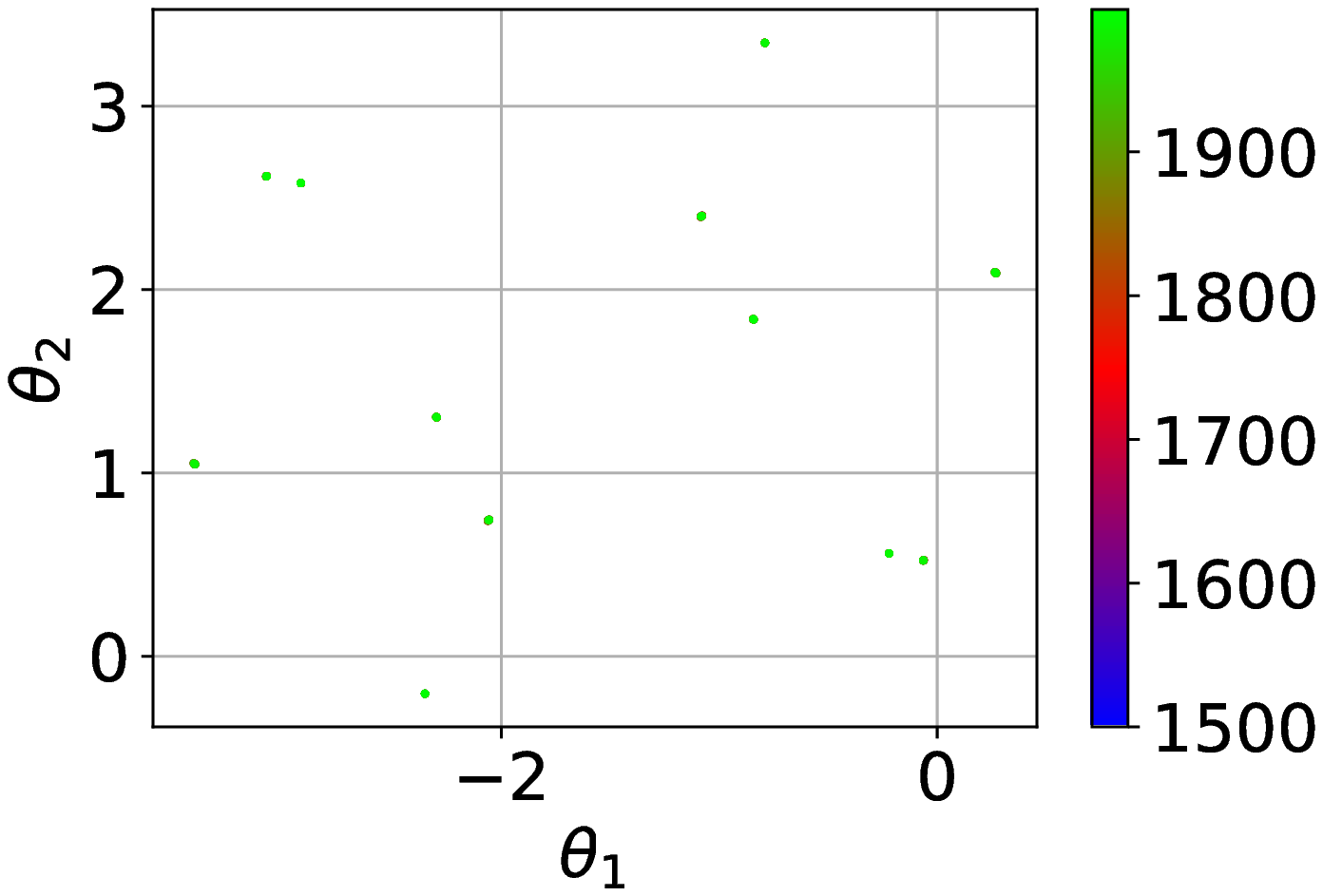}
	}
	{
		\includegraphics[scale=0.22]{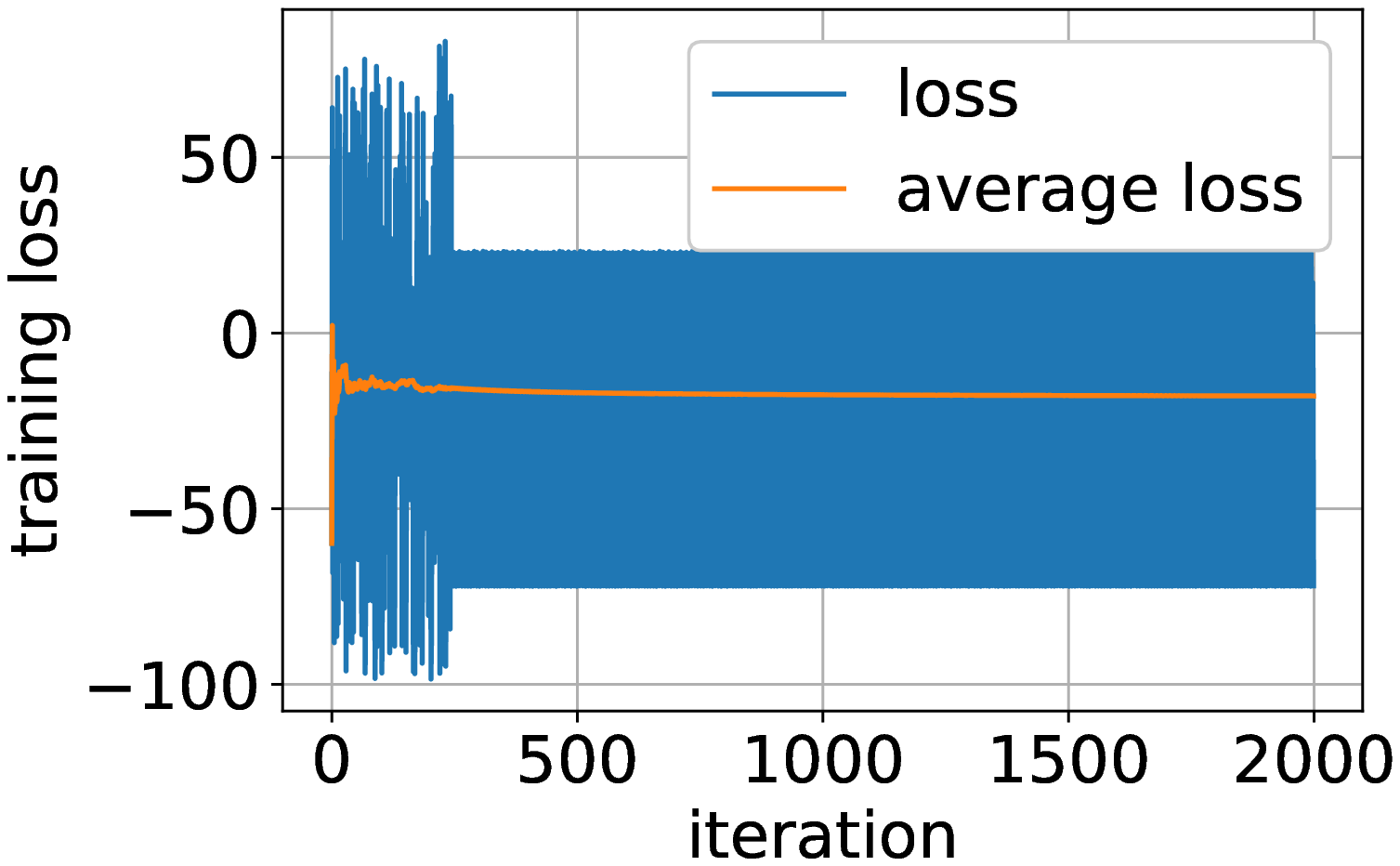}
	}	
	{
		\includegraphics[scale=0.22]{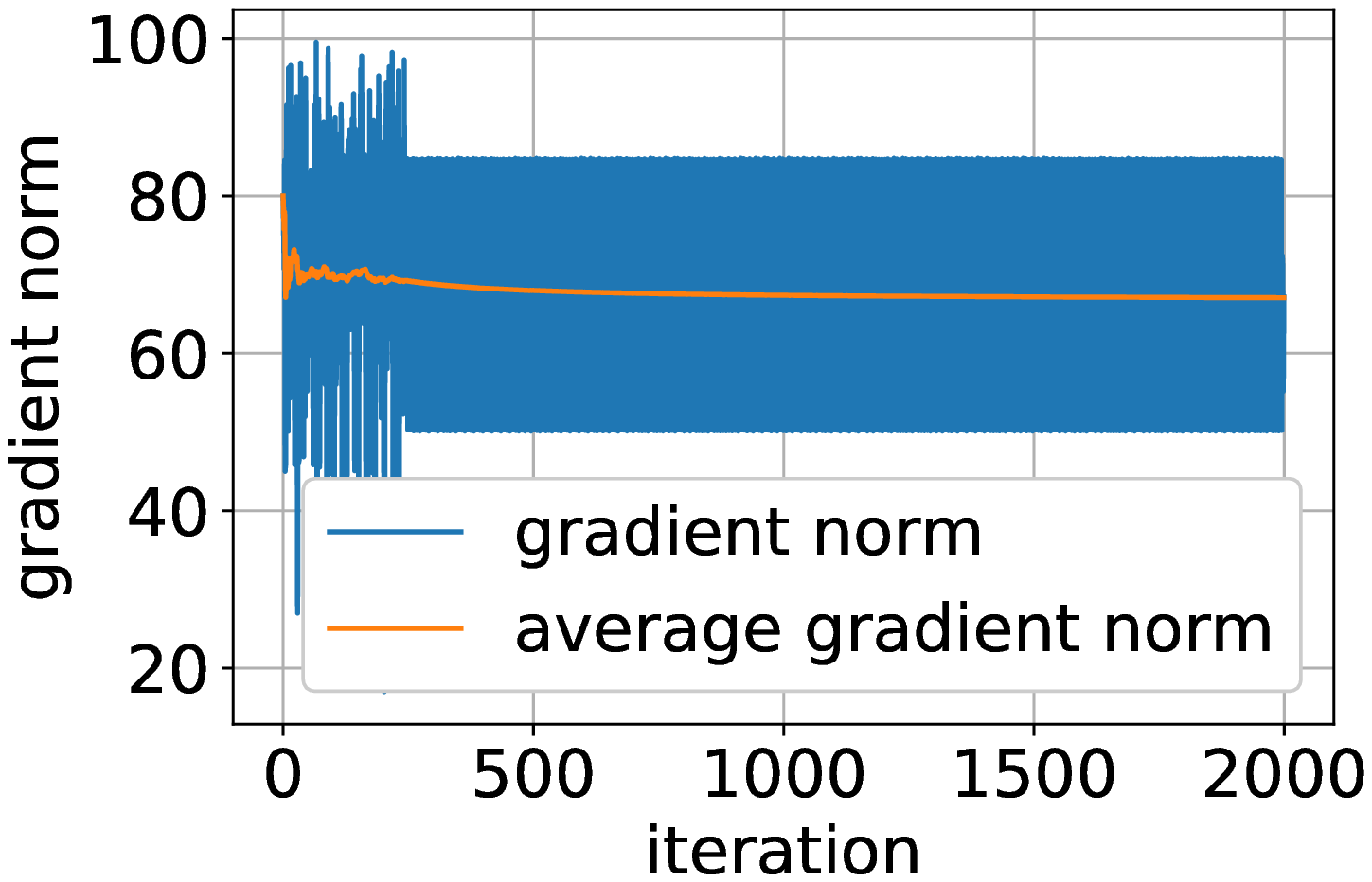}
	}
	\vspace{-0.4cm}
	\caption{
		Synthetic experiment. The learning rate is set to be $0.01$ and $0.04$ for the first and second row respectively. Column I: the whole trajectory in $2000$ iterations, where the scatter points correspond to iterates and the color of a point represents which iteration it is at; Column II: the trajectory in the last $500$ iterations to show the convergence behavior; Column III: training loss and average training loss vs iteration, where the average is taken over iterations; Column IV: gradient norm and average gradient norm vs iteration.
	}\vspace{-0.3cm}
	\label{fig:synthetic}
\end{figure*}

With the above set of experiments, we can already \textbf{exclude} the hypotheses at the beginning of this section.

\textbf{First}, in the rightmost column of Figure~\ref{fig:imagenet-constant}, 
we find that after running 300 epochs with a smaller step size, though the training loss dropped significantly, the gradient norm did not decrease. \emph{This confirms that even the qualitative analysis (let alone the quantitative convergence rates) on when gradient norm gets smaller from canonical optimization theory is not applicable to neural network training.}

\textbf{Second}, we see that the nonzero-gradient phenomenon in TransformerXL~\citep{dai2019transformer} training is even more prominent. In addition, as TransformerXL is differentiable, this also excludes the chance that oscillation is caused by non-differentiability. The \textbf{third} conjecture is refuted due to our estimation precision discussed later in Appendix~\ref{sec:exp-precision} with additional experimental details. We also observed that in our Cifar10 experiment in Appendix~\ref{sec:exp-cifar}, the gradient norm can indeed go to zero.

Given the above evidence, we believe that the convergence of training loss without reaching stationary points is caused by more fundamental and nontrivial reasons. To understand this phenomenon, we later will develop a notion of convergence based on the theory of dynamical systems. 

\begin{figure}[htbp]
	\centering

  \begin{minipage}[c]{0.2\textwidth}
  \centering
	\includegraphics[width=\textwidth]{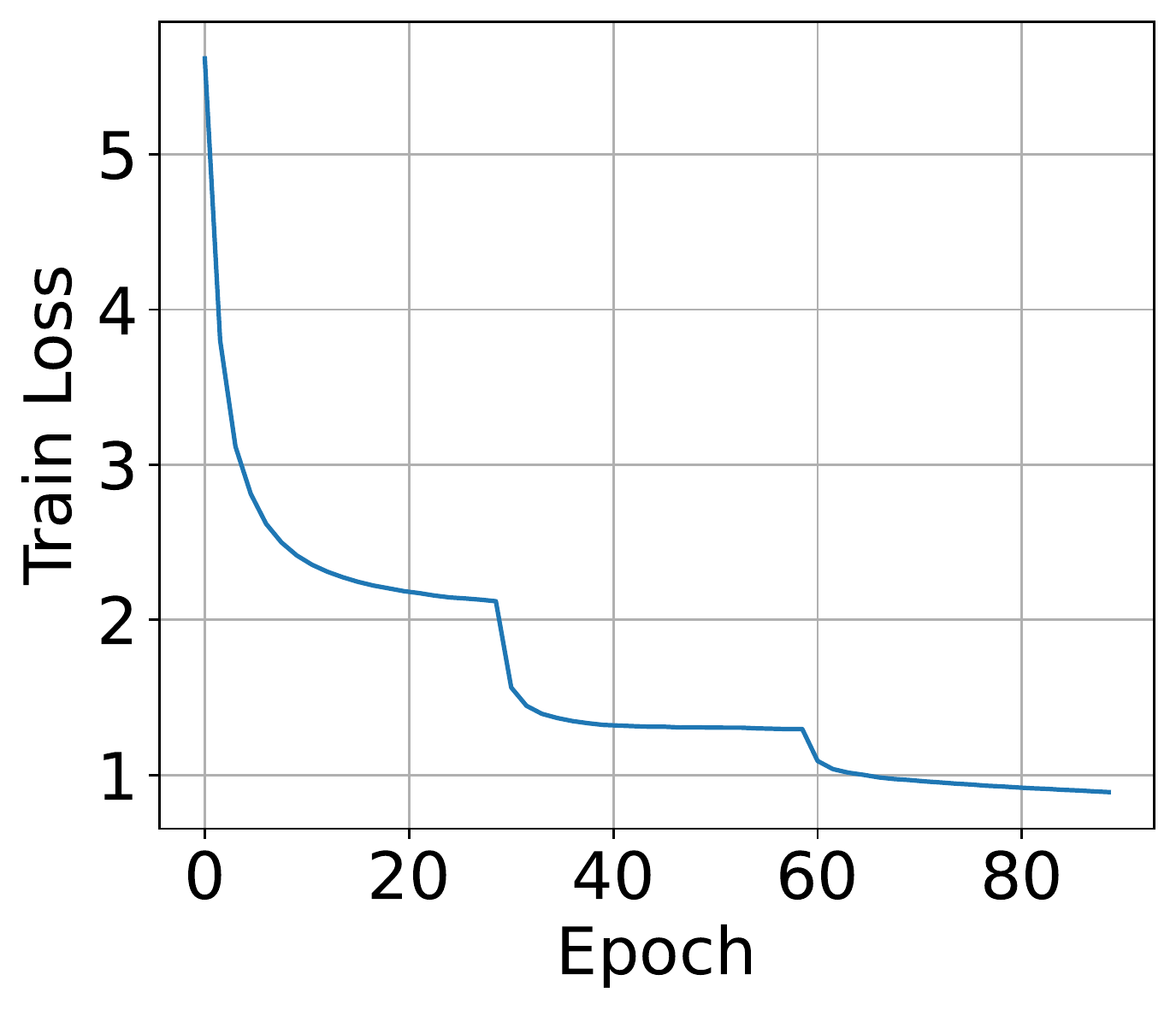}
   \end{minipage}
  \begin{minipage}[c]{0.2\textwidth}
  \centering
	\includegraphics[width=\textwidth]{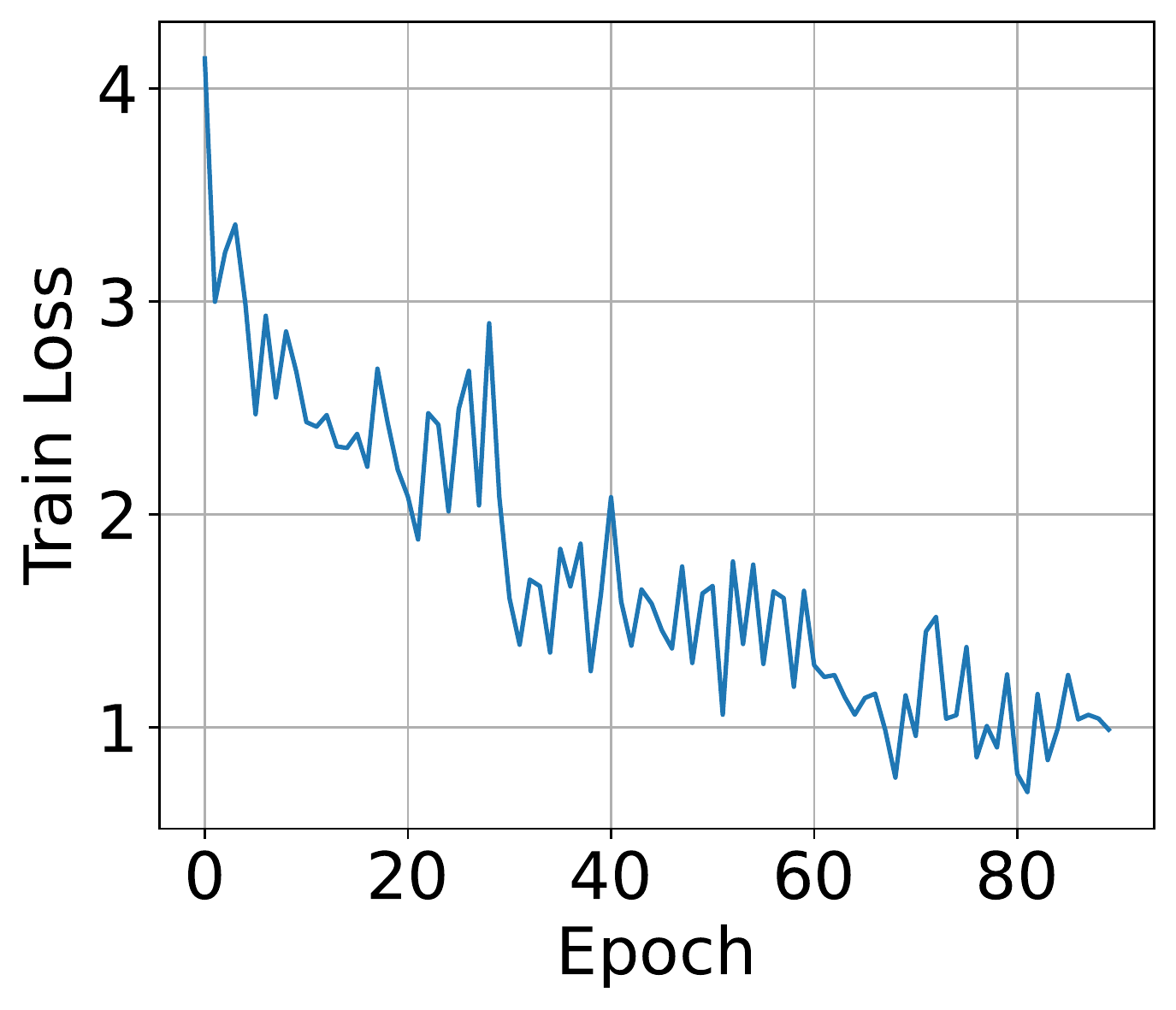}
  \end{minipage}
    \caption{
The \textbf{ full batch} training loss vs epochs in default ImageNet training. The left plot computes loss in the usual way as an moving average during training~\eqref{eq:moving_average}, where as the right plot computes the loss at the last iteration of a training epoch using~\eqref{eq:fixed_average}.
    }\label{fig:imagenet-last}\vspace{-0.3cm}

\end{figure}

Before diving into our theorems, we start with an interesting observation in Figure~\ref{fig:imagenet-last}. Here, we evaluate the \emph{full batch} training loss in two ways. The left one is to compute a moving average during the training epoch:
\begin{align}\label{eq:moving_average}
    \text{Loss} = \tfrac{1}{N}\textstyle\sum_{i=1}^N \ell(\theta_{\textcolor{red}{i}}, x_i),
\end{align}
where $i$ denotes the iteration number within an epoch of $N$ minibatches, $x_i$ denotes data from $i_{th}$ minibatch and $\theta_i$ denotes the network parameter at iteration $i$. The other is to compute the full batch loss at the last iteration $N$,
\begin{align}\label{eq:fixed_average}
    \text{Loss} = \tfrac{1}{N}\textstyle\sum_{i=1}^N \ell(\theta_{\textcolor{red}{N}}, x_i).
\end{align}
We notice that though both evaluations consumed the entire dataset, averaging the minibatch losses across all training iterations leads to a much more smooth loss curve than evaluating all the minibatch losses at a fixed iteration. Hence, one explanation is that the \textbf{time average of the  loss} rather than the \textbf{iteration-wise loss} converges,  while the gradient norm is nonzero due to nonsmoothness, and that the actual weight iterates keep oscillating.  To provide more intuition, in the next subsection, we provide a conceptual explanation through a synthetic experiment.

\subsection{An explanatory synthetic experiment}

The curious phenomenon discussed above is not limited to neural network training.  In what follows we present a simple synthetic example to illustrate the intuition behind the convergence behavior to unstable cycles rather than stationary points.

To this end, we simulate gradient descent on the objective function $f(\theta_1,\theta_2)=100\sin \theta_1\sin \theta_2$ whose smoothness and Lipschitzness parameters are both $L_f=100$. It is well known that gradient descent with a learning rate $\eta<2/L_f=0.02$ provably converges to stationary points for such a smooth function. As shown in the first row of Figure~\ref{fig:synthetic}, the iterates converge to a fixed point very fast with $\eta=0.01$. Moreover, the gradient norm converges to zero, which means a stationary point is reached at convergence. 

However, when $\eta>2/L_f$, which is often the case for neural network training, gradient descent no longer converges to stationary points as shown in the second row of Figure~\ref{fig:synthetic} with $\eta=0.04$. 
During the last $500$ iterations,
the iterates only take values around a few points and keep oscillating among them. As a result, the training loss and gradient norm also oscillate and do not converge in the usual sense.
However, the oscillation among these points follows some periodic pattern. If we collect all the iterates during a long enough training process, their empirical distribution will converge to a discrete distribution over those points that capture the periodic pattern. Then if we take an average of the training losses or gradient norms over time, it must converge to the average value of the periodic points, as shown in the last two images in Figure~\ref{fig:synthetic}. However, although the average gradient norm converges, the convergence value can not be zero in presence of oscillation, as gradient descent makes no updates if the gradient is zero.

The above example shows that the key to function value convergence could be that a time average rather than a spatial average is taken in evaluating the loss. Hence, the convergence only happens in \textbf{time average }sense.

% In fact, we could verify this intuition through the example below. Recall that in ImageNet training, the plotted training loss is a moving average of previous iterations. In the left plot of Figure~\ref{fig:imagenet-last} we instead plot the training loss of the last 50 epochs in the experiment shown in Figure~\ref{fig:imagenet-long} and see that the variation across iterations is quite large considering the learning rate is $0.01$ and the gradient norm is about $0.6$. On the right plot, we show the last a few iterations of transformerXL training (see section~\ref{sec:exp} for more details) and observe even larger oscillations. Furthermore, even in Cifar10 experiments, both~\citet{li2020reconciling,lobacheva2021periodic} show very strong periodic divergence in training loss when the number of training epoch is huge ($>1000$ epochs).  

\section{ Convergence beyond stationary points}

We saw above that even though the per-iteration loss does not converge, the time average with a long enough window size can converge. In this section, we provide a simple mathematical analysis to explain why that happens. In particular, we prove that the change in training loss evaluated as a time average converges to $0$ for neural networks. Our analysis is motivated by, and follows the proof of the celebrated Krylov-Bogolyubov theorem. As a result, we refer to our interpretation as the \emph{invariant measure perspective}. 

Particularly, we say a measure $\mu$ is an \textbf{invariant measure} for the map $F: \cX \to \cX$ if for any measurable set $A$
\vspace{-0.1cm}
\begin{align*}
	\mu (A) = \mu(F^{-1}(A)) = \int_{\theta} \indicator\{F(\theta) \in A\} d \mu(\theta),
\end{align*}
where $F^{-1}(A) = \{\theta | F(\theta) \in A\}.$ Notice that if $F$ is a stochastic update, then this should be read as
\begin{align} \label{eq:pushforward-st}
	\mu (A) = \mu(F^{-1}(A)) = \int_{\theta} \Pr\{F(\theta) \in A\} d \mu(\theta).
\end{align}
In other words, the pushforward of $\mu$ under $F$ stays unchanged, $F\#\mu = \mu$.

Invariance of measure is closely related to convergence of function values. To see this, consider the dynamical system
\vspace{-0.1cm}
\begin{align*}
	\theta_{t+1} = F(\theta_t).
\end{align*}
In such a scenario, for any continuous function $\phi$, the function value  does not change after update when the variable is sampled from an invariant measure,
\vspace{-0.1cm}
\begin{align*}
	\E_{\theta\sim\mu} [\phi(\theta)] = \E_{\theta\sim\mu}[\phi(F(\theta))].
\end{align*}
Recall that our key insight is that the convergence of the training loss occurs in a time-average sense, which naturally leads to the following notion of empirical measure:
\vspace{-0.1cm}
\begin{align}\label{eq:empirical}
	\mu_k :=  \textstyle\frac{1}{k} \sum_{t=1}^k \delta_{\theta_t},
\end{align}
where $\delta_{\theta}$ denotes the Dirac measure supported on the value $\theta$, i.e., $\delta_{\theta}(A) = 1$ if and only if $\theta \in A$, and $\{\theta_1, \theta_2, \cdots\}$ are the sequence of iterates generated by the dynamical system. With this notation, we can conveniently write the time average of a scalar function $\phi: \cX \to \R$ as
\vspace{-0.1cm}
\begin{align}
	\mu_k (\phi) = \E_{\theta \sim \mu_k}[\phi(\theta)].
\end{align}
We focus on the case when the dynamic system $F(\theta_t)$ denotes the SGD update, i.e.,
\begin{align*}
 F(\theta_t) = \theta_t - \eta g(\theta_t),
\end{align*}
 where $g(\theta_t)$ denotes the stochastic gradient and $\eta$ denotes the step size. Next, we will show that the empirical measure converges to an approximately invariant measure as the number of iterations grows.

\subsection{Vanishing change in neural network training}
\label{sec:neural}
 We are now ready to provide a theoretical analysis to prove the vanishing gain of training losses in neural network training, and thus explain how the training loss can stabilize even when the norm of the loss function gradient is non-zero. Our analysis is distinct from previous ones (e.g. \cite{chizat2018global,mei2019mean,jacot2018neural}) in the literature in that it \emph{does not} assume global Lipschitzness or smoothness, \emph{does not} rely on bounded noise assumptions, and it \emph{does not} require perfectly fitting the data as in Neural tangent kernel models or mean-field style arguments. Instead, it builds upon minimal practical conditions. The downside of this generality is that we only prove convergence of function values and do not comment on local or global optimality or generalization. We believe much remains to be done here and we have just scratched the surface. We will make more comments on this in Section~\ref{sec:discussion}.

To start the discussion, we  define the following $L$-layer deep neural network $f(x,\theta)$, where $x$ is the input and $\theta=(W_0,\ldots,W_{L-1})$ is the network weights:
% 	&x_0=x,\nonumber, \quad
\begin{align}
\label{eq:nn}
	&x_{l+1}=\sigma_{l+1}(W_lx_l), \quad l=0,\ldots L-1\\
	&f(x,\theta)=x_L,
\end{align}
where $\sigma_l$ is a coordinate-wise activation function (e.g., ReLU or sigmoid). In practice, the last layer usually does not use any activation function so $\sigma_{L}$ is the identity mapping. We do not consider pooling layers, convolutional layers, or skip connections for now and it should be easy to extend our analysis to these settings.  Iteration~\eqref{eq:nn} does not include batch normalization layers which we will analyze later in this section. Given a training dataset $S=\{(x^i,y^i)\}^N_{i=1}$, the empirical training loss is defined as
\begin{align*}
	L_S(\theta):=\tfrac{1}{N}\textstyle\sum_{i=1}^N \ell(f(x^i,\theta),y^i),
\end{align*}
where $\ell:\R^d\times [d]\to \R$ is a loss function and we assume $\norm{x^i}_2\le 1$.
The network is trained by SGD with weight decay, which is equivalent to running SGD on the following regularized loss
\vspace{-0.2cm}
\begin{align*}
	\ \ L_S^\gamma(\theta):=	L_S(\theta)+\frac{\gamma}{2}\norm{\theta}^2_2,
\end{align*}
where $\norm{\theta}_2$ denotes the $\ell_2$ norm of vectorized $\theta$. We will focus on the most widely used loss function for classification tasks, the cross-entropy after softmax, defined as follows.
\begin{align}
	\ell(x,y)=x_y-\log\left(\textstyle\sum_{j=1}^d e^{x_j}\right),
	\label{eq:cross-entroy}
\end{align}
which has the following properties that we will use later.
\begin{lemma}
	The cross-entropy after softmax loss $\ell:\R^d\times [d]\to \R$ defined in \eqref{eq:cross-entroy} satisfies
    \vspace{-0.2cm}
    \begin{enumerate}[leftmargin=*,itemsep=0.03cm]
    \item If $\max_i x_i-\min_i x_i\le c$, we have $\ell(x,y)\le c+\log d$ for any $y\in [d]$.
	\item $\ell(x,y)$ is $c_{\ell}$ Lipschitz w.r.t.\ $x$ for a global constant $c_{\ell}$.
	\end{enumerate}    \vspace{-0.2cm}
	\label{lem:cross_entropy}
\end{lemma}

Next, we make the following assumption for the activation function. It holds for ReLU and tanh activation functions.
\begin{assumption}\label{assump:activation-lipschitz}
	Each activation function $\sigma_l$ is (sub)-differentiable and $c_\sigma$ coordinate-wise Lipschitz for some numerical constant $c_\sigma>0$. Also assume $\sigma_l(0)=0$.
\end{assumption}
 Now we can prove the vanishing gain of the function values.

\begin{theorem}\label{thm:compact_domain}
Suppose Assumption~\ref{assump:activation-lipschitz} holds and $\theta$ is initialized within the compact set $C_w:=\{(W_0,\ldots,W_{L-1}): \norm{W_l}_{\text{op}}\le w\}$ for some $w\le (\gamma/c_\ell c_\sigma^{L})^{1/(L-2)}$. Then the iterate $\theta_k$ for every $k$ lies in $C_w$ and the empirical measure generated by SGD with a stepsize $\eta\le 1/\gamma$ satisfies with probabiblity $1-\delta$ that
	\begin{align*}
		\E_{\theta \sim \mu_n}[L_S(\theta) - L_S(F(\theta))] = \cO(\tfrac{\log(1/\delta)}{\sqrt{n}}).
	\end{align*} \vspace{-0.3cm}
\end{theorem} 
The above theorem states that the change in loss vanishes in a time average sense. The key step in the previous proof is to show that all iterates lie in a compact region almost surely even though the function may not be smooth or Lipschitz continuous.  One may suspect that a stronger result should hold that the limit will exist. We show in Section~\ref{sec:existlimit} that such a statement is highly nontrivial and sometimes false.

We notice that the initialization choice may not always hold in practice, especially when there is batch normalization design. We further note that similar to the above theorem, all (piece-wise) continuous scalar functions including the noise norm are bounded by compactness, and hence should stabilize after long enough training. However, in the third row of Figure~\ref{fig:imagenet-constant}, the noise norm does not really converge. To explain this observation, we propose the following theorem that studies neural networks with batch normalization.

For simplicity of analysis, we assume the last layer is one of the layers with batch normalization. For a vector $x$, we use $x^2$, $\abs{x}$ and $\sqrt{x}$ to denote its coordinate-wise square, absolute value, and square root respectively. In the $l$-th layer, if it uses batch normalization, given a batch $\cB=\{(x^i,y^i)\}_{i=1}^m$ sampled from a distribution $\mathcal{P}_{\cB}$, batch normalization makes the following update from $\{x_{l-1}^i\}_{i\in\cB}$ to $\{x_{l}^i\}_{i\in\cB}$:
\begin{alignat*}{2}
    \mu_{\cB,l-1} =& \tfrac{1}{m}\textstyle\sum_{i\in\cB} x_{l-1}^i, \\
    \sigma_{\cB,l-1}^2 =&\tfrac{1}{m}\textstyle\sum_{i\in\cB} \left( x_{l-1}^i-\mu_{\cB,l-1}\right)^{ 2},\\
    \hat{x}_l^i=&\tfrac{x_{l-1}^i-\mu_{\cB,l-1}}{\sqrt{\sigma_{\cB,l-1}^2+\epsilon}}, \quad 
    {x}_l^i=a_l \cdot\hat{x}_l^i+b_l,
\end{alignat*}
where $a_l$ and $b_l$ are the scale and shift parameters to be trained. We also use SGD with weight decay in training.

\begin{theorem}[With batch normalization]
	Suppose the parameter of batch normalization layer $a_L$ is initialized within the compact set $|a_L|\le 2\sqrt{m}/\gamma$. Then the empirical measure generated by SGD with $\eta\le 1/\gamma$ satisfies with probability $1-\delta$ that,
	\begin{align*}
		\E_{\theta \sim \mu_n,\cB\sim\mathcal{P}_{\cB}}[L_{\cB}(\theta) - L_{\cB}(F(\theta))] = \cO(\tfrac{\log(1/\delta)}{\sqrt{n}}).
	\end{align*}
	
	where $\mathcal{P}_{\cB}$ denotes the distribution of random minibatches.
	\label{thm:bn_bounded}
\end{theorem}
We have shown in this section how the expected change of the training loss in per iterate update converges to zero for neural network training without any smoothness or Lipschitzness assumptions. One weakness of our analysis is that the limit of the training loss may not exist. Another caveat is that our gain is measured in terms of empirical measure instead of the last iterate distribution. In the next section, we discuss the many implications and open problems that the invariant measure view brings us.

%\hl{we can get a better bound of $2\sqrt{m}+\log d$ if not considering the last step of scale and shift, which does not need weight decay, either. Also note that we are bounding the minibatch loss $L_\cB(\theta)$ instead of $L_S$.}

\section{Theorems implications and open questions} \label{sec:discussion}

We showed that in neural network training, the change in training loss gradually converges to 0, even if the full gradient norm does not vanish. In this section, we will show how this result explains and connects to several observed phenomenons that were not captured by the canonical optimization framework. We then conclude by discussing several limitations of our result and important future directions.

\subsection{Edge-of-stability and relaxed smoothness}

The invariant measure perspective can also provide insight into the edge-of-stability observation and relaxed smoothness phenomenon. Our argument is heuristic, and somewhat speculative. We believe a rigorous analysis is both interesting and challenging and leave them as future directions.

We start from the equation~\eqref{eq:equilibrium} in the proof of the Theorem~\ref{thm:smaller-step} in Appendix~\ref{proof:small}. If the variable $\theta$ follows the distribution of an invariant measure, then by the fact that the expected loss does not change after one SGD update, 
\begin{align*}
&\E_{\theta,  g}[ \| \nabla L_S(\theta)\|_2^2] = \\
&\E_{\theta,  g} \left[\eta  \iint_0^1   \iprod{ g(\theta) }{ \nabla^2 L_S(\gamma_{\theta, g}(t\tau \eta)) g(\theta)} dtd\tau  \right],
\end{align*}
where $g(\theta)$ is the stochastic gradient and $\gamma_{\theta, g(\theta)}(r) = \theta - r g(\theta)$ denotes the line segment.
Then we boldly extract an equation that holds in expectation 
\begin{align}
    (\nabla)^2 = \eta \mathcal{L} G^2,
\end{align}
where $\nabla$ denotes the gradient norm, $\mathcal{L}$ denotes the sharpness in the update direction and $G^2$ denotes the second moment of stochastic gradients. The only approximation we made is that we replaced the hessian integral along the line segment $\gamma_{\theta, g(\theta)}$ by sharpness. This equation has some interesting connections to the following two observations. 

First, we recall the edge-of-stability framework~\citep{Cohen2021}, which observes that the actual smoothness constant during training neural network has an inverse relation to step size. This is true from the above equation if we hold $\nabla, G$ constant. Second, in~\citep{Zhang2019e}, the authors identified a positive correlation between the gradient norm and the smoothness constant. This relation can also be extracted from the equation if $\eta, G$ are held constant.

In fact, as we observe that in practice, the relation between the sharpness and step size is not a direct inverse but indeed has some negative correlation. Therefore, we believe that through a more rigorous analysis of the property of invariant measures, one could understand why many counter-intuitive behaviors can happen, and provide a more accurate model of the interaction between different quantities.

\subsection{Decreasing stepsize leads to smaller objective values}

One well-known observation in neural network training is that when the training loss plateaus, reducing the learning rate can further reduce the objective. This phenomenon can be proved in theory if the function has globally bounded noise and smoothness constant. However, as we showed, the smoothness and noise level change adversarially to the step size. In this section, we provide a partial explanation on when a smaller step size can decrease the function value. In particular, we consider the neural network setup introduced in Section~\ref{sec:neural}. We make the following assumption:
\begin{assumption}
	The neural network is continuously second-order differentiable, though it need not necessarily have bounded smoothness.
\end{assumption}

Then we can prove that reducing the step size would result in a decrease in function value.

\begin{theorem}\label{thm:smaller-step}
	Consider the stochastic gradient update $F:\cX \to \cX$ on a compact set defined as $F(\theta) = \theta -\eta g(\theta)$ for a fixed step size $\eta > 0$. Let $\mu$ be the invariant distribution  such that $\E_{F, \theta\sim \mu}[L_S(\theta)] = \E_{F, \theta\sim\mu}[L_S(F(\theta))]$. If $\mu$ is not supported on stationary points (i.e. $\E_{\theta \sim \mu}[\|\nabla f(\theta)\|_2^2] > 0$), then there exists a small enough $c\in(0,1)$ such that for any positive step size $\eta' < c\eta$, the update $F'(\theta) = \theta -\eta' g(\theta)$  will lead to a smaller function value, i.e.
	\begin{align*}
		\E_{F', \theta \sim \mu}[L_S(F'(\theta))] < \E_{\theta \sim \mu}[L_S(\theta)].
	\end{align*}
\end{theorem}

The above theorem states that once the change in loss vanishes, by selecting a smaller step size, one could further reduce the loss. This reflects the observation in Figure~\ref{fig:imagenet-baseline}. The challenge in the proof is that reducing the step size might lead to worse smoothness that is too large for the step size, and hence may increase the training objective. The proof can be found in Appendix~\ref{proof:small}.

However, Theorem~\ref{thm:smaller-step} only depicts what happens after one-step update rather than the long-term behavior after the iterates generated by a smaller step size converge. We believe that characterizing the shift from one invariant measure to another due to step size update could lead to a better understanding of the convergence rates of optimization algorithms, and is worth future studies.

\subsection{Vanishing change vs existence of a limit}\label{sec:existlimit}

Theorems~\ref{thm:compact_domain} and~\ref{thm:bn_bounded} show that the update vanishes to zero, yet they do not imply whether the limit $\lim_{t \to \infty} \mu_t(\phi)$ exists. In fact,  an explicit counterexample shows that the limit may not exist even for a dynamic system defined on a compact domain with Lipschitz maps. 
\begin{theorem}[\cite{Yoccoz}]
  	There exist a compact set $\cX$, a dynamic system with deterministic continuous map $F: \cX \to \cX$ and a scalar function $\phi\in C^\infty: \cX \to \mathbb{R}$, such that sequence $\frac{1}{n} \sum_{k \le n} \phi(\theta_{k})$ has no limit, where $\theta_{k+1} = F(\theta_k)$.
\end{theorem}

Given the above negative result, we only know that a subsequence of the series of empirical measures will converge to the invariant set.

\begin{theorem}[convergence of distribution]\label{thm:invariant}
	Assume that  $F$ maps a compact set  $\cX$ to itself. Then the empirical distribution has a subsequence converging weakly to an ergodic distribution. In other words, there exists an invariant distribution $\mu$, and a \textbf{subsequence} of positive integers $\{n_k\}_{k\in\Z}$ such that
$		\mu_{n_k} \to_w \mu. $
\end{theorem}

The proof of the above theorem is similar to the proof for the Krylov-Bogolyubov theorem. We include the proof in Appendix~\ref{proof:invariant} for completeness. However, we note that the above two theorems do not make use of the gradient descent update or the neural network architecture. Whether the dynamic system resulting from gradient descent has exactly the same property is left as a challenging future problem.

\subsection{Discussion}
Our work introduces a paradigm shift in how the convergence of weights and loss function should be analyzed and defined. It suggests neural network training converges to approximate invariant measures \emph{when iterates fail to converge to a single point and the distribution of weights does not converge to a globally unique stationary distribution.} Our results, however, lead to more questions than answers. For example, what do we know about the last iterate instead of empirical distribution? How do the invariant measures of different neural network structures and gradient-based optimizers differ from one another? What kind of dynamics converge to invariant measures faster? We believe that answering these questions will require vastly different techniques compared to standard optimization theory. 

\subsection*{Acknowledgments} SS acknowledges support from an NSF CAREER award (number 1846088), and NSF CCF-2112665 (TILOS AI Research Institute). JZ acknowledges support from a IIIS young scholar fellowship.
\nocite{langley00}

\bibliography{main}
\bibliographystyle{icml2022}

%%%%%%%%%%%%%%%%%%%%%%%%%%%%%%%%%%%%%%%%%%%%%%%%%%%%%%%%%%%%%%%%%%%%%%%%%%%%%%%
%%%%%%%%%%%%%%%%%%%%%%%%%%%%%%%%%%%%%%%%%%%%%%%%%%%%%%%%%%%%%%%%%%%%%%%%%%%%%%%
% APPENDIX
%%%%%%%%%%%%%%%%%%%%%%%%%%%%%%%%%%%%%%%%%%%%%%%%%%%%%%%%%%%%%%%%%%%%%%%%%%%%%%%
%%%%%%%%%%%%%%%%%%%%%%%%%%%%%%%%%%%%%%%%%%%%%%%%%%%%%%%%%%%%%%%%%%%%%%%%%%%%%%%
\newpage
\appendix
\onecolumn

\section{Additional experiments details}\label{sec:exp}

In this section, we add some additional experiments and experimental details that supplement the results in Section~\ref{sec:example}. We showed that the observed phenomenon happens in large scale tasks. To supplement the result, we briefly comment on how smaller dataset presents different behavior by taking Cifar experiment as an example. In the end, we will discuss some experimental details on how the quantities in~\eqref{eq:quantities} are estimated.

\begin{figure*}[htbp]
	\centering
	
	{
		\includegraphics[scale=0.3]{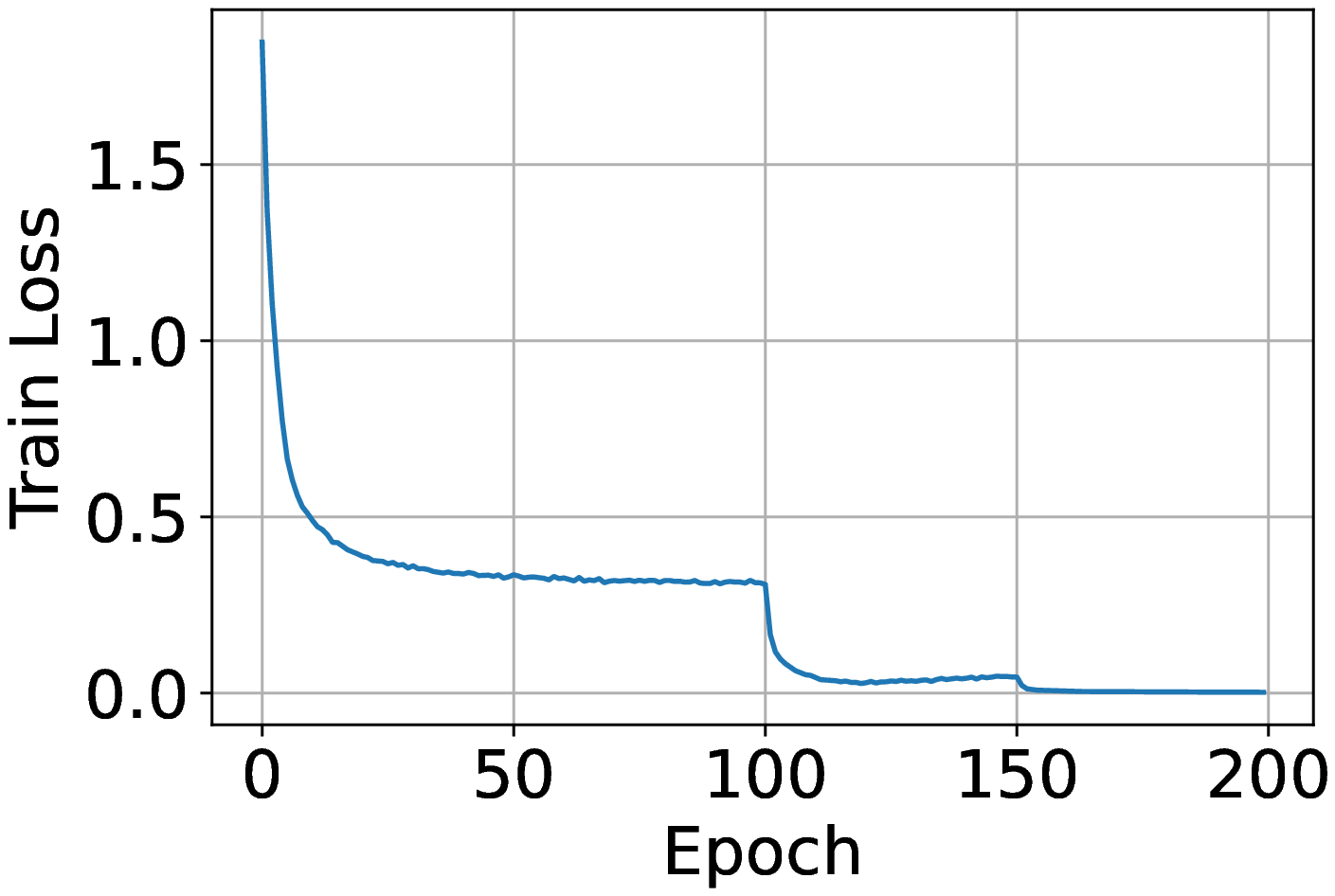}
	}	
	{
		\includegraphics[scale=0.3]{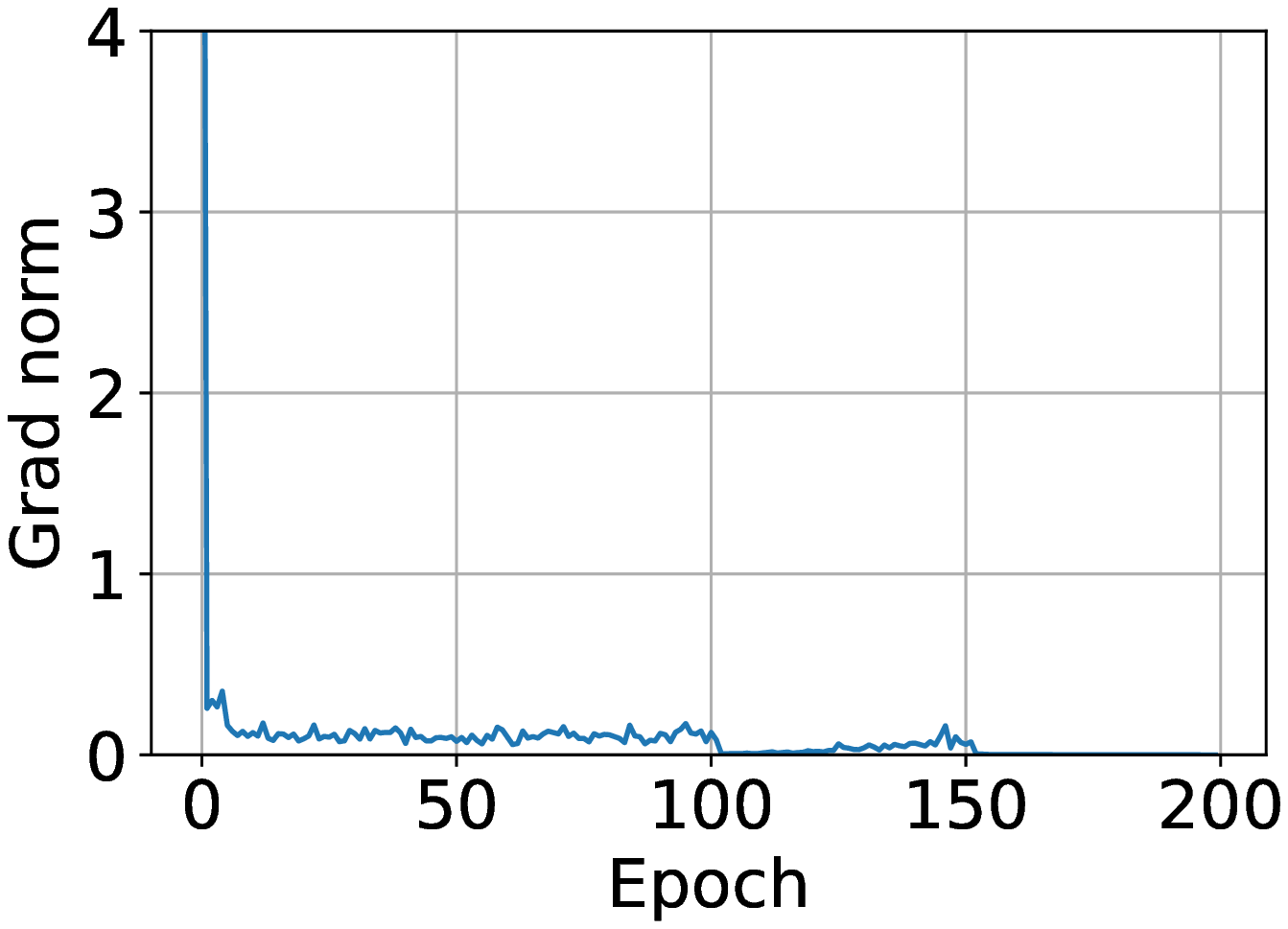}
	}
	{
		\includegraphics[scale=0.3]{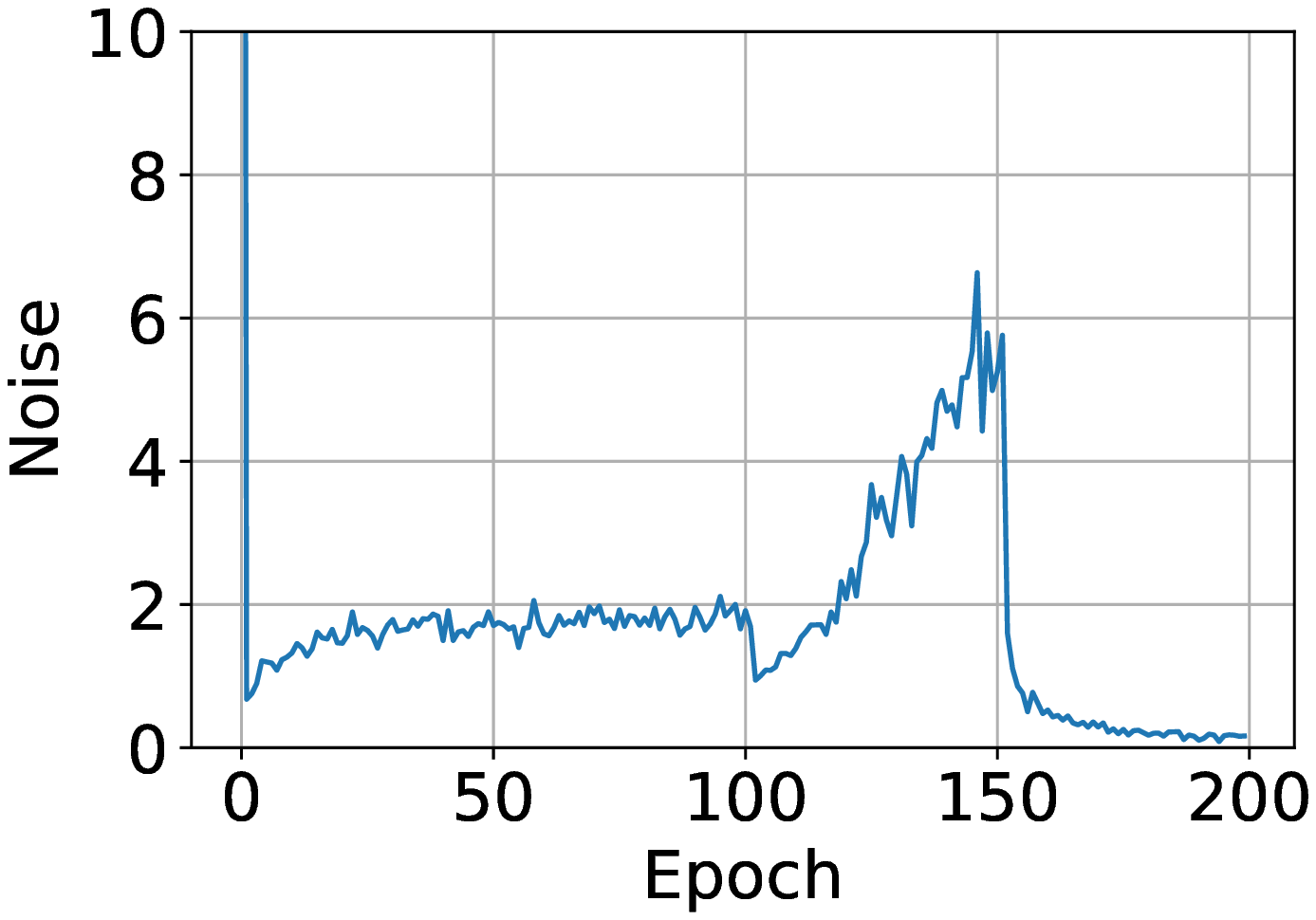}
	}
	\caption{
	The estimated stats vs epoch for Cifar10 training. The learning rate starts at 0.1 and decay by a factor of 10 at epoch 100 and epoch 150.
}
\label{fig:cifar}
\end{figure*}
\subsection{Cifar10 Experiment}\label{sec:exp-cifar}

In this section, we show how noise, gradient norm and training loss evolve in Cifar10 with ResNet training. Our training procedure is based on the implementation\footnote{\url{https://github.com/kuangliu/pytorch-cifar}}. The key result is demonstrated in Figure~\ref{fig:cifar}. We observe that in this case, the gradient norm indeed converges to 0. In fact, this is expected, as for cross entropy loss, the train loss could bound the gradient norm when weights are bounded. 

The implications of the above observations are many. First, this separation behavior between small overfitting model on Cifar10 and larger model on ImageNet shows that the study of overparametrization and convergenece to stationary point may still be true in many cases. However, we should be careful that these analysis does not apply to larger models that do not overfit the data. Second, this shows that the SDE modeling in~\citep{li2020reconciling, lobacheva2021periodic} can also be valid. It also shows that our work studies a problem of a different nature (non-zero grad norm).

\subsection{Estimating the statistics}\label{sec:exp-precision}

Here we provide additional details on how the values in~\eqref{eq:quantities} are estimated. Notice that these quantities are defined using all $N$ data points in the entire dataset, which is too large in practice. Therefore, we use a random batch $m < N$ to estimate the quantities. For training loss, gradient norm, and noise norm, the estimation is straight-forward. For the sharpness, we follow the implementation in \citep{wu2018sgd}\footnote{\url{https://github.com/leiwu0/sgd.stability}} and estimate the sharpness via power iterations.

\begin{figure*}[htbp]
	\centering

	{
		\includegraphics[scale=0.4]{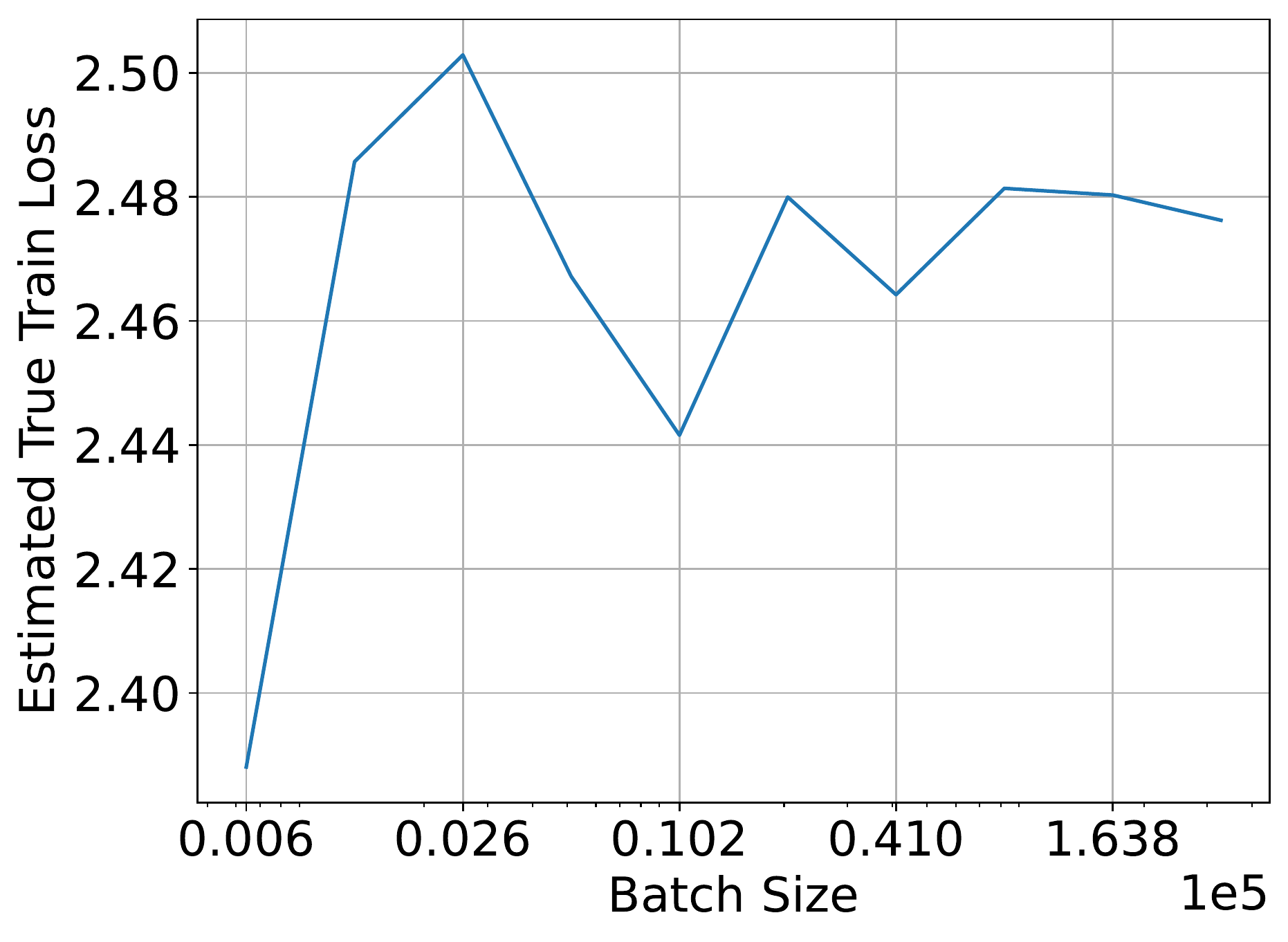}
	}
	{
		\includegraphics[scale=0.4]{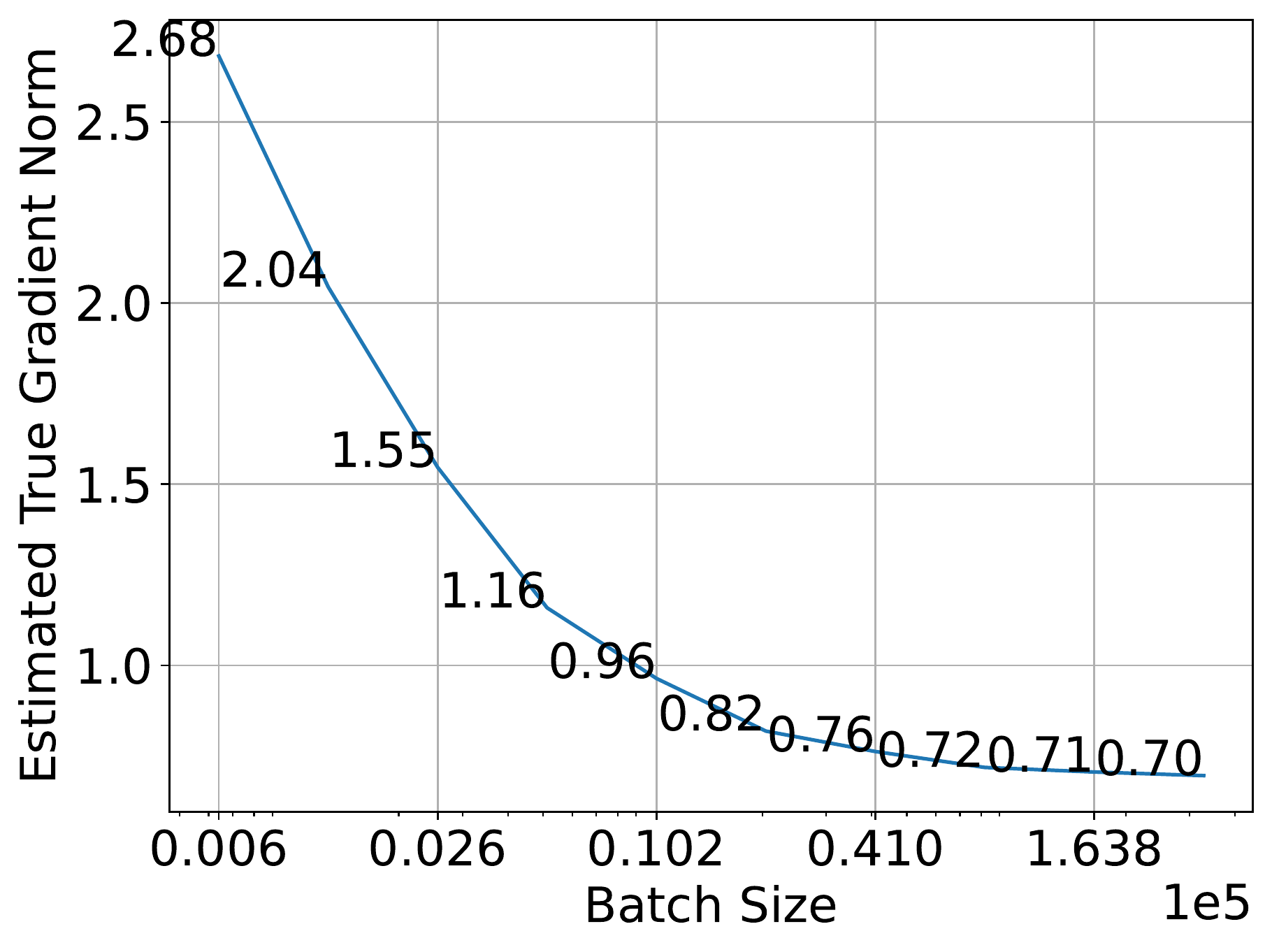}
	}
	
	{
		\includegraphics[scale=0.4]{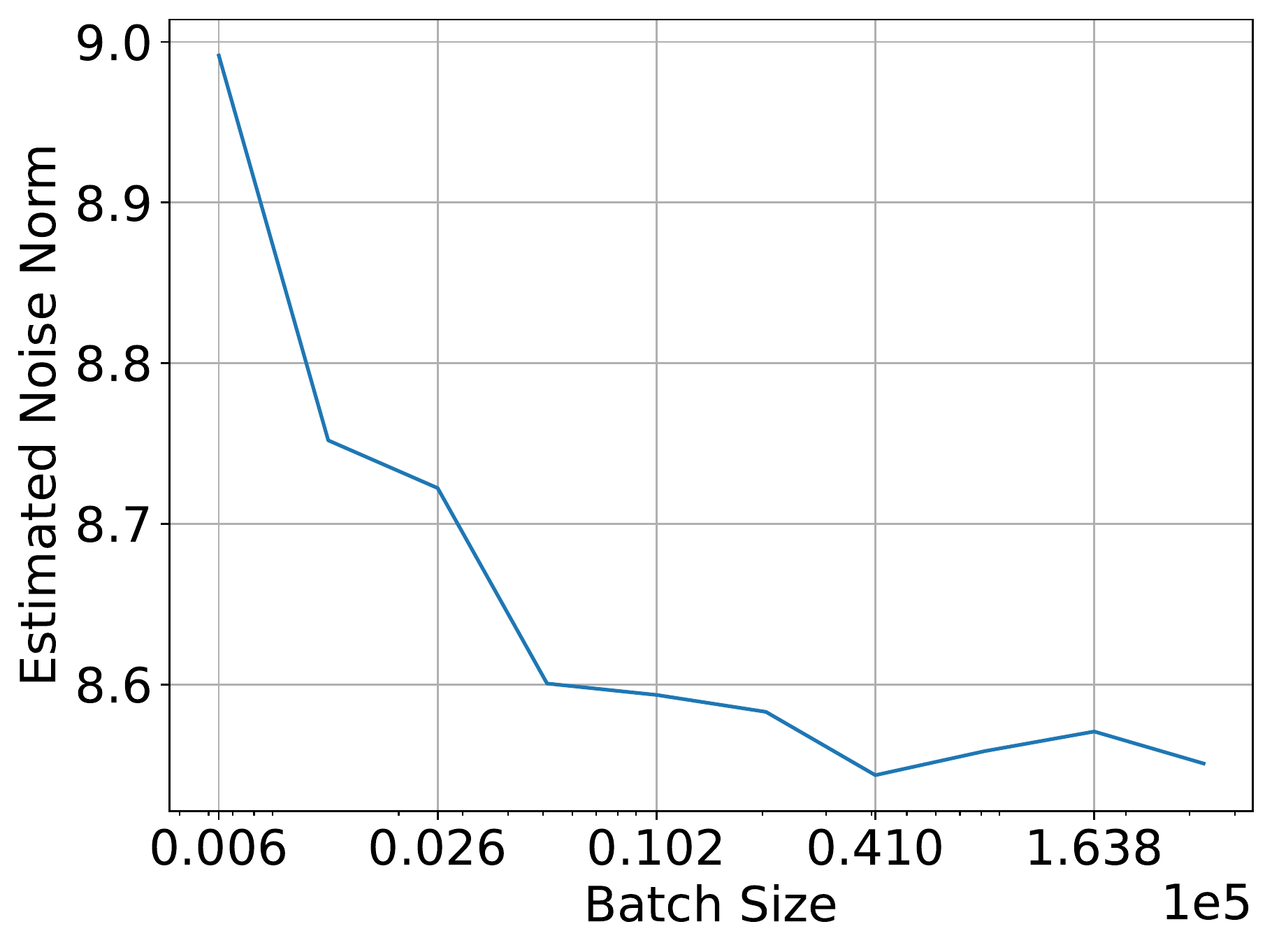}
	}{
		\includegraphics[scale=0.4 ]{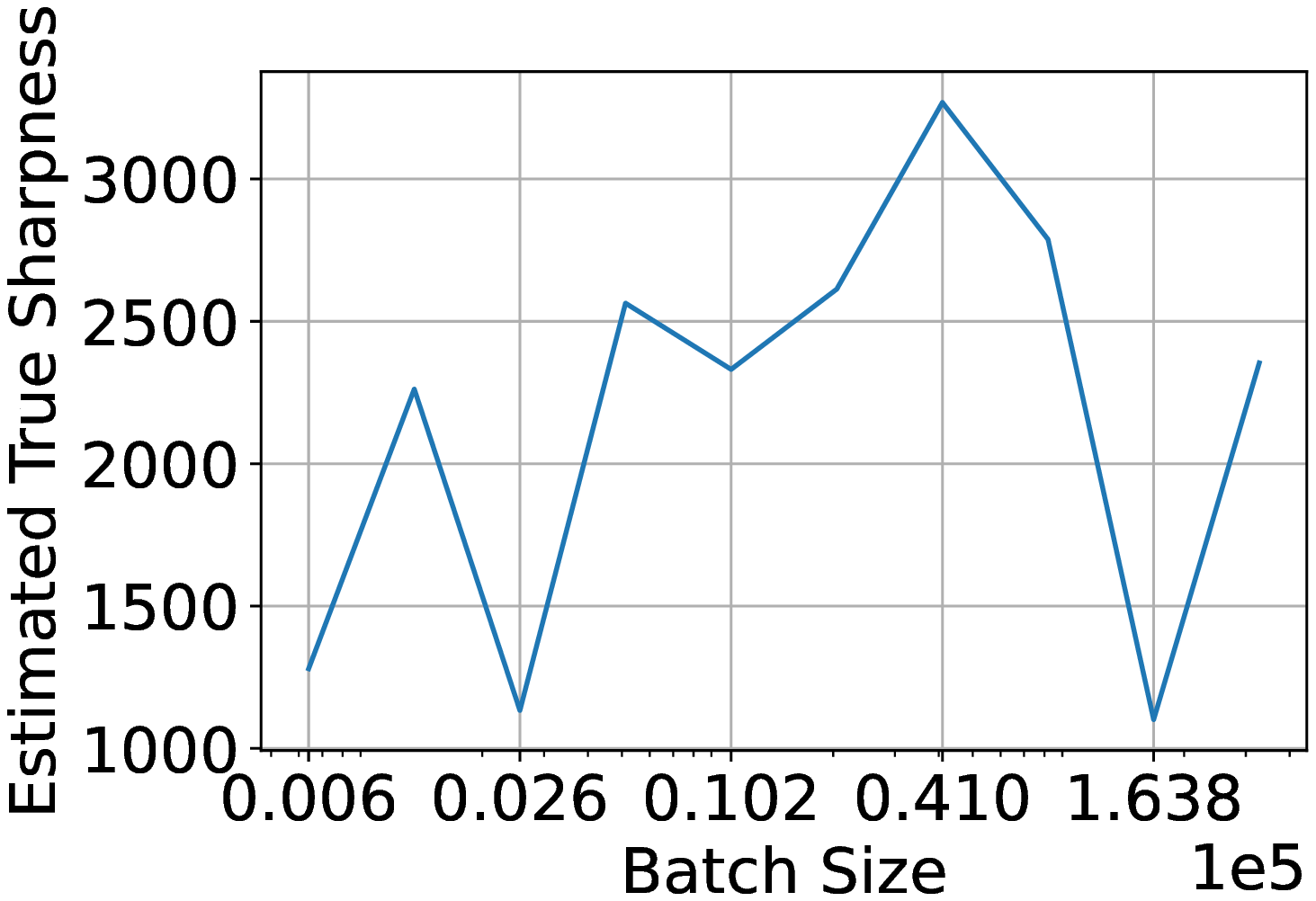}
	}
	\caption{
		The estimated stats vs batch size for ImageNet training.
	}
	\label{fig:imagenet-batch}
\end{figure*}

\begin{figure*}[htbp]
	\centering

	{
		\includegraphics[scale=0.4]{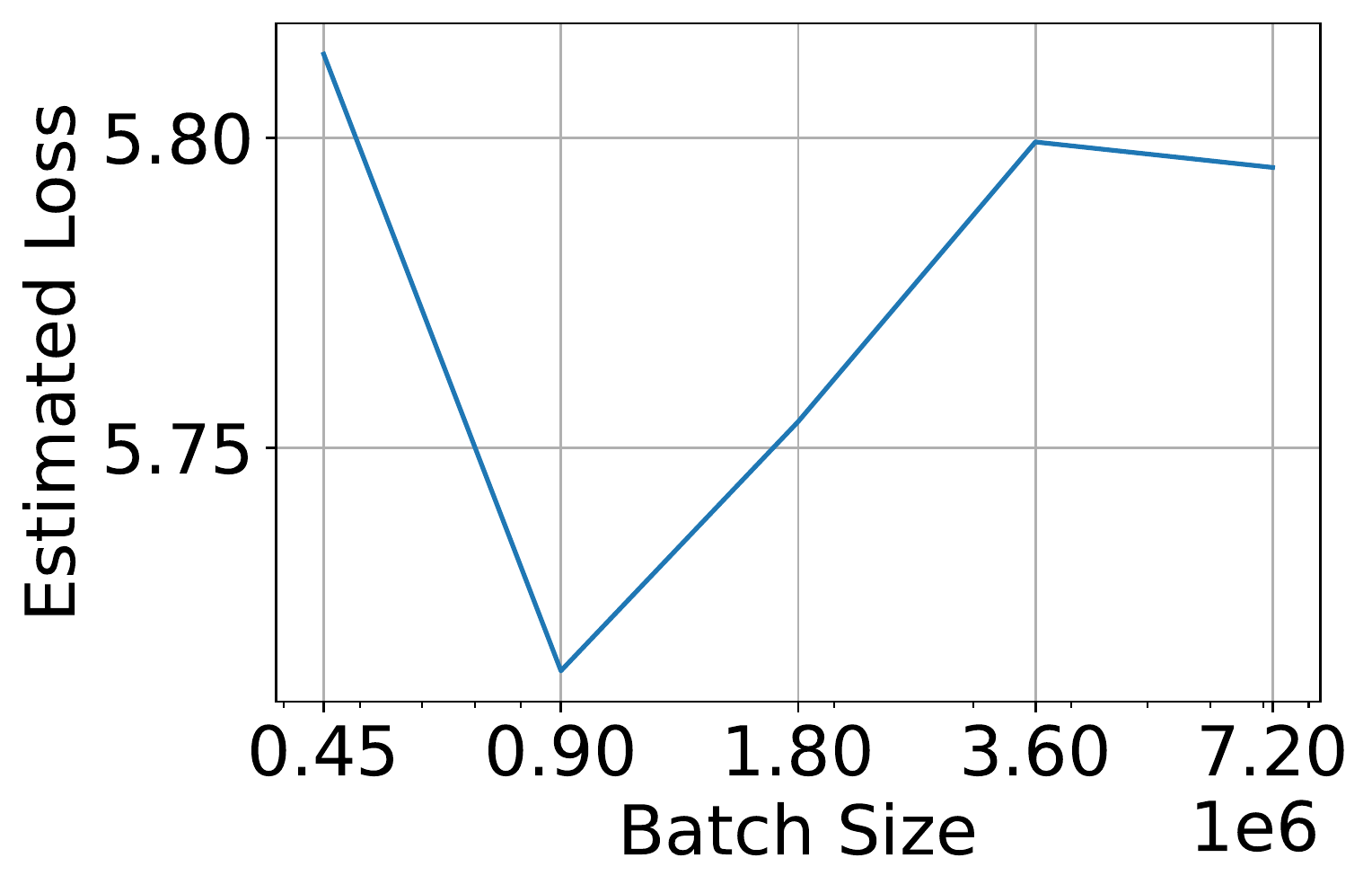}
	}
	{
		\includegraphics[scale=0.4]{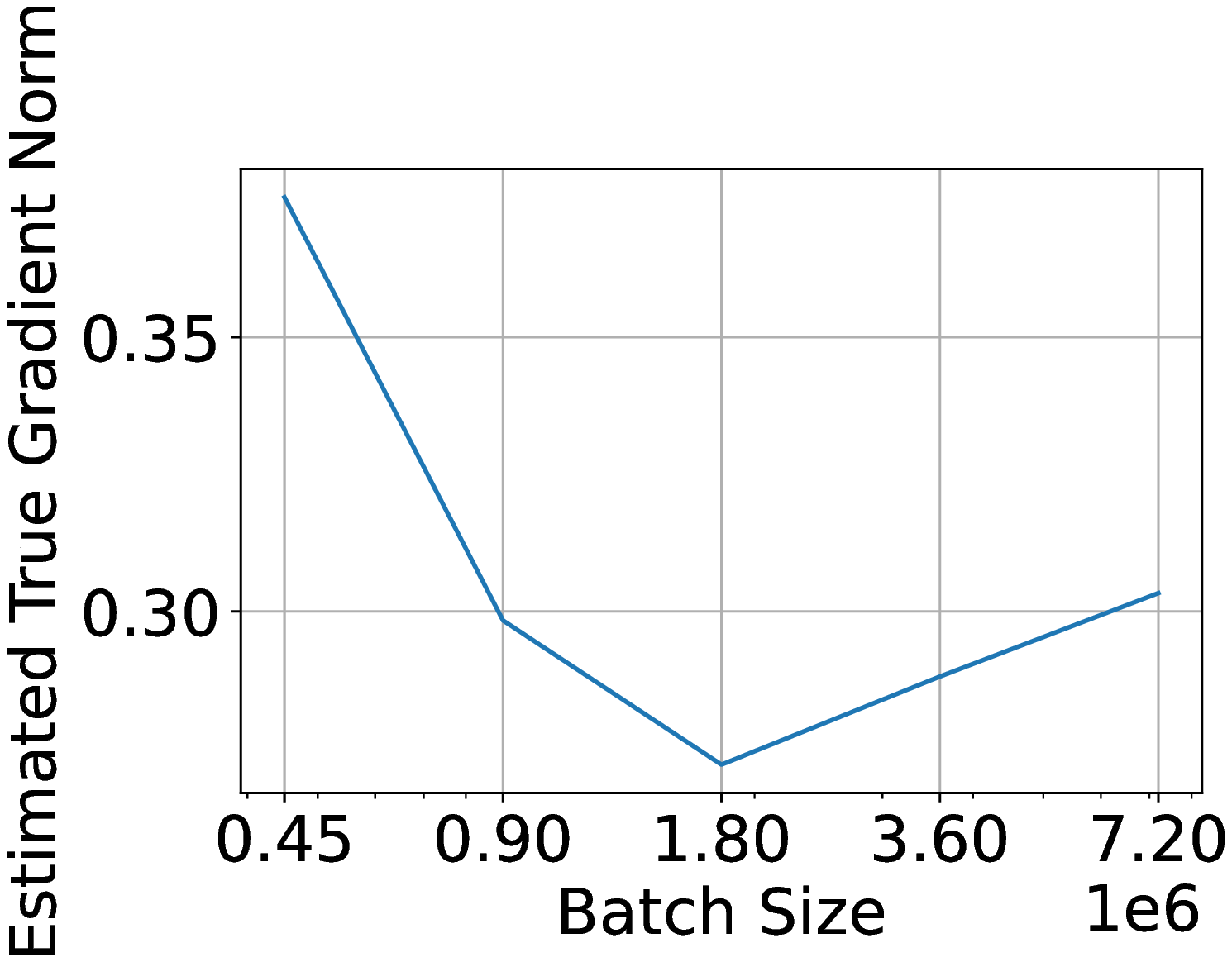}
	}
	
	{
		\includegraphics[scale=0.4]{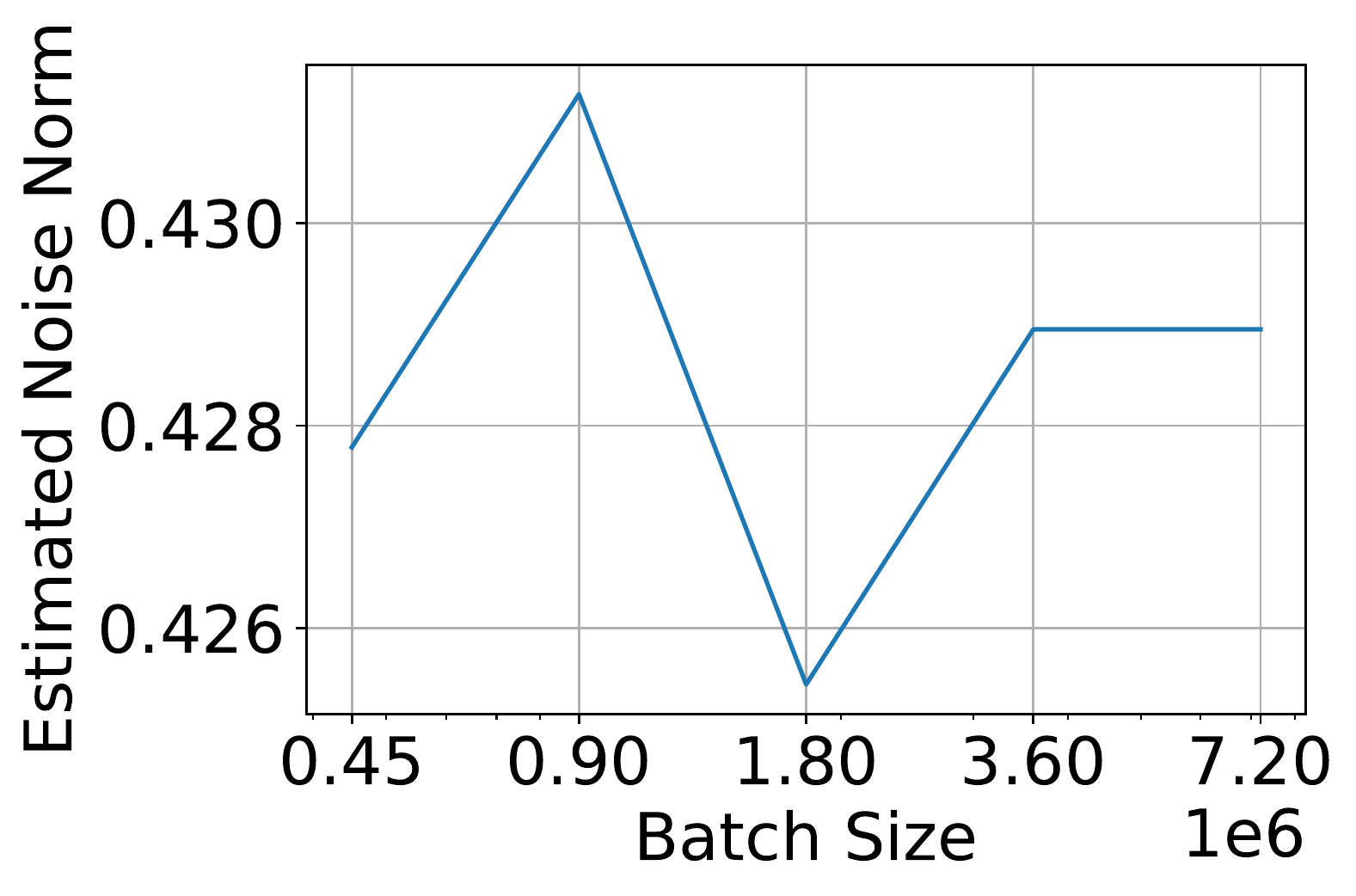}
	}{
		\includegraphics[scale=0.4]{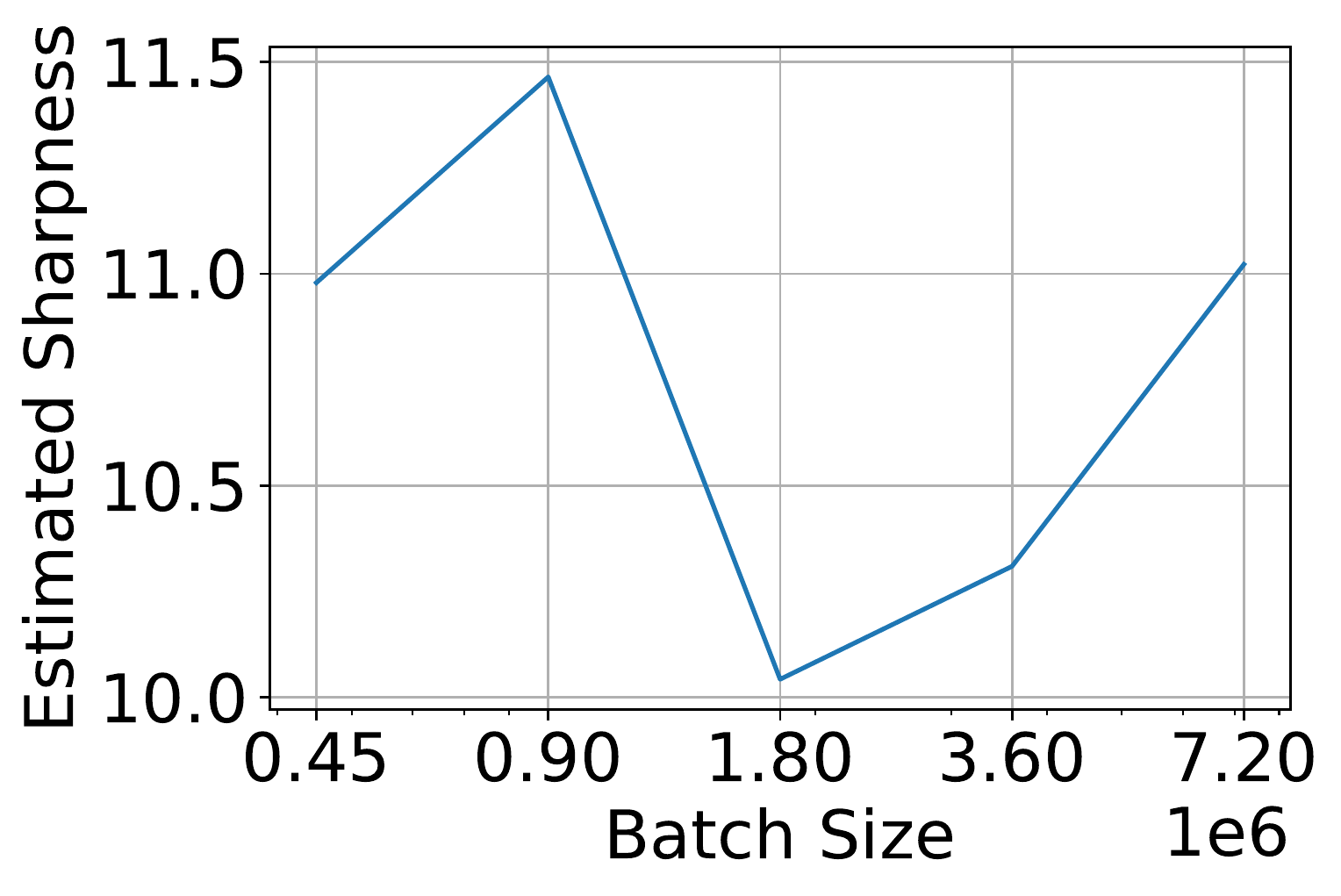}
	}
	\caption{
		The estimated stats vs batch size for WT103 training.
	}
	\label{fig:transformer-batch}
\end{figure*}

By Jensen's inequality, the estimated norms would be larger than the true value. However, the value should converge as the sampled batch size $m$ converges to the total data number $N$. We show in Figure~\ref{fig:imagenet-batch} and Figure~\ref{fig:transformer-batch} how these estimator values converge in practice. Based on these plots, we select the batch size to be $1.6 \times 10^5$ for ImageNet training and the token size to be $9 \times 10^5$ for the WT103 training. These sample sizes give a high enough precision level for making the observations in previous sections. Note that the estimated smoothness for the ImageNet experiment has very large variations, and hence we didn't make many comments on that plot throughout this work.

\newpage

\section{Convergence of function values}\label{proof:func-value}

One key intermediate result for the proof main theorem relies on the convergence of empirical measures when the iterates are updated by a compact continuous function $F$.  

\begin{theorem}[convergence of function values]\label{thm:function-value}
	Consider a continuous scalar function $\phi: \cX \to \R$.  Assume that the update map $F$ has the property  that $\phi \circ F : \cX \to [-M, M]$ has a bounded value for any $\theta \in \cX$ , then with probability $1 - \delta$ over the randomness of $F$,
	\begin{align}
		\abs{\E_{\theta \sim \mu_n}\left[\phi(\theta) -  \phi(F(\theta))\right]} = \cO\left(\frac{\log(1/\delta)}{\sqrt{n}}\right).
	\end{align}
\end{theorem}
\begin{proof}
The proof can be found in Appendix~\ref{proof:func-value}. 
\end{proof}

\begin{proof}
By the fact that $\phi \circ F: \cX \to \R$ has bounded value $[-M, M]$, we can denote the subgaussian norm at $\theta$ as 
\begin{align*}
	\sigma(\theta) = \inf \{\sigma > 0 | \Pr (\| \phi(F(\theta)) -  \E[\phi(F(\theta))] \| \ge t) \le 2e^{-t^2/2\sigma^2}\}.
\end{align*}
In fact, $\forall \theta, \sigma(\theta)  \le M < \infty$. Hence, we can further denote the upperbound on the sub-Gaussian norm as
$$\sigma = \sup_\theta  \sigma(\theta). $$
Then we consider two distributions. One is the empirical distribution of a sampled trajectory,
\begin{align*}
	\mu_n = \frac{1}{n} \sum_{t=1}^n \delta_{\theta_t}.
\end{align*}
The other one is the pushforward distribution $\mu_k(F^{-1})$ as defined in \eqref{eq:pushforward-st}. Then,

\begin{align*}
\E_{\theta \sim \mu_n}\left[\phi(\theta) -  \phi(F(\theta))\right] & = \frac{1}{n} \sum_{t=1}^n  \phi(\theta_t) -  \frac{1}{n} \sum_{t=1}^n  \phi(F(\theta_t))  \\
	&= \frac{1}{n} (\phi(\theta_0) - \phi(F(\theta^n))) + \frac{1}{n} \sum_{t=1}^{n}  \phi(\theta_t) - \phi(F(\theta_{-1})) \\
	&= \cO\left(\frac{1}{n}\right) +  \frac{1}{n} \sum_{t=1}^{n}  \phi(\theta_t) - \phi(F(\theta_{t-1})).
\end{align*}
Then the claim follows by applying Hoeffding's inequality on the second term.
\end{proof}

% \section{Proof of Lemma~\ref{lem:periodic}}\label{proof:periodic}

% \begin{proof}
% First, we construct a function 

% \begin{equation}
% 	f(\theta) =
% 	\begin{cases}
% 		-\frac{1}{n+1}\theta & \theta\le 0, \\
% 		\frac{n}{n+1}\theta & \theta> 0.
% 	\end{cases}
% \end{equation}
% We notice that if $\theta_0 = \frac{0.2}{n+1}$, then $\theta_k = -1 + \frac{k -0.8}{n+1}, k \le n$ and $\theta_0, \theta_1, \theta_2 ... \theta_{n}$ is periodic with period $n+1$. We further notice that within a period $\theta_1, ..., \theta_{n+1}$, the function values are distinct, and    $|f(\theta_0) - f(\theta_{1})  | \ge 0.5$. Therefore, for any $n > 0$, there exists a $t > n$ such that
% \begin{align*}
% 	|M_t - M_{t-1}| = \frac{1}{n} |f(\theta_{t+n-1}) - f(\theta_{t-1})|\ge \frac{1}{2n}.
% \end{align*}
% \end{proof}

\section{Proof of Theorem~\ref{thm:invariant}}\label{proof:invariant}
\begin{proof}
	
	Since $\cX$ is a compact metric space, we can find a dense countable set of the family of continuous functions $C(\cX)$, denoted as $\{\phi_1,  \phi_2...\}$.  Since $\cX$ is compact, we have that $\mu_k (\phi_j)$ exists for any $k, j$. Therefore, by the diagonal argument, there exists a subsequence $\{n_k\}_k$ such that for all $j = 1, 2, ...$, 
	\begin{align*}
		\lim_{k \to \infty} \frac{1}{n_k} \sum_{l \le n_k} \phi_j(\theta_{l}) = J(\phi_j).
	\end{align*}
	Then by denseness of the set $\{\phi_1,  \phi_2...\}$, we know that the above limit also exists for any $\phi \in C(\cX)$.  Denote the functional as 
	\begin{align}
		J(\phi) = \lim_{k \to \infty} \frac{1}{n_k} \sum_{l \le n_k} \phi(\theta_{l}).
	\end{align}
	Since $J$ is obviously linear and bounded, there exist a unique probability measure $\phi$ such that $J(\phi) = \mu (\phi)$.
	
	The invariance of $\mu$ follows by the fact that for any continuous $\phi$,
	\begin{align}
		\lim_{k \to \infty} |\E_{\theta \sim \mu_k}\left[\phi(\theta) -  \phi(F(\theta))\right] | =& \lim_{k \to \infty} \frac{1}{n_k} \sum_{l\le n_k} \phi(\theta_{l}) - \frac{1}{n_k} \sum_{l\le n_k} \phi(F(\theta_{l}) ) \nonumber    \\
		=& \lim_{k \to \infty} \frac{1}{n_k} \sum_{l\le n_k} \phi(\theta_{l}) - \frac{1}{n_k} \sum_{l\le n_k} \phi(\theta_{l+1}) \nonumber   \\
		&+ \lim_{k \to \infty} \frac{1}{n_k} \sum_{l\le n_k} \phi(\theta_{l+1}) - \frac{1}{n_k} \sum_{l\le n_k} \phi(F(\theta_{l})) \nonumber  \\
		=& 	\lim_{k \to \infty} \frac{1}{n_k}  (\phi(\theta_1) - \phi(\theta_{n_k+1}))\nonumber   \\
		&+ \lim_{k \to \infty} \frac{1}{n_k} \sum_{l \le n_k}(\phi(\theta_{l+1}) - \phi(F(\theta_{l}))
		)\to 0.
	\end{align}
	In the last line, the first term goes to zero by boundedness of function value on the compact set. The second term goes to zero by noticing that the sequence 
	\begin{align*}
		M_n = \frac{1}{n} \sum_{l \le n}(\phi(\theta_{l+1}) - \phi(F(x_{l}))
	\end{align*}
	is a martingale sequence. By the fact that each the induced martingale difference sequence has uniformly bounded sub-Gaussian norm, we can apply Hoeffding's inequality and know that $M_n$ converge in probability to 0, which implies convergence in distribution.

\end{proof}

\section{Proof of Lemma~\ref{lem:cross_entropy}}

\begin{proof}\ \newline

	\begin{enumerate}
		\item Let $x_m=(\max_i x_i+\min_i x_i)/2$ and define $z_i=x_i-x_m$. We know $\abs{z_i}\le c/2$. Then we have
		\begin{align*}
			\abs{\ell(x,y)}=& \abs{z_y+x_m-\log\left(\textstyle\sum_{j=1}^d e^{z_j+x_m}\right)}\\
			=& \abs{z_y-\log\left(\textstyle\sum_{j=1}^d e^{z_j}\right)}\\
			\le& c/2+\log\left(de^{c/2}\right)\\
			=& c+\log d.
		\end{align*}
		
		\item As $\ell$ is differentiable, it suffices to bound its gradient norm. For any fixed $1\le k\le d$, we have
		\begin{align*}
			\frac{\partial \ell(x,y)}{\partial x_k}=\delta_{y,k}-\frac{e^{x_k}}{\textstyle\sum_{j=1}^d e^{x_j}}.
		\end{align*}
		Then we can bound
		\begin{align*}
			\norm{\frac{\partial \ell(x,y)}{\partial x}}_2=& \sqrt{\textstyle\sum_{k=1}^d \left(\frac{\partial \ell(x,y)}{\partial x_k}\right)^2}\\
			\le& \sqrt{1+\frac{\textstyle\sum_{k=1}^d e^{2x_k}}{\left(\textstyle\sum_{j=1}^d e^{x_j}\right)^2}}\\
			\le& \sqrt{2}.
		\end{align*}
	\end{enumerate}
	
\end{proof}

\section{Proof of Theorem~\ref{thm:compact_domain}}

\begin{proof}
	Denote $\rho=w c_\sigma$. Then it is easy to show that within $C_w$, we have $\norm{x_l}_2\le \rho^l$ for every $l$.
	We define $z_{l+1}=W_l\theta_l$ and thus $x_l=\sigma_l(z_l)$. For any mini-batch $\cB=\{(x^i,y^i)\}_{i=1}^m$, we can bound the gradient norm of the loss.
	\begin{align*}
		\abs{\frac{\partial L_\cB}{\partial W_l}}&=\abs{\frac{1}{m}\sum_{i\in\cB}x^i_l (\nabla_x\ell(x_L^i,y^i))^\top D_L^{(i)} \left(\prod_{s=l+1}^{L-1}W_sD_s^{(i)}\right)}\le c_{\ell}c_{\sigma}\rho^{L-1}\le \gamma w,
	\end{align*}
	where we define $D_l^{(i)}=\text{Diag}(\sigma'_l(z_l^i))$.
	By the SGD rule, we have
	\begin{align*}
		W_{l}^{k+1}=&(1-\eta\gamma)W_{l}^k-\eta \frac{\partial L_\cB}{\partial W_l}.
	\end{align*}
	Choosing $\eta\le 1/\gamma$, if $\norm{W_l^{k}}_{\text{op}}\le w$, we also have
	\begin{align*}
		\norm{W_l^{k+1}}_{\text{op}}\le (1-\eta\gamma)w+\eta\gamma w \le w.
	\end{align*}
	By induction on $k$, the iterates of SGD optimizing the above objective always lie in $C_w$ if the stepsize satisfies $\eta\le 1/\gamma$. Then we have $\norm{x_L^i}_2\le \rho^L$ and can bound that for any $k$ $$\abs{L_{S}(\theta_k)}\le \log d +c_\ell \norm{x_L^i}_2\le \log d+\left(\frac{\gamma}{c_\ell c_\sigma^2}\right)^{L/(L-2)}.$$  The claim then follows by applying Theorem~\ref{thm:function-value}.
\end{proof}

\section{Proof of Theorem~\ref{thm:bn_bounded}}

\begin{proof}
	We first show that the coordinates of $\hat{x}_L^i$ are bounded. 
	\begin{align*}
		\abs{\hat{x}_L^i}=\frac{\abs{x_{L-1}^i-\mu_{\cB,L-1}}}{\sqrt{\frac{1}{m}\sum_{i\in\cB} \left( x_{L-1}^i-\mu_{\cB,L-1}\right)^{ 2}+\epsilon}}\le \sqrt{m}. 
	\end{align*}
	As in Theorem~\ref{thm:compact_domain}, we show that during the training process, $a_L$ always satisfies $|a_L|\le 2\sqrt{m}/\gamma$. Note that
	\begin{align*}
		\abs{\frac{\partial L_{\cB}}{\partial a_L}}=&\abs{\frac{1}{m}\sum_{i\in\cB}\hat{x}_L^i\cdot \nabla_x  \ell({x}_L^i,y^i)}\\
		=&\abs{\frac{1}{m}\sum_{i\in\cB} \sum_{k=1}^d (\hat{x}_L^i)_k \left( \delta_{y,k}- \frac{e^{({x}_L^i)_k}}{\sum_{j=1}^d e^{({x}_L^i)_j}} \right)}\\
		\le & \max_{i\in\cB}\abs{(\hat{x}_L^i)_y-\frac{\sum_{k=1}^d (\hat{x}_L^i)_ke^{({x}_L^i)_k}}{\sum_{j=1}^d e^{({x}_L^i)_j}}}\\
		\le& 2\sqrt{m}.
	\end{align*}
	Therefore if $\abs{a_L^k}\le 2\sqrt{m}/\gamma$, we have
	\begin{align*}
		\abs{a_L^{k+1}}=\abs{(1-\eta\gamma)a_L^k-\eta \frac{\partial L_{\cB}}{\partial a_l}}\le (1-\eta\gamma)\cdot 2\sqrt{m}/\gamma+2\eta\sqrt{m}\le 2\sqrt{m}/\gamma.
	\end{align*}
	Then by induction, the above is true for every $k$. By Lemma~\ref{lem:cross_entropy}, we have for every $k$
	\begin{align*}
		L_\cB(\theta_k)\le 4m/\gamma+\log d.
	\end{align*}
	Then the training loss $\abs{L_\cB(\theta_k)}\le 4m/\gamma+\log d$ is bounded during the training process if the stepsize satisfies $\eta\le 1/\gamma$. The theorem follows by applying Theorem~\ref{thm:function-value}.
\end{proof}

\section{Proof of Theorem~\ref{thm:smaller-step}}\label{proof:small}

\begin{proof}
	For simplicity, we denote
	\begin{align*}
		f(\theta) &:= L_S(\theta),\\
		\delta &:= \E_{\theta \sim \mu}[\|\nabla f(\theta)\|_2^2] > 0.
	\end{align*}
	
	By compactness of  $\cX$, we could denote the following quantities:
	\begin{align*}
		G  &= \sup_{\theta \in \cX} \|g(\theta)\|_2 < \infty,\\
		M^2 &= \sup_{\theta, \zeta \in \cX} \E_{z \sim \text{unif}[\theta, \zeta]}[\| \nabla^2 f(z) - \E_{z' \sim \text{unif}[\theta, \zeta]}[\nabla^2 f(z')] \|_{\text{op}}^2]  < \infty,\\
	\end{align*}

	For clarity, note that for any function $f: \cX \to \R^d$,
	\begin{align*}
		\E_{z \sim \text{unif}[\theta, \zeta]}[f(z)] = \int_0^1 f(t\theta + (1-t)\zeta)dt.
	\end{align*}
	Therefore, we have that for any $c \in (0, 1)$
	\begin{align*}
		&  \int_0^1 \| \nabla^2 f(ct\theta + (1-ct)\zeta) - \E_{z \sim \text{unif}[\theta, \zeta]}[\nabla^2 f(z)] \|_{\text{op}}^2  dt\\
		=   & \frac{1}{c} \int_0^c \| \nabla^2 f(t\theta + (1-t)\zeta) - \E_{z \sim \text{unif}[\theta, \zeta]}[\nabla^2 f(z)] \|_{\text{op}}^2  dt\\
		\le & \frac{1}{c}\E_{z \sim \text{unif}[\theta, \zeta]}[\| \nabla^2 f(z) - \E_{z'}[\nabla^2 f(z')] \|_{\text{op}}^2].
	\end{align*}

	Therefore, by Jensen's inequality, we have
	\begin{align}\label{eq:curve-concentrate}
		\int_0^1 \| \nabla^2 f(ct\theta + (1-ct)\zeta) - \E_{z \sim \text{unif}[\theta, \zeta]}[\nabla^2 f(z)] \|_{\text{op}} dt \le \sqrt{\frac{M^2}{c}}.
	\end{align}
	
	By applying Taylor expansion twice we get the following equations,
	\begin{align*}
		\E_{\theta, F}[f(F(\theta)) - f(\theta) ]  & = \E_{\theta, g}[f(\theta - \eta g(\theta)) - f(\theta) ] \\
		&= \E_{\theta, g}[ -\eta \int_{0}^1 \iprod{g(\theta)}{\nabla f(\gamma_{\theta, g(\theta)}(\eta t))} dt ] \\
		&= \E_{\theta, g}[ -\eta \| \nabla f(\theta)\|_2^2   - \eta  \int_{0}^1 \iprod{g(\theta) - \nabla f(\theta)}{\nabla f(\gamma_{\theta, g(\theta)}(\eta t))} dt ]   \\
		& - \E_{\theta, g}[\eta  \int_{0}^1 \iprod{ \nabla f(\theta)}{\nabla f(\gamma_{\theta, g(\theta)}(\eta t)) -  \nabla f(\theta)} dt ] \\
		& = \E_{\theta, g}[ -\eta \| \nabla f(\theta)\|_2^2   - \eta  \int_{0}^1 \iprod{g(\theta) - \nabla f(\theta)}{\nabla f(\gamma_{\theta, g(\theta)}(\eta t)) - \nabla f(\theta)} dt  ]  \\
		& - \E_{\theta, g}[\eta  \int_{0}^1 \iprod{ \nabla f(\theta)}{\nabla f(\gamma_{\theta, g(\theta)}(\eta t)) -  \nabla f(\theta)} dt ] \\
		&= \E_{\theta,  g}[ -\eta \| \nabla f(\theta)\|_2^2   -  \eta ^2 \int_0^1 \int_0^1  \iprod{g(\theta)}{ \nabla^2 f(\gamma_{\theta, g}(t\tau \eta)) g(\theta)} dtd\tau  ].
	\end{align*}
	where $\gamma_{\theta, g(\theta)}(r) = \theta - r g(\theta)$ denotes the line segment. In the second line, we applied the fundamental theorem of calculus. In the second line, we add and subtracted same terms. In the fourth equality, we used the unbiasedness of noise. In the last line, we combined the last two terms and applied Taylor expansion again.
	
	By invariance of the function value, we get that 
	\begin{align}\label{eq:equilibrium}
		&\E_{\theta,  g}[ -\eta \| \nabla f(\theta)\|_2^2   -  \eta ^2 \int_0^1 \int_0^1  \iprod{g(\theta) }{ \nabla^2 f(\gamma_{\theta, g}(t\tau \eta)) g(\theta)} dtd\tau  ] = 0 \nonumber\\
		\implies &\E_{\theta,  g}[ \| \nabla f(\theta)\|_2^2]    = \E_{\theta,  g}[\eta  \int_0^1 \int_0^1  \iprod{ g(\theta) }{ \nabla^2 f(\gamma_{\theta, g}(t\tau \eta)) g(\theta)} dtd\tau  ] .
	\end{align}
	
	Therefore we have that 
	\begin{align*} 
		&\E_{\theta, F'}[f(F'(\theta)) - f(\theta) ] \\
		= &\E_{\theta,  g}[ -c\eta \| \nabla f(\theta)\|_2^2   -  c^2\eta ^2 \int_0^1 \int_0^1  \iprod{g(\theta) }{ \nabla^2 f(\gamma_{\theta, g}(t\tau \eta)) g(\theta)} dtd\tau  ] \\
		\le & c\eta \left( - \delta + c G^2 \sqrt{\tfrac{M}{c}}\right),
	\end{align*}
	where in the last line we used ~\eqref{eq:curve-concentrate}. The claim follows by setting $c$ small enough.
	
\end{proof}

\end{document}